\documentclass[times,twocolumn,final]{elsarticle}
\usepackage{medima_arxiv}
\usepackage{framed,multirow}
\usepackage{amssymb}
\usepackage{latexsym}
\usepackage{url}
\usepackage{cite}
\usepackage{amsmath,amssymb,amsfonts,bm}
\usepackage{booktabs}
\usepackage{graphicx,epstopdf}
\usepackage{textcomp}
\usepackage{subcaption}
\usepackage{url}
\usepackage{float}
\usepackage{tabularx,threeparttable}
\usepackage{makecell}
\usepackage{xcolor}
\usepackage[normalem]{ulem} 
\usepackage{tabulary}
\usepackage[colorlinks=true,citecolor=blue,urlcolor=blue]{hyperref}
\usepackage[]{color}
\usepackage{hhline}
\usepackage{array}
\usepackage{amsfonts}
\usepackage{fancyhdr}
\usepackage{soul}
\usepackage{makecell}
\usepackage{bbm}
\usepackage{amssymb}% http://ctan.org/pkg/amssymb
\usepackage{pifont}% http://ctan.org/pkg/pifont
\newcommand{\xmark}{\ding{55}}%
\usepackage[noend]{algpseudocode}
\usepackage{algorithm}
\usepackage{flexisym}
\usepackage{eqparbox}
\usepackage{breqn}
\usepackage{afterpage}
\usepackage[leftcaption]{sidecap}
\newcommand{\blue}{}

\newcommand\Tstrut{\rule{0pt}{2.6ex}}         % = `top' strut
\newcommand\Bstrut{\rule[-0.9ex]{0pt}{0pt}}   % = `bottom' strut

\renewcommand{\algorithmiccomment}[1]{\textsf\bgroup\tiny//~#1\egroup}

\definecolor{newcolor}{rgb}{.8,.349,.1}
\hypersetup{pdfborder={0 0 0},
	colorlinks=true,
	linkcolor=black,
	citecolor=black,
	urlcolor=black}
\journal{Medical Image Analysis}
\begin{document}

\verso{Vinkle Kumar Srivastav \textit{et~al.}}

\begin{frontmatter}

\title{Unsupervised domain adaptation for clinician pose estimation and instance segmentation in the operating room}
%\tnotetext[tnote1]{Code will be made public at \url{https://github.com/CAMMA-public/HPE-AdaptOR} upon acceptance}

\author[1]{Vinkle \snm{Srivastav}\corref{cor1}}
\ead{srivastav@unistra.fr}
\author[2]{Afshin \snm{Gangi}}
\author[1,3]{Nicolas \snm{Padoy}\corref{cor1}}
\ead{npadoy@unistra.fr}
\cortext[cor1]{Corresponding authors: Tel.:+33 (0) 3 904 13530}

\affliation[1]{ICube, University of Strasbourg, CNRS, France}
\affliation[2]{Radiology Department, University Hospital of Strasbourg, France}
\affliation[3]{IHU Strasbourg, France}

\received{-}
\finalform{-}
\accepted{-}
\availableonline{-}
\communicated{-}

\begin{abstract}
    \begin{center}
        \rule{2.07\linewidth}{1.0pt}\\[0.15in]
    \end{center}
    \noindent{\bf{Abstract:}}
    The fine-grained localization of clinicians in the operating room (OR) is a key component to design the new generation of OR support systems. Computer vision models for person pixel-based segmentation and body-keypoints detection are needed to better understand the clinical activities and the spatial layout of the OR. This is challenging, not only because OR images are very different from traditional vision datasets, but also because data and annotations are hard to collect and generate in the OR due to privacy concerns. To address these concerns, we first study how joint person pose estimation and instance segmentation can be performed on low resolutions images {\blue with downsampling factors from 1x to 12x}. Second, to address the domain shift and the lack of annotations, we propose a novel unsupervised domain adaptation method, called \emph{AdaptOR}, to adapt a model from an \emph{in-the-wild} labeled source domain to a statistically different unlabeled target domain. We propose to exploit explicit geometric constraints on the different augmentations of the unlabeled target domain image to generate accurate pseudo labels and use these pseudo labels to train the model on high- and low-resolution OR images in a \emph{self-training} framework. Furthermore, we propose \emph{disentangled feature normalization} to handle the statistically different source and target domain data. Extensive experimental results with detailed ablation studies on the two OR datasets \emph{MVOR+} and \emph{TUM-OR-test} show the effectiveness of our approach against strongly constructed baselines, especially on the low-resolution privacy-preserving OR images. Finally, we show the generality of our method as a semi-supervised learning (SSL) method on the large-scale \emph{COCO} dataset, where we achieve comparable results with as few as \textbf{1\%} of labeled supervision against a model trained with 100\% labeled supervision. Code is available at \url{https://github.com/CAMMA-public/HPE-AdaptOR}.

     \vspace{-0.15in}
\end{abstract}
%along with \emph{mean-teacher} approach for stable training

\end{frontmatter}

\section{Introduction}
%\afterpage{
	\begin{figure*}[htb!]
		\includegraphics[clip, trim=0.0cm 1.3cm 0.0cm 0.0cm, width=1.0\linewidth]{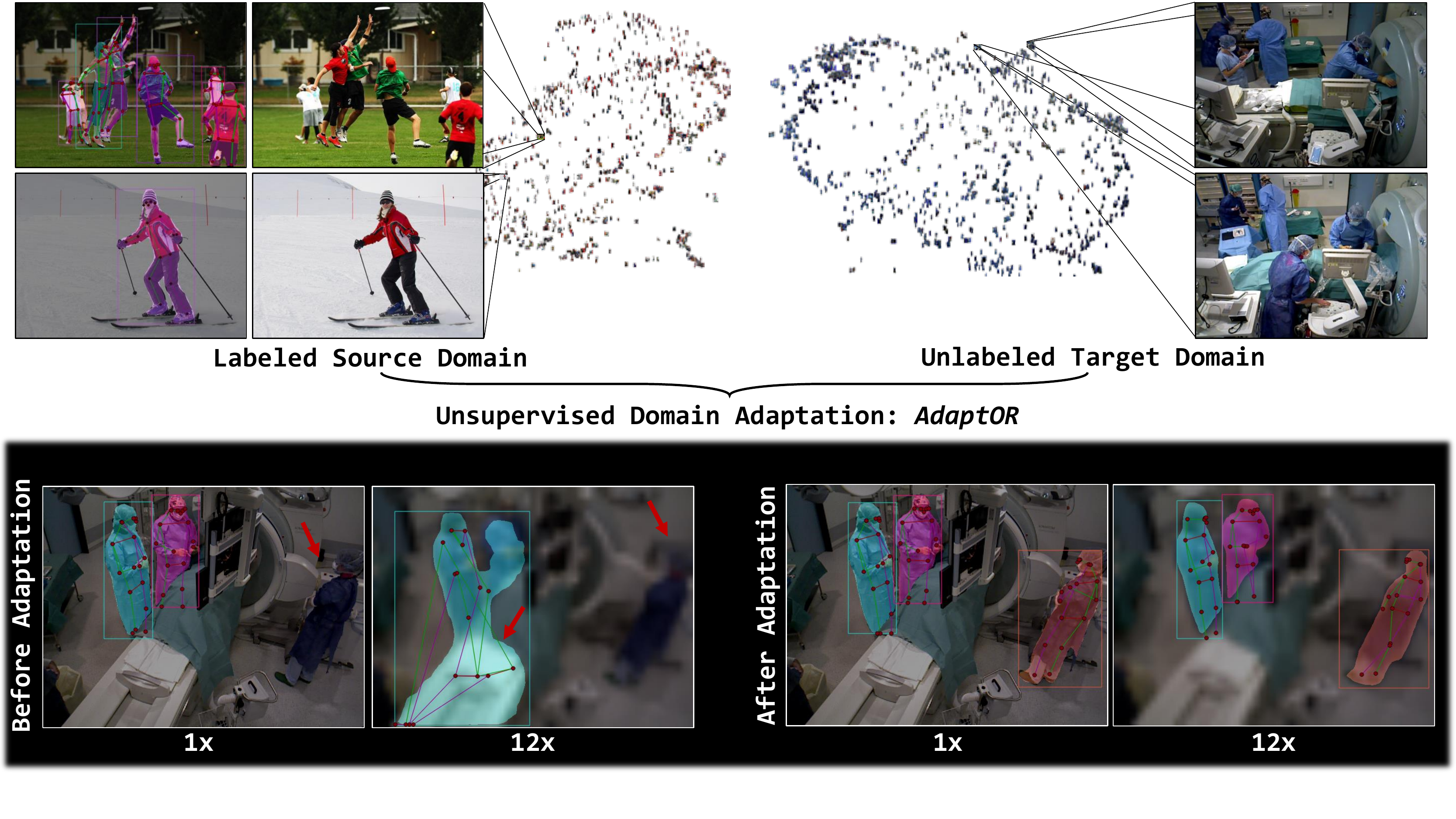}
		\caption{Global and instance-level visual differences between \emph{source domain} natural images and \emph{target domain} OR images. When a model trained on the source domain is applied to the unseen target domain, we see a substantial decrease in the localization accuracy and an increase in the missed detections. Our unsupervised domain adaptation method significantly improves the results on high and low-resolution OR images. The separate clusters of the source domain and the target domain images are obtained by running a dimension reduction technique: Uniform Manifold Approximation and Projection for Dimension Reduction (UMAP) \citep{mcinnes2018umap,pixplot}. The source and the target domain images are a subset of the COCO \citep{lin2014microsoft} and the MVOR \citep{srivastav2018mvor} datasets, respectively.}
		\label{fig:introduction}
	\end{figure*}
	
%}
The significant rise in the supervised deep-learning methods has paved the way for the visual understanding of persons in challenging environments. Recent progress has pushed its boundaries from coarse bounding box detection to more fine-grained pose estimation, providing keypoint-level understanding, and instance segmentation, providing pixel-level understanding. Joint person pose-estimation and instance segmentation aim to localize the body keypoints and estimate segmentation masks for all persons in a given image using a single model. It can support a variety of computer vision applications ranging from virtual try-on \citep{han2018viton},  smart video synthesis \citep{chan2019everybody}, human activity recognition \citep{song2021human} to self-driving cars \citep{liang2020polytransform}. The healthcare sector, especially the modern operating room (OR), could hugely benefit from such models to enable novel context-aware computer-assisted systems.

Novel context-aware systems in sensor-enhanced and visually complex modern ORs have the potential to streamline clinical workflow processes, detect adverse events, and support real-time decision making by automatically analyzing clinical activities \citep{padoy2019machine,vercauteren2019cai4cai,maier2020surgical,mascagni2021or}. This has been illustrated by the recent development of new OR applications such as activity analysis in robot-assisted surgery \citep{sharghi2020automatic}, semantic scene understanding of OR \citep{li2020robotic}, surgical workflow recognition in the OR \citep{kadkhodamohammadi2020towards,zhang2021real}, and radiation risk monitoring during hybrid surgery \citep{rodas2017see}. As clinicians are the main dynamic actors in the OR, models for joint person pose estimation and instance segmentation are key components in building various smart assistance applications. The radiation risk monitoring \citep{rodas2017see}, for example, needs such models to understand harmful exposure of radiations to the clinicians at pixel- and keypoint-level. Team activity analysis as another example \citep{dias2019physiological,soenens2021assessment} needs such models to understand interactions, non-verbal communications, and cognitive load, especially during critical phases of the surgery.

These systems, with immense promise to improve patient safety and care, however face hindrance due to the privacy-sensitive nature of the OR environment. Continuous monitoring by the ceiling-mounted cameras raises potential privacy concerns for the patients and clinicians. Therefore, the data from the cameras is often recorded at low-resolution to improve privacy, as suggested in the literature \citep{haque2018activity,srivastav2019human,srivastav2020human}. Developing person localization approaches for these spatially-degraded but privacy-preserving low-resolution images is consequently an important challenge that we introduce and tackle in this paper.

While person pose estimation and instance segmentation models have improved substantially in the unconstrained environments, they fail remarkably on the unseen \emph{target domains} due to the visual differences \citep{recht2018cifar,srivastav2018mvor}. The OR as a target domain also presents many challenges due to notable changes in the visual appearance at a global and instance level. Indeed, the OR has particular lighting conditions, and the clinicians wear loose clothes and surgical masks and occlude one another due to close proximity and instrument clutter. Fig.~\ref{fig:introduction} shows such global and instance-level visual differences between natural and OR images. One way to overcome such domain differences is to fine-tune a model on the manually labeled data from the target domain. For example, the authors in \citep{kadkhodamohammadi2017multi,belagiannis2016parsing,li2020robotic} developed fully-supervised approaches for multi-view 3D pose estimation and semantic segmentation for the OR. However, collecting the labels for the data is not only time-consuming and expensive - annotating a single image with pixel-level segmentation can take up to 90 minutes \citep{cordts2016cityscapes} -  but also particularly unscalable for the OR due to privacy concerns. The scalable and successful crowd-sourcing platforms, for example, Amazon Turk, can not be easily used for the privacy-sensitive OR environment to provide large-scale manually labeled data. Approaches that can adapt a model to the unseen and unlabeled target domain are therefore very promising.

In this work, we propose a novel unsupervised domain adaptation (UDA) approach, called \emph{AdaptOR}, for joint person pose estimation and instance segmentation. We aim to adapt a model from a labeled source domain, i.e., unconstrained natural images from \emph{COCO} \citep{lin2014microsoft} to an unlabeled target domain, i.e., constrained low-resolution OR images with {\blue downsampling factors from 1x to 12x}. The UDA methods have been extensively studied for various computer vision tasks ranging from image classification \citep{zhuang2020comprehensive}, object detection \citep{oza2021unsupervised} to semantic segmentation \citep{toldo2020unsupervised}. Unlike the existing UDA approaches that have primarily been applied to general object classes, we aim to study the UDA for a single but highly challenging ``person'' class inside the visually complex OR environment while simultaneously exploiting articulated ``person'' class properties for effective domain {{\blue adaptation}.

We choose Mask R-CNN \citep{he2017mask} as our backbone model for joint person pose estimation and instance segmentation, which is primarily designed for a single domain fully supervised training. Inspired from UDA for image {\blue classification} \citep{chang2019domain}, we propose \emph{disentangled feature normalization} (DFN) for our backbone model to train it on two statistically different domains. DFN replaces every feature normalization layer in the feature extractor of the backbone model with two feature normalization layers: one for the source domain and another for the target domain. With the improved design, the backbone model expects an input batch containing half the images from the source domain and another half from the target domain. DFN therefore {\blue modifies} the multi-task loss function to compute and weigh the loss differently for the two domains. The use of separate feature normalization layers for the two domains effectively disentangle the feature learning and stabilizes the training.

Given a backbone model with the ability to train on two statistically distinct domains, we build our approach based on a \emph{self-training} framework \citep{sohn2020fixmatch,liu2021unbiased,deng2021unbiased}, where we aim to predict similar predictions from a model under different augmentations of the same image, thereby taking the confident predictions from one augmented image - called \emph{weakly} augmented image - as pseudo labels for the other augmented image - called \emph{strongly} augmented image. {\blue Unlike} image classification tasks \citep{berthelot2019remixmatch,sohn2020fixmatch} where the model predictions need to be invariant to the different augmentations applied to the input image. The spatial localization tasks such as pose estimation or instance segmentation however can change the model predictions under certain geometric augmentations, e.g., random-flip or random-resize. Thankfully, these changes in the predictions need to satisfy \emph{transformation equivariant constraints} i.e., prediction labels also need to be transformed according to the applied geometric augmentations. We therefore use the \emph{transformation equivariant constraints} to add explicit geometric constraints on the  \emph{weakly} and the \emph{strongly} augmented unlabeled images to generate high-quality pseudo labels; for example, the random-flip operation has to exploit the chirality property \citep{yeh2019chirality} for pose estimation to map the keypoints to the horizontally flipped image.

To improve the performance of the model on low-resolution OR images as needed {\blue to improve the privacy}, we also propose to extend the data augmentation pipeline with a \emph{strong-resize} augmentation for the \emph{strongly} augmented image by applying two resize operations on the input image: a down-sampling and an up-sampling operation with a scaling factor randomly chosen between 1x to 12x. It generates heavily blurred images (see example downsampled images in figure \ref{figure:mvor-tum}) that naturally extend our approach to the privacy-preserving low-resolution images. Training the model using the two sets of weak and strong augmentations also enforces consistency regularization \citep{tarvainen2017mean,sajjadi2016regularization, sohn2020fixmatch}: a popular regularization technique utilized in a semi-supervised learning (SSL). The SSL is closely related to the UDA and aims to generalize a model to the same domain with limited labeled and large-scale unlabeled data.

We further extend our approach with \emph{mean-teacher} for stable training \citep{tarvainen2017mean}, where instead of using a single model to generate and consume the pseudo labels, we create two copies of a given source domain trained model: a \emph{teacher} and a \emph{student} model. The \emph{teacher} model generates the pseudo labels on the \emph{weakly} augmented image that is used by the \emph{student} model to train on the corresponding \emph{strongly} augmented image. The weights of the \emph{teacher} model are updated using temporal ensembling of the weights of the \emph{student} model, thereby helping it to improve its predictions due to ensembling while simultaneously generating better pseudo labels to improve the \emph{student} model. Fig.~\ref{fig:architecture} illustrate the complete architecture of our approach.

We evaluate our approach on the two OR datasets: \emph{MVOR+} \citep{srivastav2018mvor,srivastav2020human} and \emph{TUM-OR-test} \citep{belagiannis2016parsing}. The \emph{MVOR+} dataset is extended from the public \emph{MVOR} \citep{srivastav2018mvor} with full-body keypoints in COCO format for all the persons. The default annotation in the \emph{TUM-OR-test} contains only the six common COCO keypoints in the upper body bounding box. Therefore, we re-annotate the \emph{TUM-OR-test} using a semi-automatic approach\footnote{We will release \emph{MVOR+} dataset and the new \emph{TUM-OR-test} annotations along with the source code at \url{https://github.com/CAMMA-public/HPE-AdaptOR}.}. Both \emph{MVOR+} and \emph{TUM-OR-test} do not contain ground-truth for the person instance masks. We therefore evaluate the mask segmentation results by computing tight bounding boxes around the prediction masks and comparing them with the corresponding ground-truth bounding boxes, along with qualitative results\footnote{\url{https://youtu.be/gqwPu9-nfGs}}. We show that our approach performs significantly better after domain adaption and against strongly constructed baselines, especially on low-resolution OR images even downsampled up to \textbf{12x}. As our backbone model based on Mask R-CNN performs person bounding box detection by design, we use the model to evaluate for the person bounding boxes and show significant improvements in the bounding box detection results. We also conduct extensive ablation studies to shed light on the different components of our approach and their contributions to the results. The figure \ref{fig:introduction} shows a comparative qualitative result before and after the domain adaptation.

Finally, without bells and whistles, our UDA approach can be easily used as an SSL approach on the same domain dataset - by using regular feature normalization instead of DFN. We show the generality of our approach as an SSL method on the same domain COCO dataset with different percentages of supervision. With as few as \textbf{1\%} of labeled supervision, we obtain 57.7\% (38.2 keypoint AP) in the pose estimation and 72.3\% (36.1 mask AP) in instance segmentation, a strong improvement against the model trained with 100\% of labeled supervision (66.2 keypoint AP and 49.9 mask AP). These initial valuable baselines for the joint person pose estimation and instance segmentation could help foster SSL research on large-scale public datasets. We can summarize the contributions of our work mainly in the following five aspects:

\begin{enumerate}
	\item We propose to study joint pose estimation and instance segmentation on OR images at different low resolutions to address the privacy concerns in the OR.
	\item We propose a novel UDA approach to adapt the model to the unlabeled OR target domain by exploiting advanced data augmentations, explicit geometric constraints, disentangled feature normalization (DFN), and mean-teacher training.
	\item We show the generality of our approach as a novel SSL approach by using regular feature normalization instead of DFN.
	\item We extend two challenging OR datasets with person bounding box and 2D keypoint annotations for the evaluation.
	\item We achieve significantly better results against strongly constructed baselines on the high- and low-resolution OR images.
\end{enumerate}

\begin{figure*}[t!]
	%\hfill
	\centering
	\setlength{\fboxsep}{0pt}%
	\setlength{\fboxrule}{1.5pt}%	
	\begin{subfigure}[t]{0.24\textwidth}
		\centering
		\fbox{\includegraphics[width=1.6in]{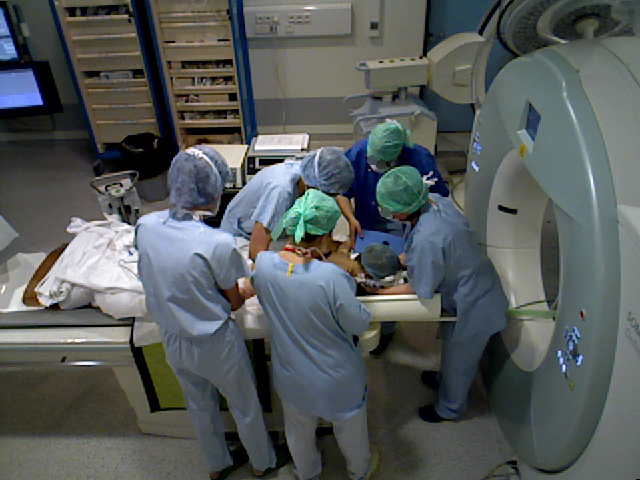}}
		\caption{640x480 (1x)}
	\end{subfigure}
	\begin{subfigure}[t]{0.24\textwidth}
		\centering
		\fbox{\includegraphics[width=1.6in]{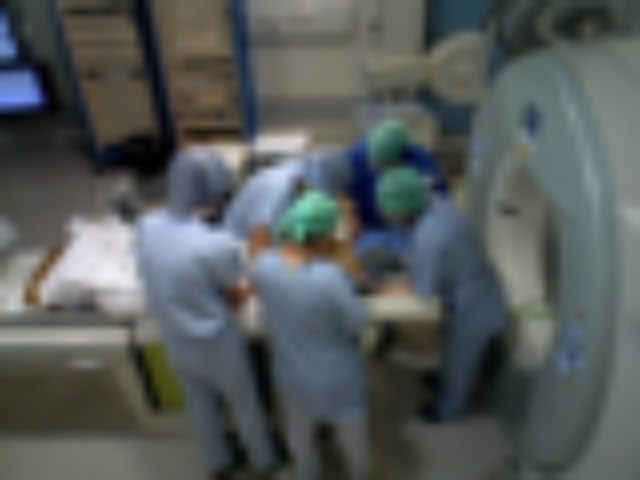}}
		\caption{80x60 (8x)}
	\end{subfigure}
	\begin{subfigure}[t]{0.24\textwidth}
		\centering
		\fbox{\includegraphics[width=1.6in]{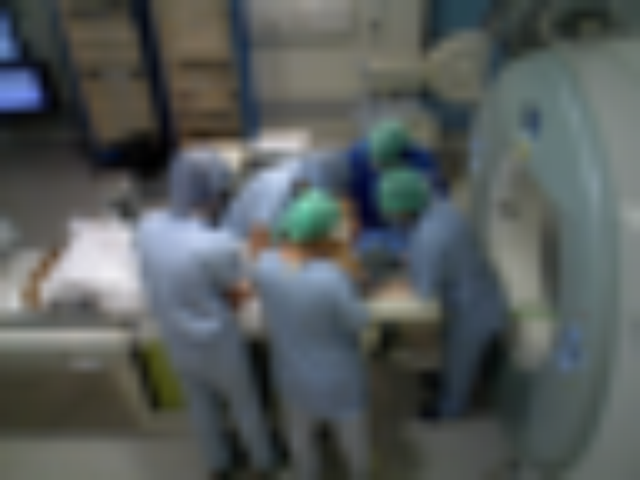}}
		\caption{64x48 (10x)}
	\end{subfigure}
	\begin{subfigure}[t]{0.24\textwidth}
		\centering
		\fbox{\includegraphics[width=1.6in]{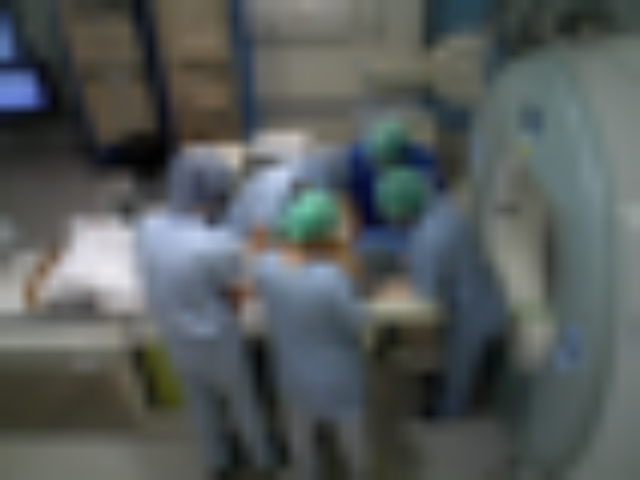}}
		\caption{53x40 (12x)}
	\end{subfigure}
	\caption{Sample image from the OR downsampled at different resolutions. The downsampled images contain little information to identify clinicians and the patients, making them more suitable for activity analysis in privacy-sensitive OR environments.}
	\label{figure:mvor-tum}
	\vspace{-4mm}
\end{figure*}
\section{Related works}
\subsection{Human pose estimation} 
Human pose estimation (HPE) has been mainly studied either using bottom-up (keypoint-first) or top-down (person-first) approaches. The bottom-up approaches first detect all the keypoints for all the persons and then use a group post-processing method to associate keypoints to person instances; conversely, the top-down approaches first obtain the bounding box for each person instance using an \emph{off-the-shelf} object detector and then employ a single-person pose estimation method to get the keypoints. The group post-processing methods in bottom-up approaches include Part Affinity Fields in CMU-Pose \citep{cao2016realtime}, Part Association Field in PifPaf \citep{kreiss2019pifpaf}, and Associative Embedding (AE) in \citep{newell2017associative,cheng2020higherhrnet}. The leading methods for single-person pose estimation in the top-down approaches include Simple-Baseline \citep{xiao2018simple}, Alpha-Pose \citep{fang2017rmpe}, Cascaded-Pyramid-Network \citep{Chen2018CPN}, HRNet \citep{SunXLW19}, and EvoPose2D \citep{mcnally2020evopose2d}. The bottom-up approaches are computationally faster due to their person-agnostic keypoint localization but yield inferior accuracy compared to the top-down approaches. The two-stage design in the top-down approaches helps them achieve significantly better accuracy but at a more computational cost. Built on top of anchor-free detector \citep{tian2019fcos}, some recent approaches such as DirectPose \citep{tian2019directpose} and FCPose \citep{mao2021fcpose} consider the keypoints as a special bounding-box with more than two corners and propose to regress the keypoint coordinates directly.

HPE in the OR is a relatively new field with approaches applied to either single or multi-view images and on color (RGB), depth (D), or both color and depth (RGB-D) images. The initial work \citep{kadkhodamohammadi2014temporally} propose a method to consistently track the upper body poses by offline optimization using discrete Markov Random Field (MRF) on the short RGB-D video sequences. The authors further propose an approach using the pictorial structure model \citep{fischler1973representation,felzenszwalb2005pictorial} initially designed for the RGB images to the RGB-D images with a handcrafted histogram of depth difference (HDD) features \citep{kadkhodamohammadi2015pictorial}. Subsequent work use the multi-view RGB images \citep{belagiannis2016parsing} and multi-view RGB-D images \citep{kadkhodamohammadi2017multi,kadkhodamohammadi2017-ar} for 3D HPE along with the corresponding multi-view RGB and multi-view RGB-D extensions to the pictorial structure model. Some recent work utilizes multi-view depth data for 3D HPE in the OR either using a voxel-based model \citep{hansen2019fusing} or point R-CNN model \citep{bekhtaoui2020view}. Previous work from the authors \citep{srivastav2019human,srivastav2020human} has also studied unsupervised domain adaptation for the OR. The authors in \citep{srivastav2019human} adapt the RT-Pose RGB model \citep{cao2016realtime} to the low-resolution OR depth images for 2D HPE, and the authors in \citep{srivastav2020human} adapt the Mask R-CNN model \citep{he2017mask} to the OR RGB images for joint 2D/3D HPE. Both these approaches use a two-stage approach. In the first stage, a complex multi-stage teacher model is used to generate pseudo labels on the target domain, and in the second stage, a student model is trained using these pseudo labels. Conversely, our approach uses a single-stage approach to generate and consume the pseudo labels \emph{on the fly} using the same given model as a teacher and a student. The visual shift of the source and the target domain data, in our single-stage design, is handled by improving the model with disentangled feature normalization.

\subsection{Instance segmentation} 
Instance segmentation has been extensively studied in the context of multi-class object detection. Like human pose estimation, the instance segmentation approaches can also be categorized into the bottom-up and top-down approaches. The top-down method also uses a two-stage design first to detect the bounding box and then either classify mask proposals or estimate segmentation masks from the bounding box proposals \citep{he2017mask,chen2019tensormask,bai2017deep,liu2017sgn,lee2020centermask}. Similarly, the bottom-up methods associate pixel-level semantic segmentation output to the object instance \citep{zhang2016instance,liang2017proposal,kirillov2017instancecut}. Inside the OR, the only related work \citep{li2020robotic} addresses a 3D scene semantic segmentation from multi-view depth images; however, the data is obtained from simulated clinical activities.

\subsection{Joint person pose estimation and instance segmentation}
A few notable works address the joint person pose estimation and instance segmentation \citep{papandreou2018personlab,he2017mask,zhang2019pose2seg,zhou2020poseg}. The authors in \citep{zhang2019pose2seg,zhou2020poseg} use pose estimation as a strong prior for the person instance segmentation. The PersonLab \citep{papandreou2018personlab} as a bottom-up method and Mask R-CNN \citep{he2017mask} as a top-down method are designed for the joint person pose estimation and instance segmentation.\\
We extend the top-down Mask R-CNN to build the backbone model in this work. Unlike the top-down approaches for HPE, which use separate networks for the two stages and do not share the features and computations, Mask R-CNN uses different heads for the end task that share common features across the heads, making it computationally faster and more easily configurable. This configurability property can be utilized to either use the model for a particular task - for example, only for instance segmentation or pose estimation - or extend it further - for example, for dense pose estimation \citep{guler2018densepose}.

\subsection{Privacy-preserving low-resolution image recognition}
{\blue The privacy-sensitive OR environment poses challenges in bringing the AI inside the OR. The recent controversies \citep{powles2017google} have raised public awareness regarding how personal data should be collected and controlled, along with how AI algorithms should use personal data in a privacy-safe way \citep{symons2017me}. One way to address these challenges is by using the federated learning \citep{mcmahan2017communication} framework that allows training the model in a decentralized manner without explicitly sharing data. The federated learning has been recently used in medical imaging for segmenting the brain tumor \citep{sheller2018multi} and detecting COVID-19 lung abnormalities in CT \citep{dou2021federated}. Unlike medical imaging data, where privacy-sensitive information essentially lies in the metadata, direct video recording of OR using ceiling cameras contains the private information in the data itself. Adapting a model to very low-resolution images has been suggested in the literature to improve privacy \citep{haque2018activity} that can further be incorporated inside the federated learning setup to improve multi-centric generalization. Indeed, as low-resolution images significantly degrade the spatial details, it could provide a viable means to improve privacy.}

The low-resolution image recognition has been studied for various computer vision tasks ranging from 2D human pose estimation on RGB \citep{neumann2018tiny} and depth \citep{srivastav2019human} images, face recognition \citep{ge2018low}, image classification \citep{wang2016studying}, image retrieval \citep{tan2018feature}, object detection \citep{haris2018task,li2017perceptual}, to activity recognition \citep{haque2018activity, ryoo2017privacy}. The low-resolution images as a means for privacy preservation have primarily been studied for the activity recognition in the hospital \citep{haque2018activity}, indoor posture recognition \citep{gochoo2020lownet}, and 2D human pose estimation on depth images inside the OR  \citep{srivastav2019human}. These approaches address the spatial degradation of the low-resolution image either at the image space \citep{haque2018activity} by using \emph{off-the-shelf} super-resolution model to enhance the spatial details of the low-resolution image or at the feature space \citep{srivastav2019human,tan2018feature} by directly optimizing the features suitable for the end task. Our approach falls into the latter, where we utilize advanced data augmentations to enforce consistency constraints between the high-and the low-resolution image derived from the pseudo labels, consequently enhancing the features for the low-resolution image. 

\subsection{Unsupervised domain adaptation}
Unsupervised domain adaptation (UDA) methods assume the availability of a labeled source domain and an unlabeled target domain sharing a common label space. The UDA approaches for the different end tasks can be broadly classified in two main areas: \emph{adversarial domain alignment} and \emph{self-training}.\\
The main idea in \emph{adversarial domain alignment} based UDA approaches is to update either the feature, input, or output space from the target domain such that they are distributed in the same way as the source domain. At the feature space, for example, the domain invariant feature space is achieved using an additional neural network called \emph{domain classifier} which essentially plays a min-max game with the \emph{feature extractor} using adversarial learning \citep{goodfellow2014generative}. I.e., the \emph{domain classifier} tries to fool the \emph{feature extractor} by accurately distinguishing the source and the target domain features using a binary classification loss on the domain labels; the \emph{feature extractor}, in turn, tries to fool the \emph{domain classifier} by producing domain invariant feature such that the \emph{domain classifier} would result in poor domain discrimination accuracy. The \emph{adversarial domain alignment} has been studied at the feature space in \citep{ben2010theory,hoffman2016fcns,chen2018domain,chen2019synergistic,du2019ssf,saito2019strong,tran2019gotta,hsu2020every,sindagi2020prior,vs2021mega}, at the input space in \citep{zhu2017unpaired,chen2019learning,chen2019crdoco,choi2019self,li2019bidirectional}, and at the output space in \citep{tsai2018learning,luo2019taking,tsai2019domain,kim2020learning}. Although these methods have made significant progress, stable training in the adversarial setup requires complicated training routines with careful adjustment to training parameters. Moreover, aligning the two domains using the \emph{domain classifier} may not guarantee a required discriminative capability for a given end task.

The \emph{self-training} based UDA methods have emerged as promising alternatives to \emph{adversarial domain alignment} as they follow a simple approach to learn the domain invariant representations. The main idea in the \emph{self-training} is to generate pseudo labels on the unlabeled target domain by refining the predictions - generated from a given source domain trained model - using domain/task-specific heuristics, for example, confidence score in object detection \citep{deng2021unbiased} or uncertainty in semantic segmentation \citep{liang2019exploring, zheng2021rectifying}. These pseudo labels are then used to train a model on the target domain jointly with the labeled source domain. The \emph{self-training} has been extensively studied for object detection and semantic segmentation tasks \citep{inoue2018cross,zou2018unsupervised,roychowdhury2019automatic,khodabandeh2019robust,kim2019self,zou2019confidence,zhao2020collaborative,wang2020unsupervised,zheng2021rectifying}. The \emph{self-training} methods could further be improved in a \emph{mean-teacher} framework to tackle noise in the pseudo labels \citep{cai2019exploring,liang2019exploring}. The \emph{mean-teacher} and the \emph{self-training} based UDA approaches have predominantly been inspired by the advances in the SSL \citep{tarvainen2017mean,berthelot2019mixmatch,sohn2020fixmatch,liu2021unbiased}. In fact, the UDA can be posed inside an SSL framework with the source domain data as the labeled and the target domain data as unlabeled along with additional complexity of the visual shift of the two domains.

Some recent works aim to learn domain-specific feature representation instead of domain invariant using disentangled feature normalization. These approaches modify the feature normalization layers - as these control the feature distribution statistics - with two separate layers to disentangle the features from the two domains \citep{chang2019domain}. The domain-specific features learning has been studied in the UDA for image classification \citep{chang2019domain,wang2019transferable}, and federated learning on medical imaging \citep{li2021fedbn}. It has also been used to boost performance in the supervised learning \citep{xie2020adversarial}, and adversarial robustness \citep{xie2019intriguing}. The authors in \citep{wu2021rethinking} comprehensively discuss the feature normalization under various visual recognition tasks. There also exist several survey papers that extensively discuss UDA for the end task of image classification \citep{patel2015visual,wang2018deep,zhuang2020comprehensive}, semantic segmentation \citep{toldo2020unsupervised,zhao2020review}, and object detection \citep{oza2021unsupervised}.

A few notable works propose to use the UDA on the medical domain for cross-domain segmentation task \citep{li2020domain,ouyang2019data,orbes2019knowledge,chen2019synergistic}, and image classification \citep{zhang2020collaborative}. The authors in \citep{dong2020can} also study the UDA to identify domain invariant transferable features for endoscopic lesions segmentation. The authors in \citep{dipietro2019automated} study the surgical workflow recognition with as few as one labeled sequence. \citep{li2020transformation,zheng2020annotation} papers on medical imaging.

\vspace{4mm}
Our UDA approach for the joint person pose estimation and instance segmentation builds on the notable contributions from the related domains. We propose disentangled feature normalization that uses separate normalization layers in the feature extractor and modifies the multi-task loss function for our joint person pose estimation and instance segmentation model. We use a generic \emph{self-training} framework along with extended data augmentation pipeline and \emph{mean-teacher} training to add explicit geometric constraints on the different augmentations of input images, thereby generating accurate pseudo labels that are especially useful to adapt the model to the low-resolution OR images. We show the effectiveness of our approach on the two OR datasets at varying downsampling scales. We further demonstrate the generality of our approach as a novel SSL method on the large-scale COCO dataset.

\section{Detailed methodology}
\begin{figure*}[tb]
	\includegraphics[clip, trim=0.0cm 0.0cm 2.0cm 0.0cm, width=1.0\linewidth]{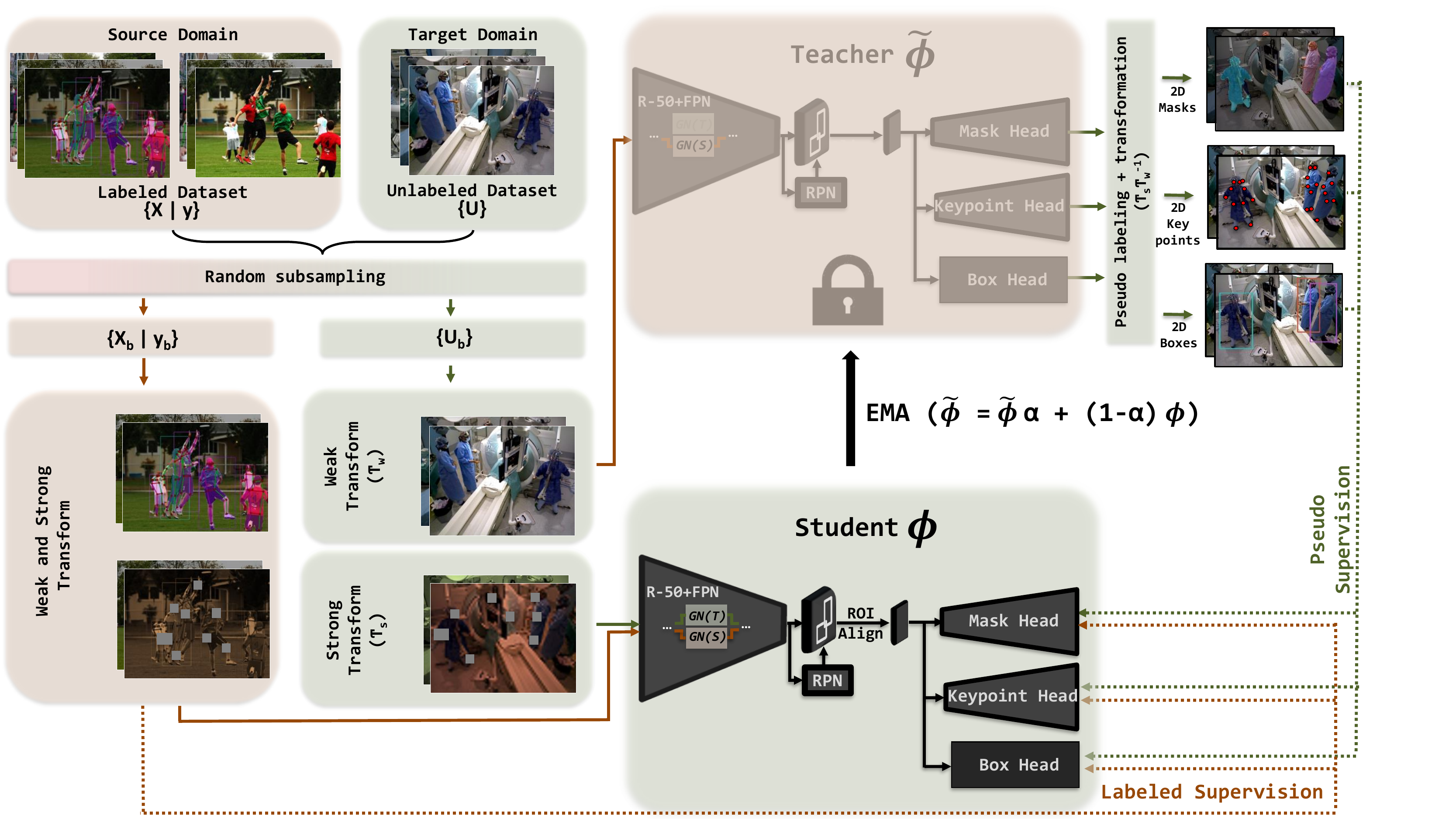}
	\caption{Overview of our approach for unsupervised domain adaptation. We generate two types of augmentations on the given unlabeled target domain images: weak and strong. The weakly augmented images pass through a frozen teacher model and a thresholding function to generate the pseudo labels. These pseudo labels are then geometrically transformed into the strongly augmented image space. A student model uses these transformed pseudo labels to train on the strongly augmented unlabeled images jointly with the labeled source domain images. The weights of the frozen teacher model are updated using the exponential moving average (EMA) of the student model's weights. We also replace every group normalization (GN) layer in the feature extractor with two GN layers (\emph{GN(S)} and \emph{GN(T)}) to normalize features of two domains separately, as needed to handle statistically different source and target domains.}
	\label{fig:architecture}
\end{figure*}
\subsection{Problem overview}
Given an end-to-end model for joint person pose estimation and instance segmentation trained on the source domain labeled dataset $\mathcal{X}= \left\{{x_i} | {y_i} \right\}_{i=1}^{N_l} $, we aim to adapt it to the unlabeled target domain dataset $ \mathcal{U} = \left\{{u_j}\right\}_{i=1}^{N_u} $. The source domain images are natural \emph{in the wild} images, whereas the target domain images are the high-resolution and low-resolution (downsampled up to 12x) images from the OR. $N_l$ and $N_u$ are the number of labeled and unlabeled images, respectively. The source domain's labeled dataset consists of images $x_i$ with the corresponding ground-truth labels $y_i$. The ground-truth labels $y_i$ consist of bounding boxes $\mathcal{P}_{bbox} \in \mathbb{R}^{m \times 4}$, keypoints $\mathcal{P}_{kp} \in \mathbb{R}^{m \times n \times 2}$, and masks $\mathcal{P}_{mask} \in \mathbb{R}^{m \times p \times 2}$, where $m$ is the number of persons, $n$ is the number of 2D keypoints for each pose, and $p$ is the number of contour points on the ground-truth binary mask. The unlabeled data from the target domain consists of only the images $u_j$.

We first explain the backbone models chosen for this work and the proposed UDA method, which we call \emph{AdaptOR}. Briefly, we first extend Mask R-CNN \citep{he2017mask} with disentangled feature normalization (DFN) to handle the statistically different datasets from the two domains. Then we develop our approach by designing geometrically constrained data augmentations to generate and use the pseudo labels for adapting the model to the unlabeled target domain consisting of high- and low-resolution images from the OR.

\subsection{Backbone models}
We choose the Mask R-CNN \citep{he2017mask} model, where the mask and the keypoint head are designed to use a single person class. We refer to this model as \emph{km-rcnn} tailored to joint person pose estimation and instance segmentation. It can also perform person bounding box detection by design. \emph{km-rcnn} works as follows: it first extracts the image features using a feature pyramid network (FPN) \citep{lin2017feature} with a Resnet-50 backbone \citep{he2016deep}. The extracted features pass through a region proposal network (RPN) to generate the bounding-box proposals. The \emph{RoiAlign} layer \citep{he2017mask} uses these proposals to extract the fixed-size feature maps. The fixed-size feature maps pass through three heads: bounding box head, keypoint head, and mask head. The bounding box head classifies and regresses for the person bounding box, the keypoint head generates the spatial heat-maps corresponding to each body keypoint, and the mask head generates segmentation masks. We use the same multi-task losses as described in \citep{he2017mask} except for bounding box classification loss where we use focal loss \citep{ross2017focal} instead of cross-entropy loss for the better handling of foreground-background class imbalance in our UDA framework \citep{liu2021unbiased}. Overall, the supervised loss term $\mathcal{L}_s$  consists of six losses: binary cross-entropy loss for RPN proposal classification $\mathbb{L}_{cls}^{rpn}$, L1 loss for RPN proposal regression $\mathbb{L}_{reg}^{rpn}$, focal loss \citep{ross2017focal} for bounding box classification $\mathbb{L}_{cls}^{bbox}$, smooth L1 loss for bounding box regression $\mathbb{L}_{reg}^{bbox}$, cross-entropy loss for the keypoint head $\mathbb{L}_{ce}^{kps}$, and the binary cross-entropy for the mask head $\mathbb{L}_{bce}$.
{\blue
\begin{dmath} \label{loss-equation_1}
	\mathcal{L}_s = \sum_{i}  \mathbb{L}_{cls}^{rpn}(f_i^l, y_i^l) + \mathbb{L}_{reg}^{rpn}(f_i^l, y_i^l) + \mathbb{L}_{cls}^{bbox}(f_i^l, y_i^l) + \mathbb{L}_{reg}^{bbox}(f_i^l, f_i^l) + \mathbb{L}_{ce}^{kp}(f_i^l, y_i^l) + \mathbb{L}_{bce}^{mask}(f_i^l, y_i^l).
\end{dmath}
Here, $f_i^l$ and $y_i^l$ correspond to the features and the ground-truth labels for the labeled input image $x_i^l$.
}
\subsubsection{Initialization}
The state-of-the-art approaches for downstream tasks such as object detection \citep{ren2015faster} and instance segmentation \citep{he2017mask} initialize the backbone network from the supervised ImageNet \citep{deng2009imagenet} weights. The feature normalization during the training is performed using frozen batch normalization (BN) in all the feature extraction layers. It, in turn, uses statistics (mean and variance) derived from the ImageNet training set and freezes its affine parameters (weights and biases).

The current advancements in self-supervised methods to learn generic feature representations exploiting large-scale unlabeled data have started to surpass the supervised ImageNet baselines on the downstream tasks \citep{chen2020simple,he2020momentum,misra2020self}. However, the backbone feature extractor weights from the self-supervised methods may not have the same distribution as supervised ImageNet methods. The use of frozen BN during the training therefore could lead to unstable training. Authors in \citep{he2020momentum} suggest training the BN layers using Cross-GPU BN \citep{peng2018megdet} to circumvent the issue. We find in our experiments that group normalization (GN) \citep{wu2018group} works equally well without the overhead of communicating the batch statistics over all the GPUs resulting in an increased training speed. We follow the network design from \citep{wu2018group,wu2019detectron2gn} to change the BN layers of \emph{km-rcnn} with the GN layers. The updated model, called \emph{km-rcnn+}, is initialized from the self-supervised method MoCo-v2 \citep{chen2020simple,he2020momentum} and trained on the source domain dataset.

\subsubsection{Disentangled feature normalization}\label{distangl}
Given the model, \emph{km-rcnn+}, trained on the labeled source domain dataset, we aim to adapt it to the unlabeled target domain. We observe in our experiments that feature normalization plays a vital role in training the model on different domains as suggested in the literature \citep{xie2020adversarial,chang2019domain,wu2021rethinking}. We propose disentangled feature normalization (DFN) to effectively disentangle the feature learning for the datasets of different domains by replacing every group normalization (GN) layer in the feature extractor with two GN layers: one for the source domain images, \emph{GN(S)}, and another for the target domain images, \emph{GN(T)}. The updated model, called \emph{km-rcnn++}, uses separate affine parameters at every normalization stage in the feature extractor for the source and the target domain images, efficiently normalizing the features of the two domains, see figure \ref{fig:architecture}. The GN parameters for the target domain, \emph{GN(T)}, are initialized from the source domain GN parameters, \emph{GN(S)}, before the domain adaptation training.

The UDA approaches require weighing the losses differently for unlabeled and labeled images, usually to weigh more the unlabeled losses than the labeled ones to overcome the over-fitting to the labeled set. It can be easily performed if the underlying model is the same for the two domains: the usual case of the existing UDA approaches. However, with our improved design, the \emph{km-rcnn++} model expects an input batch containing the first half of images from the source domain and the second half of images from the target domain. DFN therefore modifies the loss function described in equation \ref{loss-equation_1} to compute and weigh the losses on the source and the target domain images differently. The input batch passes through the feature extractor, and the obtained features are divided into two halves corresponding to the source and the target domains. Each half then passes through the RPN network and the three heads to compute the separate RPN, bounding box, keypoint, and mask losses for source and the target domain images as given below.
{\blue
\begin{dmath} \label{loss-equation_2}
	\mathcal{L}_s = \sum_{i}  \mathbb{L}_{cls}^{rpn}(f_i^l, y_i^l) + \mathbb{L}_{reg}^{rpn}(f_i^l, y_i^l) + \mathbb{L}_{cls}^{bbox}(f_i^l, y_i^l) + \mathbb{L}_{reg}^{bbox}(f_i^l, f_i^l) + \mathbb{L}_{ce}^{kp}(f_i^l, y_i^l) + \mathbb{L}_{bce}^{mask}(f_i^l, y_i^l)
\end{dmath}
\begin{dmath} \label{loss-equation_3}
	\mathcal{L}_u = \sum_{i}  \mathbb{L}_{cls}^{rpn}(f_i^u, y_i^u) + \mathbb{L}_{reg}^{rpn}(f_i^u, y_i^u) + \mathbb{L}_{cls}^{bbox}(f_i^u, y_i^u) + \mathbb{L}_{reg}^{bbox}(f_i^u, f_i^u) + \mathbb{L}_{ce}^{kp}(f_i^u, y_i^u) + \mathbb{L}_{bce}^{mask}(f_i^u, y_i^u).
\end{dmath}
Here, $f_i^l$ and $f_i^u$ correspond to the features of the labeled and unlabeled domain, respectively. The $y_i^l$ corresponds to the source domain labeled ground-truth labels, and $y_i^u$ correspond to the target domain pseudo labels. The following section explains the automatic generation of target domain pseudo labels  $y_i^u$.} The \emph{km-rcnn++} in the inference mode uses only the GN layers corresponding to the target domain, thereby maintaining the same number of parameters and inference cost compared to \emph{km-rcnn+}. 
% Fig.~2 illustrates the mean and variance distributions on the intermediate feature maps after applying the GN layer on some sample source domain and the target domain images. We observe that the affine parameters of the GN layer are tuned to the source domain images and do not normalize the target domain images appropriately.
% \begin{dmath}\label{loss-equation_2}
% \mathcal{L}=\mathcal{L}_s,\mathcal{L}_u=\sum_{i \in source}L,\sum_{i \in target}L
% \end{dmath}
\begin{algorithm}[ht!]
	\caption{: \emph{AdaptOR} algorithm to adapt a model trained on the labeled source domain dataset to the unlabeled target domain (operating room)}\label{algo:adaptor}
	\textbf{Inputs:}
	\begin{itemize}
		\item Labeled dataset from the source domain $\mathcal{X}= \left\{{x_i}| {y_i} \right\}_{i=1}^{N_l} $, unlabeled dataset from target domain $\mathcal{U} = \left\{{u_j}\right\}_{i=1}^{N_u}$, $y_i$ = ($\mathcal{P}_{bbox},\mathcal{P}_{kp},\mathcal{P}_{mask})$: ground-truth labels for the bounding box, keypoints, and mask for each person in the given labeled image.
		\item $p_{t}(y|x;{\tilde{\phi}})$: teacher model, $p_{s}(y|x;{\phi})$: student model, ${\tilde{\phi}}$, ${\phi}$: weights of the teacher and the student model respectively
		\item $\varGamma(\mathit{p}, \delta = \delta_{bbox},\delta_{kp},\delta_{mask})$: function to convert predictions ($\mathit{p}$) to pseudo labels using thresholds ($\delta$) consisting of bounding box threshold ${\delta}_{bbox}$, keypoint threshold ${\delta}_{kp}$, and mask threshold ${\delta}_{mask}$
		\item $\mathcal{T}_w(.)$: weak transform, $\mathcal{T}_s(.)$: strong transform
		\item $\mathcal{L}$: modified multi-task loss function as described in section \ref{distangl} and equations \ref{loss-equation_2} and \ref{loss-equation_3}, $\alpha$: EMA decay rate, ${\lambda}$: unsupervised weight loss value, $\eta$: learning rate
	\end{itemize}
	\textbf{Outputs:} ${\tilde{\phi}}$: Final teacher model weights
	\begin{algorithmic}[1]
		\ForAll{$(\mathcal{X}_b, y_b, \mathcal{U}_b)\in (\mathcal{X}, \mathcal{U})$} \Comment{sample a batch from the labeled and unlabeled dataset}
		\State $\mathcal{X}_{w}, y_{w}, \mathcal{U}_{w} = \mathcal{T}_w(\mathcal{X}_b, y_b, \mathcal{U}_b)$ \Comment{apply weak transform to the labeled and unlabeled batch to construct weakly augmented labeled ($\mathcal{X}_{w}, y_{w}$) and unlabeled ($\mathcal{U}_{w}$) batch}
		\State $\mathcal{X}_{s}, y_{s}, \mathcal{U}_{s} = \mathcal{T}_s(\mathcal{X}_b, y_b, \mathcal{U}_b)$ \Comment{apply strong transform to the labeled and unlabeled batch to construct strongly augmented labeled ($\mathcal{X}_{s}, y_{s}$) and unlabeled ($\mathcal{U}_{s}$) batch}
		\State $\tilde{y_{s}}  = \varGamma(p_{t}(\mathcal{U}_w; \tilde{\phi}), \delta)$ \Comment{run the teacher model $p_{t}(y | x; {\tilde{\phi}})$ on the weakly augmented unlabeled batch $\mathcal{U}_{w}$, and convert the predictions into the pseudo labels $\tilde{y_{s}}$ using the thresholding function $\varGamma (\mathit{p},\delta)$}
		\State $\bar{y_{s}}  = \mathcal{T}_{s}(\mathcal{T}_{w}^{-1}(\tilde{y_{s}}))$ \Comment{apply the transform to convert the pseudo labels $\tilde{y_{s}}$ into the coordinates of strongly augmented unlabeled batch ($\mathcal{U}_{s}$)}
		\State $ \mathcal{X}, y = concat(\mathcal{X}_{w},\mathcal{X}_{s},\mathcal{U}_{s}), concat(y_{w},y_{s},\bar{y_{s}})$
		\Comment{concatenate the strongly augmented unlabeled batch with the weakly and strongly augmented labeled batch}
		\State $\mathcal{L}_s, \mathcal{L}_u = \mathcal{L}(p_{s}(\mathcal{X}; \phi), y) $ \Comment{compute the loss using the multi-task loss function on the student model}
		\State $loss =  \mathcal{L}_s + \lambda\mathcal{L}_u  $ \Comment{add the supervised and the unsupervised losses}
		\State ${\phi} = SGD(\phi, \eta, {\nabla}_\phi(loss))$ \Comment{update the parameters of the student model $\phi$ using stochastic gradient descent with momentum}
		\State $\tilde{\phi} = \alpha \tilde{\phi} + (1 - \alpha) \phi$ \Comment{update the parameters of teacher model $\tilde{\phi}$ using the exponential moving average}
		\EndFor
		\State \textbf{end for}
	\end{algorithmic}
\end{algorithm}
\subsection{AdaptOR} \label{algo-section}
Given a model, \emph{km-rcnn++}, that can handle the datasets of different domains, we explain \emph{AdaptOR}, our proposed method for unsupervised domain adaptation. We first explain \emph{transformation equivariance constraints}, as needed to add explicit geometric constraints, and then the data augmentation pipeline, followed by the complete algorithm.
\subsubsection{Transformation equivariance constraints}\label{trans-eqv} The state-of-the-art UDA or SSL approaches for image classification exploit the \emph{transformation invariance} property on the unlabeled data, i.e., the classification labels remain unchanged irrespective of the transformation applied to the input image. However, the \emph{invariance} property does not hold for the spatial localization tasks, and labels get changed with the viewpoint changes of the image due to geometric transforms, for example, resize and horizontal flip. But, these changes in the labels are \emph{equivariant} to the applied transformations. Mathematically, if $\mathcal{F}(.)$ is a model that outputs the spatial localization labels for the input image $I$ under transformation $\mathcal{T}$, we can minimize $ \lVert \mathcal{F}(\mathcal{T}(I)) - \mathcal{T}(\mathcal{F}(I))\rVert$ under \emph{transformation equivariance constraints}, i.e., the transformation $\mathcal{T}$ can be to used map the localization labels to the transformed image space. We use this property to provide the explicit geometric constraints on the unlabeled images. Additionally, specific to the human pose estimation under horizontal flipping transformation, we exploit the chirality transform \citep{yeh2019chirality} for the mapping of the human pose to the horizontally flipped image.

\subsubsection{Data augmentations} \label{aug}
Data Augmentations construct novel and realistic samples by computing stochastic transforms on the input data. The recent advancements in data augmentations have been the key to the performance boost in the supervised as well as SSL approaches \citep{cubuk2019autoaugment, cubuk2020randaugment, devries2017improved}. We use two types of augmentations: \emph{weak} and \emph{strong}. The \emph{weak} augmentations, $\mathcal{T}_w$, consist of random-flip and random-resize whereas \emph{strong} augmentations, $\mathcal{T}_s$, consist of spatial augmentations from rand-augment \citep{cubuk2020randaugment}, random cut-out \citep{devries2017improved}, random-flip, and random-resize, along with \emph{strong-resize} augmentation to generate privacy-preserving low-resolution images. The \emph{strong-resize} data augmentation down-sample and up-sample the input image with a random scaling factor chosen between 1x to 12x. Fig.~\ref{figure:mvor-tum} shows sample images from the OR at different downsampling scales.

\subsubsection{Algorithm}
Given the \emph{weakly} augmented image, constructed using transformation $\mathcal{T}_w$, and the \emph{strongly} augmented image, constructed using the transformation $\mathcal{T}_s$, our idea is to geometrically transform the pseudo labels - obtained from the model's predictions - of the \emph{weakly} augmented image to the corresponding \emph{strongly} augmented image. As the \emph{weakly} and the \emph{strongly} augmented images are generated using different geometric transformations with the pseudo labels being in the \emph{weakly} augmented image coordinate system, we exploit \emph{transformation equivariance constraints} to transform the pseudo labels by applying a transformation, $\mathcal{T}_s\mathcal{T}_w^{-1}$, to go from \emph{weakly} augmented image space to the \emph{strongly} augmented image space. The model is trained on the \emph{strongly} augmented images with the transformed pseudo labels.

However, training the same model to generate and consume the pseudo labels may lead to unstable training. The \emph{mean-teacher} \citep{tarvainen2017mean} from semi-supervised learning has been proposed to stabilize the training using closely coupled \emph{teacher} and a \emph{student} model. We therefore adapt \emph{mean-teacher} in our approach, where we use the \emph{teacher} model to generate the pseudo labels on the \emph{weakly} augmented image, and the \emph{student} model to train on the corresponding \emph{strongly} augmented image using the pseudo labels. As the source domain GN parameters, \emph{GN(S)}, are trained under the direct supervision, we use \emph{GN(S)} layers in the \emph{teacher} model for the inference on the unlabeled target domain. The weights of the \emph{teacher} and the \emph{student} models are initialized from the same model, \emph{kmrcnn++}. The weights of the \emph{student} model are updated using the stochastic gradient descent based back-propagation, whereas the weights of the \emph{teacher} model are updated using the exponential moving average (EMA) of the weights of the \emph{student} model:
$$\tilde{\phi} = \alpha \tilde{\phi} + (1 - \alpha) \phi ,$$ where $\tilde{\phi}$ and $\phi$ are the weights of the teacher model and the student models, respectively, and $\alpha$ is a decay parameter. The EMA helps the \emph{teacher} model to generate better predictions due to its temporal ensembled weights from the \emph{student} model, in turn improving the \emph{student} model for better training. The detailed algorithm is explained in algorithm \ref{algo:adaptor} and illustrated in figure \ref{fig:architecture}.

Furthermore, we also test \emph{AdaptOR} as an SSL approach, called \emph{AdaptOR-SSL}, on a source domain dataset by making minimal changes. We use the \emph{kmrcnn+} model, without disentangled feature normalization, as the images are coming from the same domain and do not concatenate the labeled and the unlabeled batches. The labeled and the unlabeled batches pass separately through the \emph{kmrcnn+} model to calculate the separate losses on the labeled and the unlabeled data. The \emph{AdaptOR-SSL} uses x\%(x=1,2,5,10) of images from the source domain as the labeled dataset and the rest of the images as the unlabeled dataset.

\section{Baselines}
We first introduce several \emph{self-training} based baselines that we have constructed for our joint person pose estimation and instance segmentation task by extending representative approaches. We extend pseudo-label \citep{lee2013pseudo, sohn2020simple}, data-distillation \citep{radosavovic2018data}, and ORPose \citep{srivastav2020human} as our baselines approaches. We refer to the extended version of pseudo-label, data-distillation, and ORPose as KM-PL, KM-DDS, and KM-ORPose, respectively. The KM as a prefix signifies that these approaches have been extended for the joint pose (keypoint) estimation and instance (mask) segmentation tasks. The baselines approaches are two-stage approaches where the first stage generates the pseudo labels on the unlabeled data. The second stage jointly trains the model using the pseudo and the ground truth labels. \emph{AdaptOR} on the other hand generates the pseudo labels on the unlabeled data \emph{on-the-fly} during the training. For a fair comparison, we train all the baseline methods with the same training strategy, data augmentation pipeline, and \emph{kmrcnn++} model. We give a brief overview of extended baseline approaches as follows.

\begin{table}[t!]
	\centering
	\caption{\small{{{\blue An overview of the source and the target domain datasets used in this work.}}}}
	\vspace{-2mm}
	\scalebox{0.85}{
		\begin{tabular}{l|c|c|c}
			\toprule
			\textbf{Dataset}                           & \textbf{type} & \textbf{\# images} & \textbf{\# instances} \\
			\textit{Source domain labeled dataset}     &               &                                            \\
			COCO                                       & train         & 57,000             & 150,000               \\
			\textit{COCO-val}                          & test          & 5,000              & 10,777                \\
			\hline\bottomrule
			\textit{Target domain unlabelled datasets} &               &                                            \\
			MVOR                                       & train         & 80,000             & -                     \\
			\textit{MVOR+}                             & test          & 2,196              & 5,091                 \\\hline
			TUM-OR                                     & train         & 1,500              & -                     \\
			\textit{TUM-OR-test}                       & test          & 2,400              & 11,611                \\
			\hline\bottomrule
		\end{tabular}
	}
	\label{table:datasets}
	\vspace{-1mm}
\end{table}

\begin{table*}[t]
	\centering
	\caption{\small{Results on the source domain \emph{COCO-val} dataset with 100\% labeled supervision. The \emph{kmrcnn+} model using GN \citep{wu2018group} and initialized using self-supervised MoCo-v2 approach \citep{chen2020simple,he2020momentum} perform equally well with the model using Cross-GPU BN \citep{peng2018megdet} but using less training time. The first row result for the \emph{kmrcnn} model is obtained from the paper \citep{he2017mask}. Rest of the results correspond to the models that we train. Inference is performed on a single-scale of 800 pixels following \citep{he2017mask}. Automatic mixed precision (AMP) uses single- and half-precision (32 bits and 16 bits) floating operation to speed up the training while trying to maintain single-precision (32 bits) model accuracy.}}
	\vspace{-2mm}
	\scalebox{1.0}{
		\begin{tabular}{l|c|c|c|c|cccc}
			\toprule
			\textbf{{Model}} & \textbf{Initialization} & \textbf{Normalization} & \textbf{AMP} & \textbf{$\approx$ Training-time} & $\mathit{AP_{person}^{bb}}$ & $\mathit{AP_{person}^{kp}}$ & $\mathit{AP_{person}^{mask}}$ \\
			\toprule
			\emph{kmrcnn}    & Supervised-Imagenet     & Frozen BN              & \xmark       & 32 hours                         & 52.0                        & 64.7                        & 45.1                          \\
			\emph{kmrcnn}    & Supervised-Imagenet     & Frozen BN              & \checkmark   & 16 hours                         & 56.4                        & 65.7                        & 49.1                          \\
			\emph{kmrcnn}    & MoCo-v2                 & Cross-GPU BN           & \checkmark   & 22 hours                         & 57.5                        & 66.6                        & 49.8                          \\
			\emph{kmrcnn+}   & MoCo-v2                 & GN                     & \checkmark   & 18 hours                         & 57.5                        & 66.2                        & 49.9                          \\
			\hline\bottomrule
		\end{tabular}
	}
	\label{table:source}
	\vspace{-1mm}
\end{table*}

\begin{table*}[t]
	\centering
	\caption{\small{Results for the baseline approaches and \emph{AdaptOR}. We see improvements in all three metrics on both the target domain datasets, especially on the low-resolution images making the proposed approach suitable for the deployment inside the privacy-sensitive OR environment. The \emph{source-only} results correspond to the model trained on the labeled source domain without any training on the target domain images. The KM-PL, KM-DDS, and KM-ORPose are strong baselines proposed in this work.}}
	\vspace{-2mm}
	\scalebox{0.92}{
		\begin{tabular}{l|cccc|cccc}
			\toprule
			\multirow{3}{*}{Methods} & \multicolumn{4}{c|}{\textbf{\emph{MVOR+}}} \Tstrut \Bstrut                                            & \multicolumn{4}{c}{\textbf{\emph{TUM-OR-test}}} \Tstrut \Bstrut                                                                                                                                                                     \\
			\cline{2-9}
			                         & \textbf{~1x~}                                                                                         & \textbf{~8x~}                                                   & \textbf{~10x~}          & \textbf{~12x~}          & \textbf{~1x~}           & \textbf{~8x~}           & \textbf{~10x~}          & \textbf{~12x~}   \Tstrut        \\
			\cline{2-9}
			                         & \multicolumn{8}{c}{$\mathit{AP_{person}^{bb}}$ (mean$\pm$std)}   \Tstrut \Bstrut                                                                                                                                                                                                                                                   \\
			\hline
			\emph{source-only}       & 56.61$\pm$0.34                                                                                        & 40.42$\pm$2.17                                                  & 34.87$\pm$2.47          & 29.61$\pm$2.69          & 68.61$\pm$1.54          & 41.84$\pm$2.33          & 31.08$\pm$2.83          & 24.00$\pm$2.90 \Tstrut          \\\hline
			KM-PL                    & 60.21$\pm$0.51                                                                                        & 57.14$\pm$0.34                                                  & 55.88$\pm$0.39          & 54.26$\pm$0.41          & 72.28$\pm$1.51          & 65.44$\pm$1.45          & 62.84$\pm$1.02          & 62.42$\pm$1.55 \Tstrut          \\
			KM-DDS                   & 60.79$\pm$0.47                                                                                        & 57.88$\pm$0.39                                                  & 56.74$\pm$0.37          & 55.12$\pm$0.45          & 72.51$\pm$1.45          & 65.98$\pm$1.18          & 63.87$\pm$0.99          & 62.68$\pm$1.32 \Tstrut          \\
			KM-ORPose                & 58.88$\pm$0.69                                                                                        & 55.14$\pm$0.56                                                  & 53.81$\pm$0.52          & 51.96$\pm$0.47          & 69.73$\pm$1.22          & 63.46$\pm$0.93          & 60.71$\pm$0.73          & 60.14$\pm$0.94 \Tstrut          \\ \hline
			\emph{\textbf{AdaptOR}}  & \textbf{61.41$\pm$0.40}                                                                               & \textbf{59.48$\pm$0.35}                                         & \textbf{58.55$\pm$0.36} & \textbf{57.33$\pm$0.43} & \textbf{72.75$\pm$0.88} & \textbf{67.33$\pm$0.78} & \textbf{65.53$\pm$0.57} & \textbf{65.65$\pm$0.66} \Tstrut
			\\\hline
			                         & \multicolumn{8}{c}{$\mathit{AP_{person}^{kp}}$ (mean$\pm$std)} \Tstrut \Bstrut                                                                                                                                                                                                                                                     \\
			\hline
			\emph{source-only}       & 50.55$\pm$0.39                                                                                        & 23.99$\pm$2.25                                                  & 16.86$\pm$2.16          & 11.31$\pm$1.91          & 65.60$\pm$4.55          & 27.21$\pm$1.49          & 19.41$\pm$1.86          & 13.18$\pm$1.81 \Tstrut          \\\hline
			KM-PL                    & 58.72$\pm$0.44                                                                                        & 55.19$\pm$0.43                                                  & 52.81$\pm$0.55          & 49.53$\pm$0.46          & 77.49$\pm$1.87          & 67.57$\pm$1.03          & 63.46$\pm$0.89          & 58.24$\pm$1.05 \Tstrut          \\
			KM-DDS                   & 59.83$\pm$0.40                                                                                        & 55.60$\pm$0.49                                                  & 53.16$\pm$0.48          & 50.02$\pm$0.46          & 78.39$\pm$1.76          & 69.24$\pm$1.07          & 65.29$\pm$0.93          & 60.56$\pm$1.21 \Tstrut          \\
			KM-ORPose                & \textbf{62.50$\pm$0.53}                                                                               & 57.18$\pm$0.60                                                  & 54.59$\pm$0.59          & 51.24$\pm$0.47          & \textbf{80.49$\pm$1.74} & 69.90$\pm$1.03          & 65.64$\pm$0.94          & 60.67$\pm$0.73 \Tstrut          \\ \hline
			\emph{\textbf{AdaptOR}}  & 60.86$\pm$0.38                                                                                        & \textbf{57.35$\pm$0.61}                                         & \textbf{55.42$\pm$0.66} & \textbf{52.60$\pm$0.60} & 77.84$\pm$1.24          & \textbf{70.65$\pm$1.04} & \textbf{67.36$\pm$0.96} & \textbf{63.27$\pm$1.21} \Tstrut \\
			\hline
			                         & \multicolumn{8}{c}{$\mathit{AP_{person}^{bb\ (from\ mask)}}$ (mean$\pm$std)} \Tstrut \Bstrut                                                                                                                                                                                                                                       \\
			\hline
			\emph{source-only}       & 54.95$\pm$0.37                                                                                        & 37.98$\pm$2.21                                                  & 32.58$\pm$2.37          & 27.56$\pm$2.48          & 69.33$\pm$1.46          & 40.38$\pm$2.30          & 30.11$\pm$2.79          & 22.97$\pm$2.93  \Tstrut         \\\hline
			KM-PL                    & 56.50$\pm$0.60                                                                                        & 54.06$\pm$0.44                                                  & 52.90$\pm$0.48          & 51.33$\pm$0.46          & 71.93$\pm$1.34          & 65.43$\pm$1.46          & 63.16$\pm$0.89          & 62.67$\pm$1.11 \Tstrut          \\
			KM-DDS                   & 57.12$\pm$0.47                                                                                        & 54.76$\pm$0.50                                                  & 53.78$\pm$0.49          & 52.06$\pm$0.67          & 71.99$\pm$1.18          & 65.96$\pm$1.07          & 64.02$\pm$0.70          & 63.01$\pm$1.02 \Tstrut          \\
			KM-ORPose                & 55.46$\pm$0.76                                                                                        & 52.37$\pm$0.62                                                  & 51.23$\pm$0.55          & 49.34$\pm$0.46          & 68.05$\pm$1.13          & 61.15$\pm$1.09          & 58.53$\pm$0.86          & 57.89$\pm$1.00 \Tstrut          \\\hline
			\emph{\textbf{AdaptOR}}  & \textbf{59.34$\pm$0.40}                                                                               & \textbf{57.44$\pm$0.42}                                         & \textbf{56.62$\pm$0.41} & \textbf{55.39$\pm$0.51} & \textbf{72.13$\pm$0.91} & \textbf{66.55$\pm$0.80} & \textbf{65.04$\pm$0.52} & \textbf{65.15$\pm$0.65} \Tstrut \\
			\hline
			\bottomrule
		\end{tabular}
	}
	\label{table:source-target-da}
	\vspace{-1mm}
\end{table*}

\begin{figure*}[htb!]
	\includegraphics[clip, trim=0.0cm 1.0cm 0.0cm 0.0cm, width=1.01\linewidth]{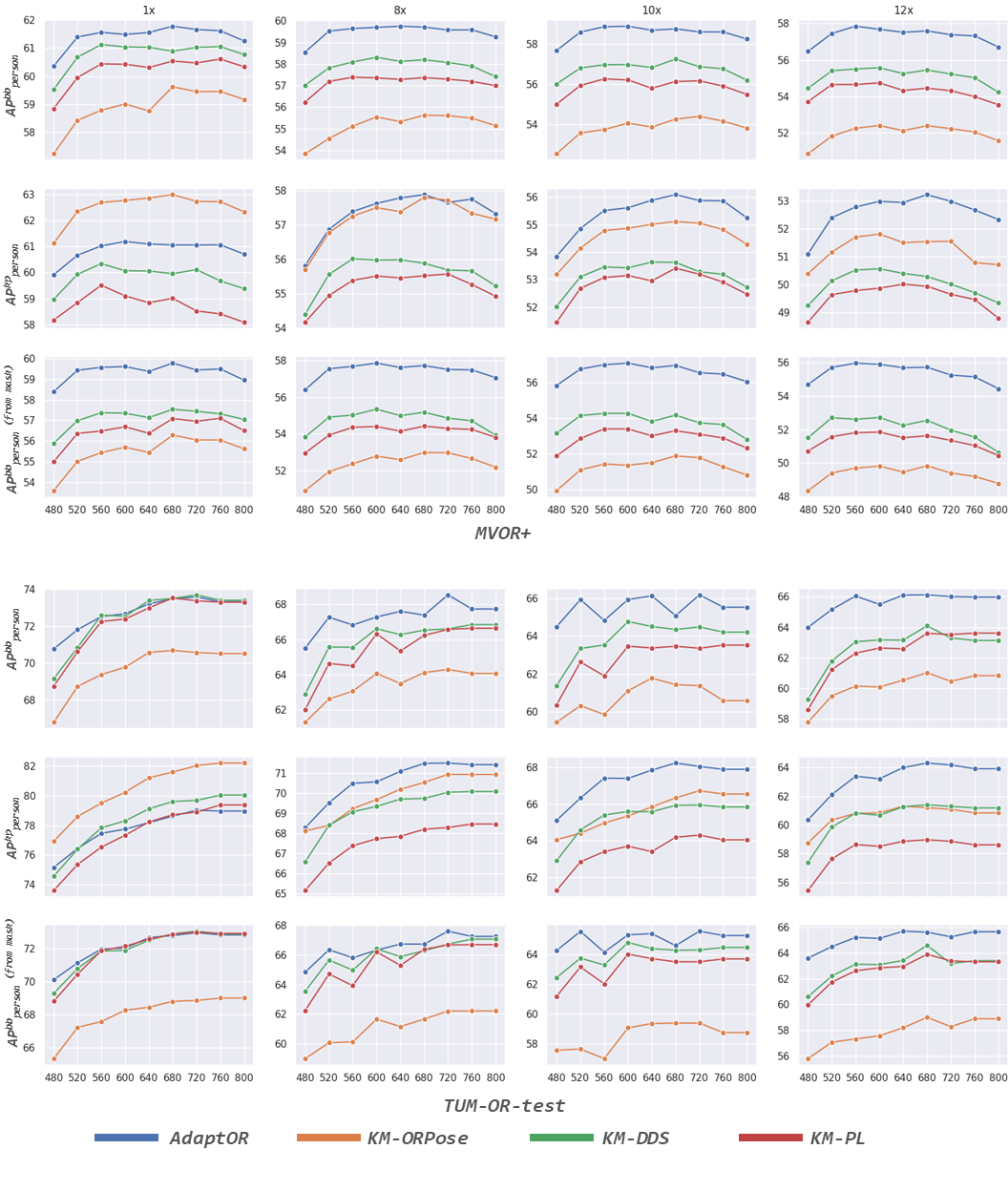}
	\caption{\small{Bounding box detection \emph{$AP_{person}^{bb}$}, pose estimation \emph{$AP_{person}^{kp}$}, and instance segmentation \emph{$AP_{person}^{bb}$ (from mask)} results for unsupervised domain adaptation experiments on four downsampling scales (1x, 8x, 10x, and 12x) and nine target resolution (480, 520, 560, 600, 640, 680, 720, 760, and 800) corresponding to the shorter side of the image for \emph{MVOR+} and \emph{TUM-OR-test} datasets. We see an increase in the accuracy with the increase in target resolution for the \emph{TUM-OR-test} dataset.  We also observe an increase in accuracy for the \emph{MVOR+} dataset but only up to around 680 pixels.}}
	\label{fig:results_graphs}
\end{figure*}
\begin{table*}[t!]
	\centering
	\setlength{\fboxsep}{0pt}%
	\setlength{\fboxrule}{1.5pt}%
	\setlength\tabcolsep{0.5pt}%
	\scalebox{0.95}{
		\begin{tabular}{cccc}
			\centering
			                                                                                                                & 1x                                                                                    & 8x                                                                                    & 12x                                                                                    \\
			\raisebox{5.0\normalbaselineskip}[0pt][0pt]{\parbox[b]{3mm}{\rotatebox[origin=c]{90}{Source-only}}}             & \fbox{\includegraphics[width=2.3in]{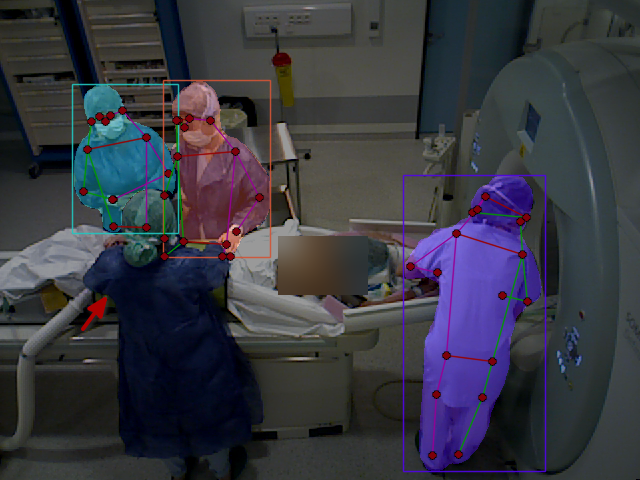}}     & \fbox{\includegraphics[width=2.3in]{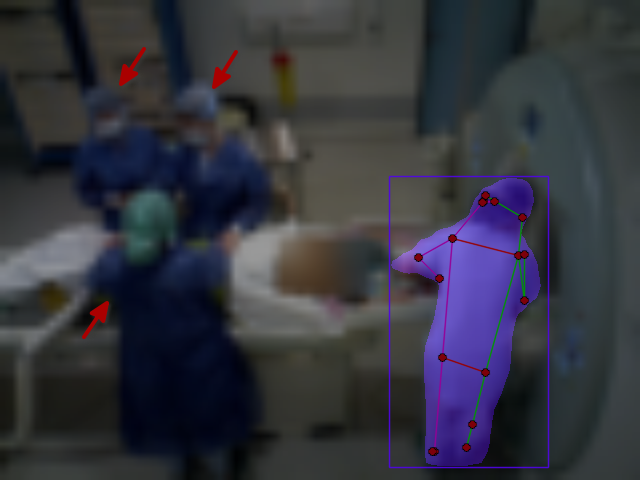}}     & \fbox{\includegraphics[width=2.3in]{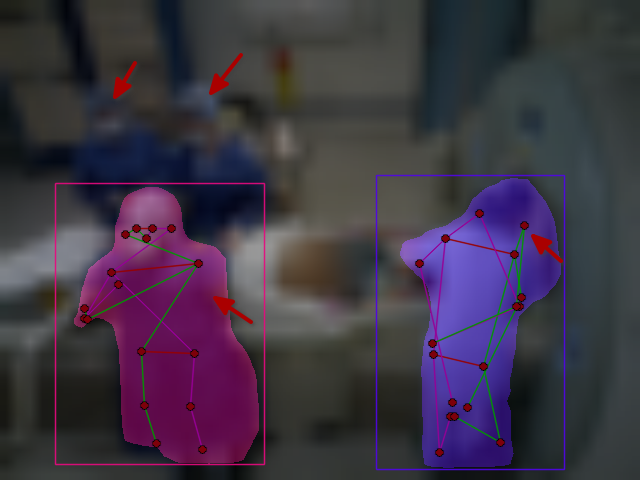}}     \\

			\raisebox{5.0\normalbaselineskip}[0pt][0pt]{\parbox[b]{3mm}{\rotatebox[origin=c]{90}{Pseudo-label}}}            & \fbox{\includegraphics[width=2.3in]{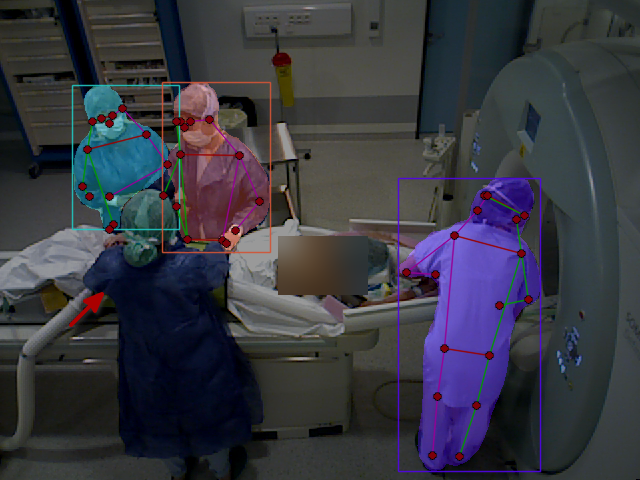}} & \fbox{\includegraphics[width=2.3in]{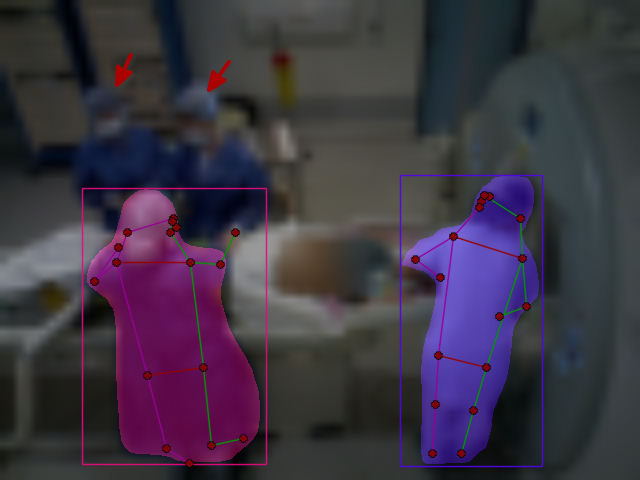}} & \fbox{\includegraphics[width=2.3in]{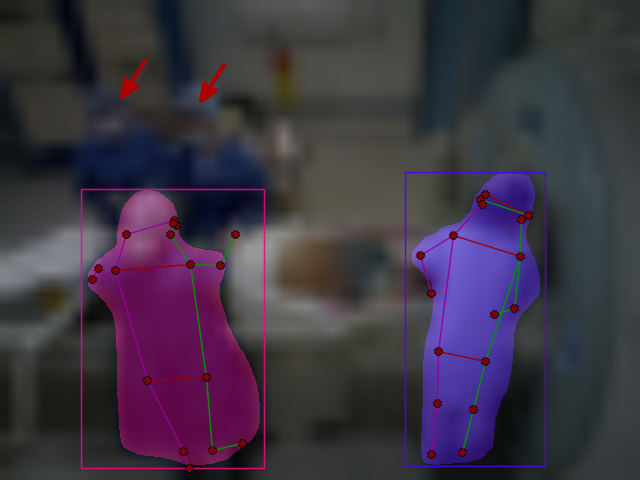}} \\

			\raisebox{5.0\normalbaselineskip}[0pt][0pt]{\parbox[b]{3mm}{\rotatebox[origin=c]{90}{Data-distillation}}}       & \fbox{\includegraphics[width=2.3in]{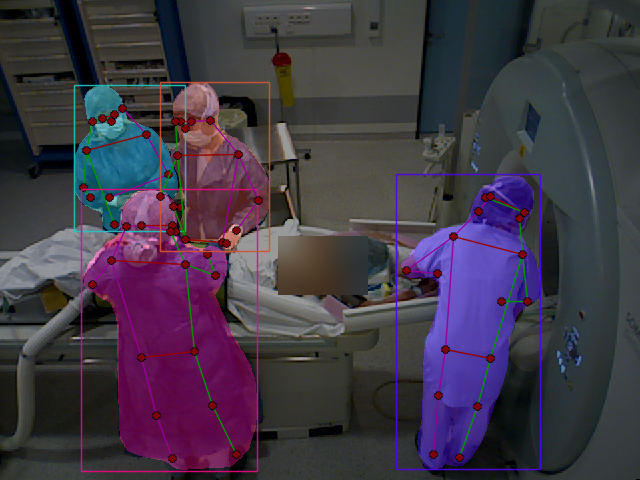}}   & \fbox{\includegraphics[width=2.3in]{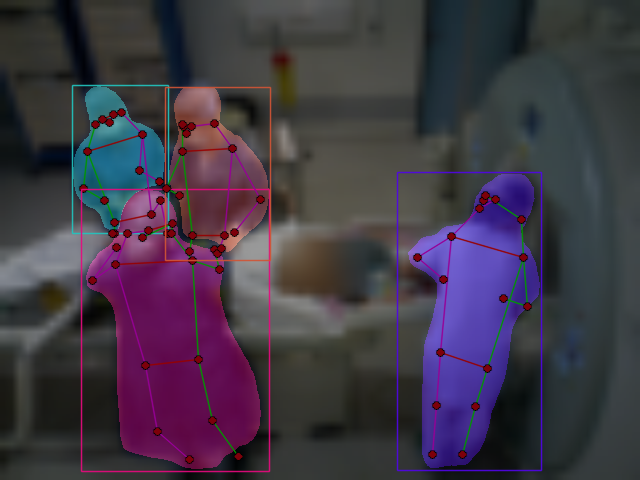}}   & \fbox{\includegraphics[width=2.3in]{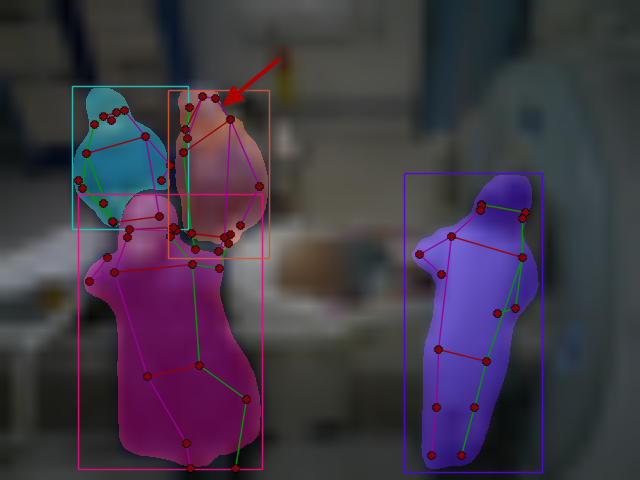}}   \\

			\raisebox{5.0\normalbaselineskip}[0pt][0pt]{\parbox[b]{3mm}{\rotatebox[origin=c]{90}{ORPose}}}                  & \fbox{\includegraphics[width=2.3in]{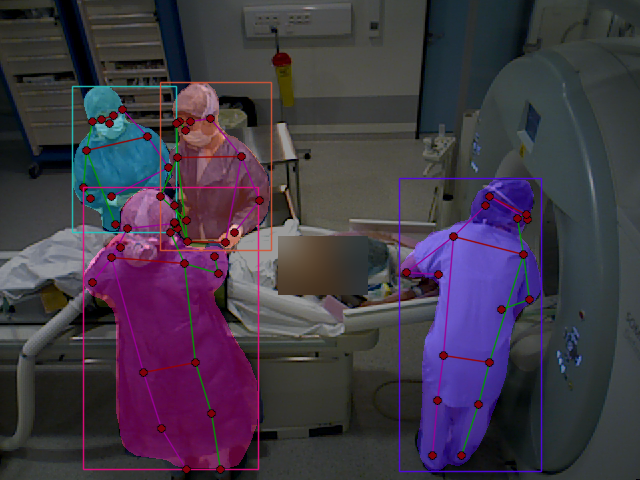}}       & \fbox{\includegraphics[width=2.3in]{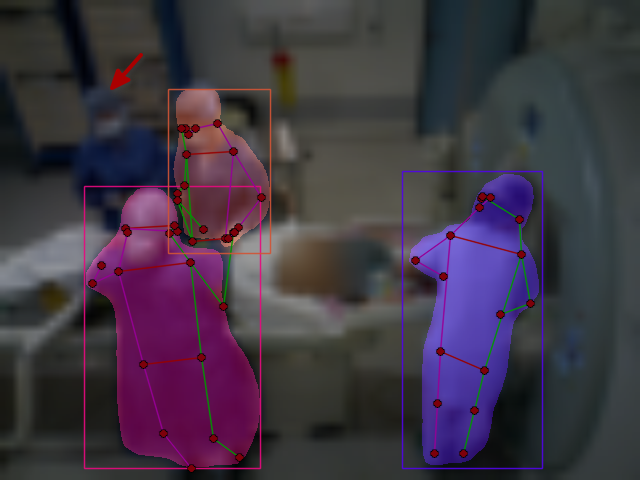}}       & \fbox{\includegraphics[width=2.3in]{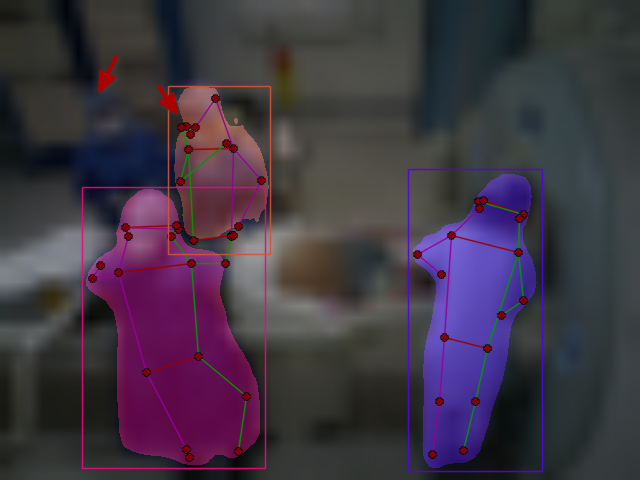}}       \\

			\raisebox{5.0\normalbaselineskip}[0pt][0pt]{\parbox[b]{3mm}{\rotatebox[origin=c]{90}{\emph{\textbf{AdaptOR}}}}} & \fbox{\includegraphics[width=2.3in]{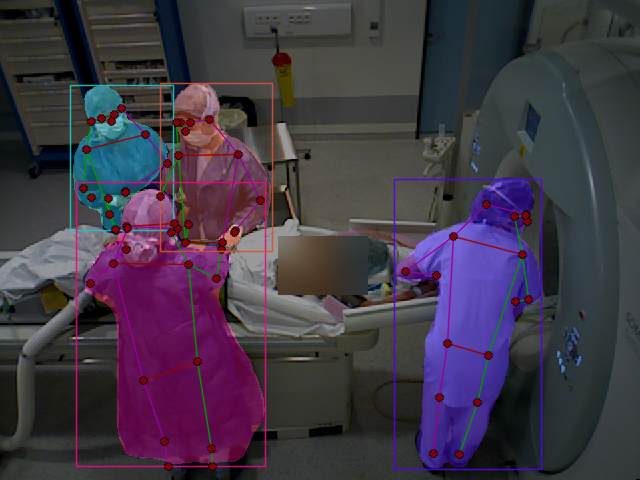}}      & \fbox{\includegraphics[width=2.3in]{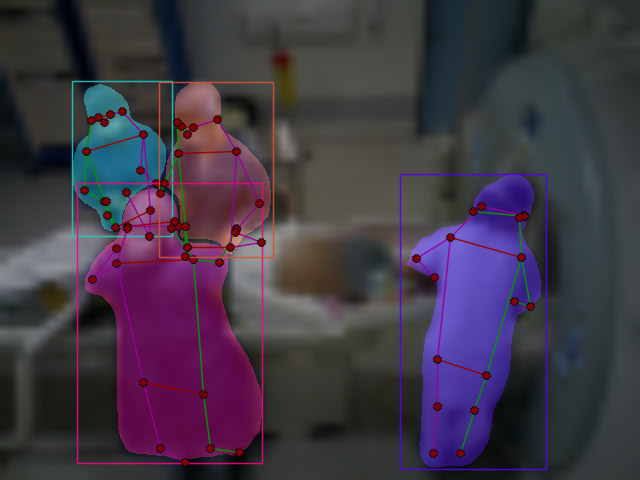}}      & \fbox{\includegraphics[width=2.3in]{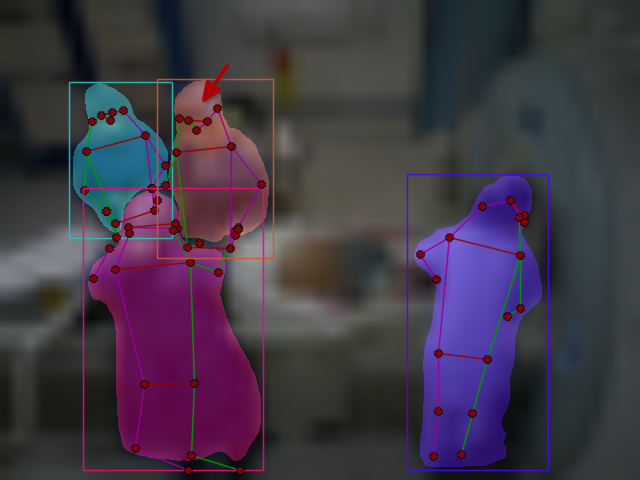}}      \\
		\end{tabular}
	}
	\captionof{figure}[]{\small{Qualitative results for bounding box detection, pose estimation, and instance segmentation on a sample \emph{MVOR+} image for the baseline approaches and \emph{AdaptOR}. Results are displayed on the for original image and corresponding downsampled images with downsampling factor 8 and 12. The red arrows show either missed detections or localization errors. Localization errors are noticeable on the low-resolution images}}
	\label{tab:table-qual1}
\end{table*}

\begin{table*}[t!]
	\centering
	\setlength{\fboxsep}{0pt}%
	\setlength{\fboxrule}{1.5pt}%
	\setlength\tabcolsep{0.5pt}%
	\scalebox{0.95}{
		\begin{tabular}{cccc}
			\centering
			                                                                                                                & 1x                                                                                & 8x                                                                                & 12x                                                                                \\
			\raisebox{5.0\normalbaselineskip}[0pt][0pt]{\parbox[b]{3mm}{\rotatebox[origin=c]{90}{Source-only}}}             & \fbox{\includegraphics[width=2.3in]{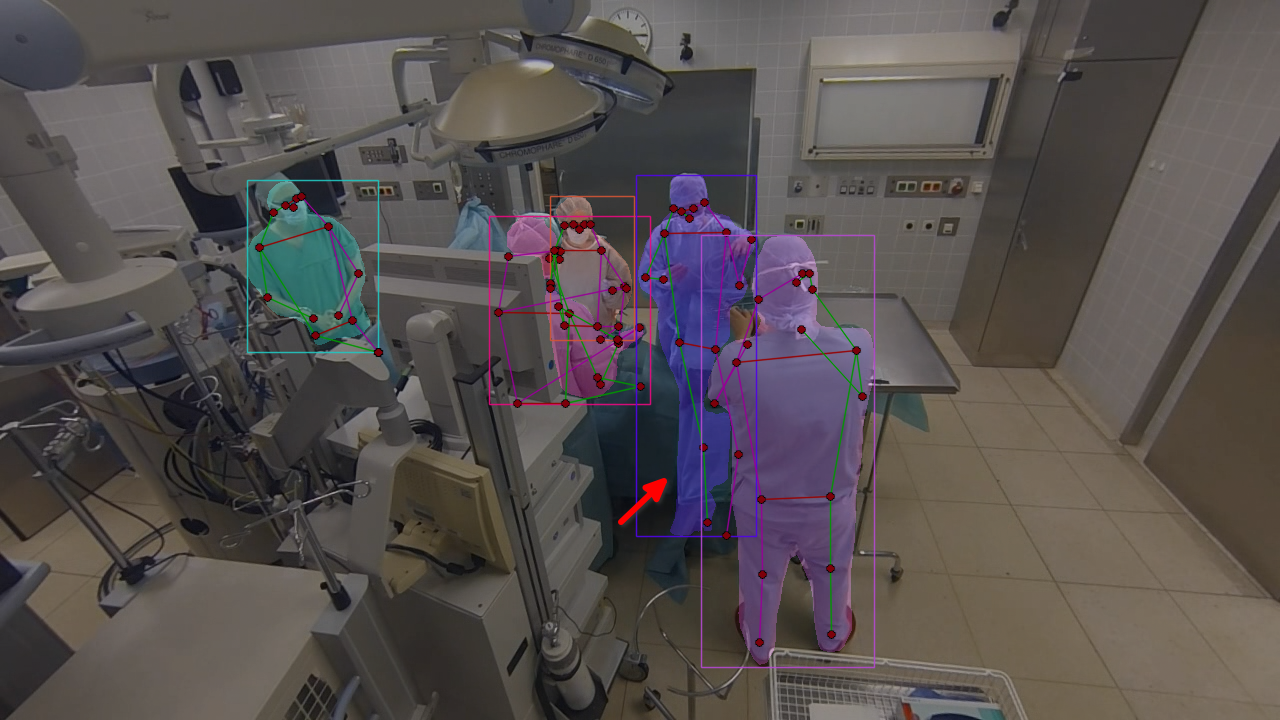}}     & \fbox{\includegraphics[width=2.3in]{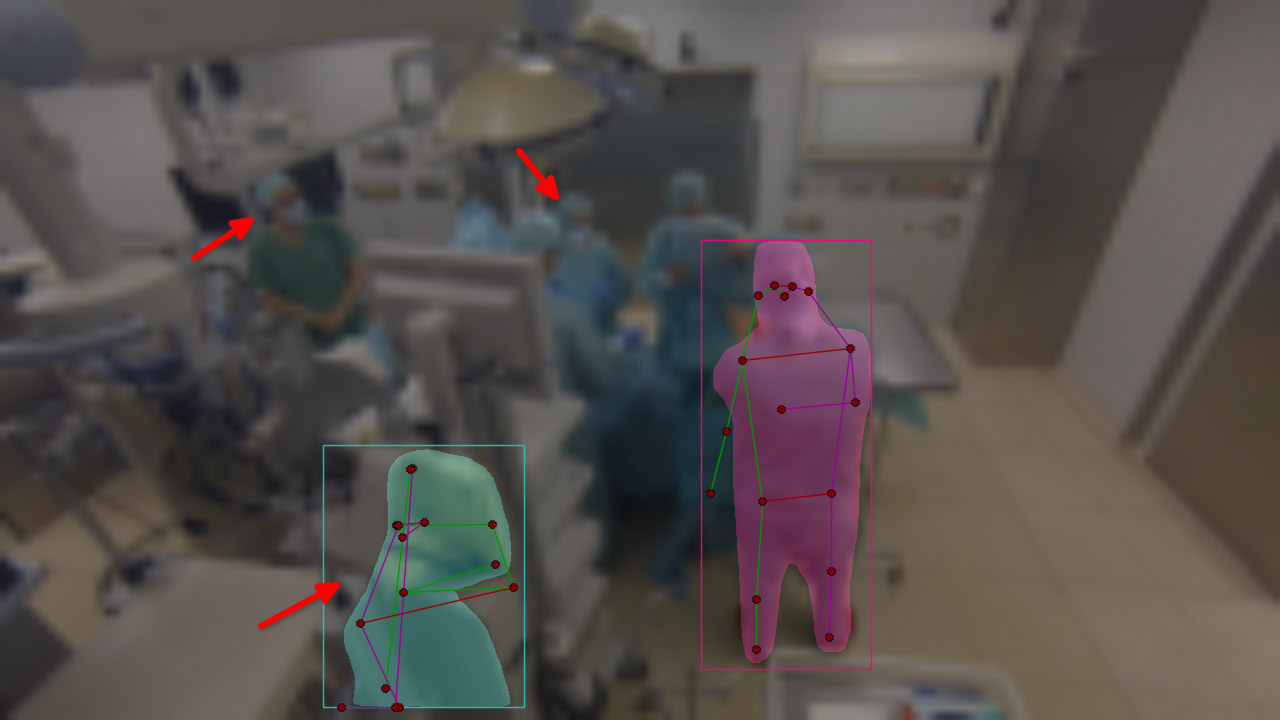}}     & \fbox{\includegraphics[width=2.3in]{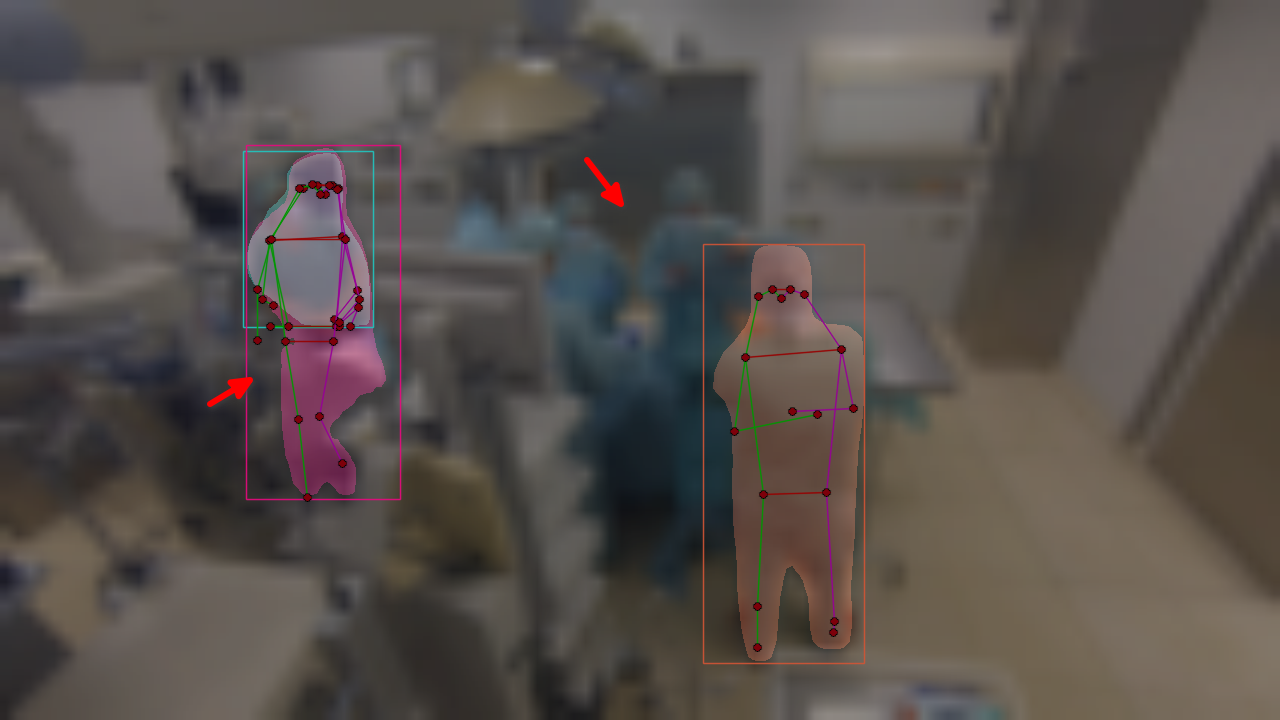}}     \\

			\raisebox{5.0\normalbaselineskip}[0pt][0pt]{\parbox[b]{3mm}{\rotatebox[origin=c]{90}{Pseudo-label}}}            & \fbox{\includegraphics[width=2.3in]{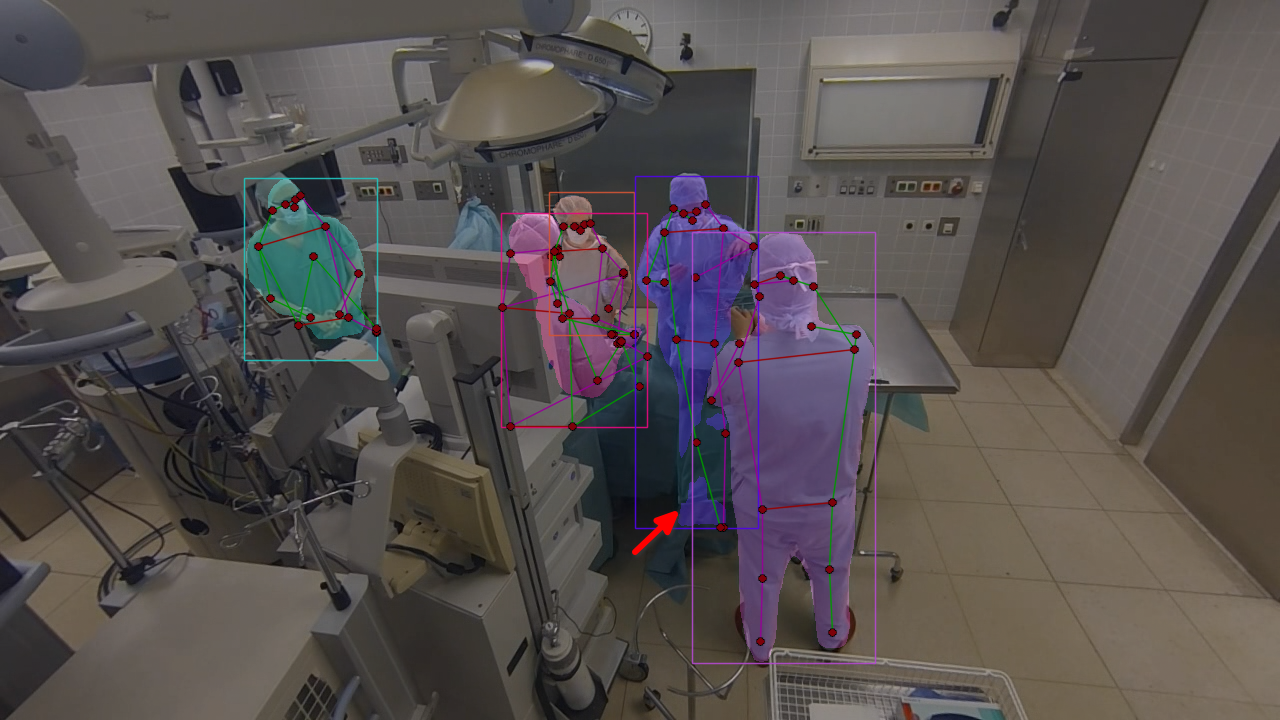}} & \fbox{\includegraphics[width=2.3in]{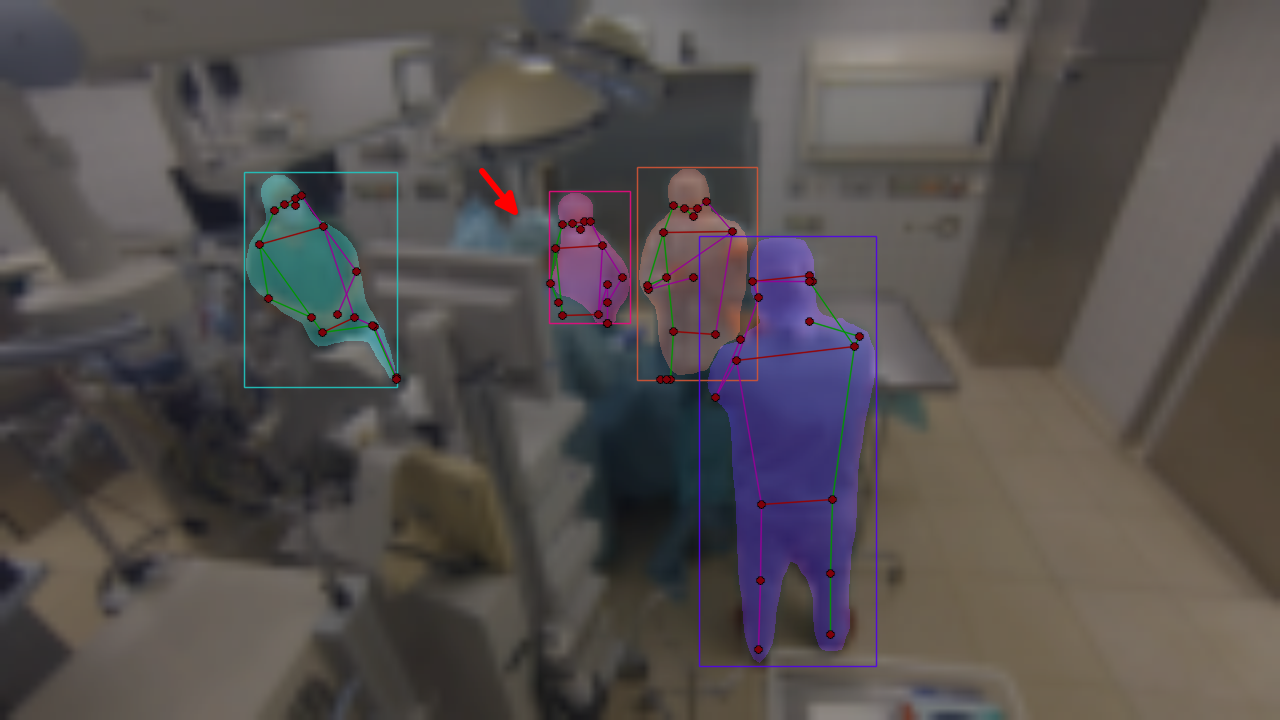}} & \fbox{\includegraphics[width=2.3in]{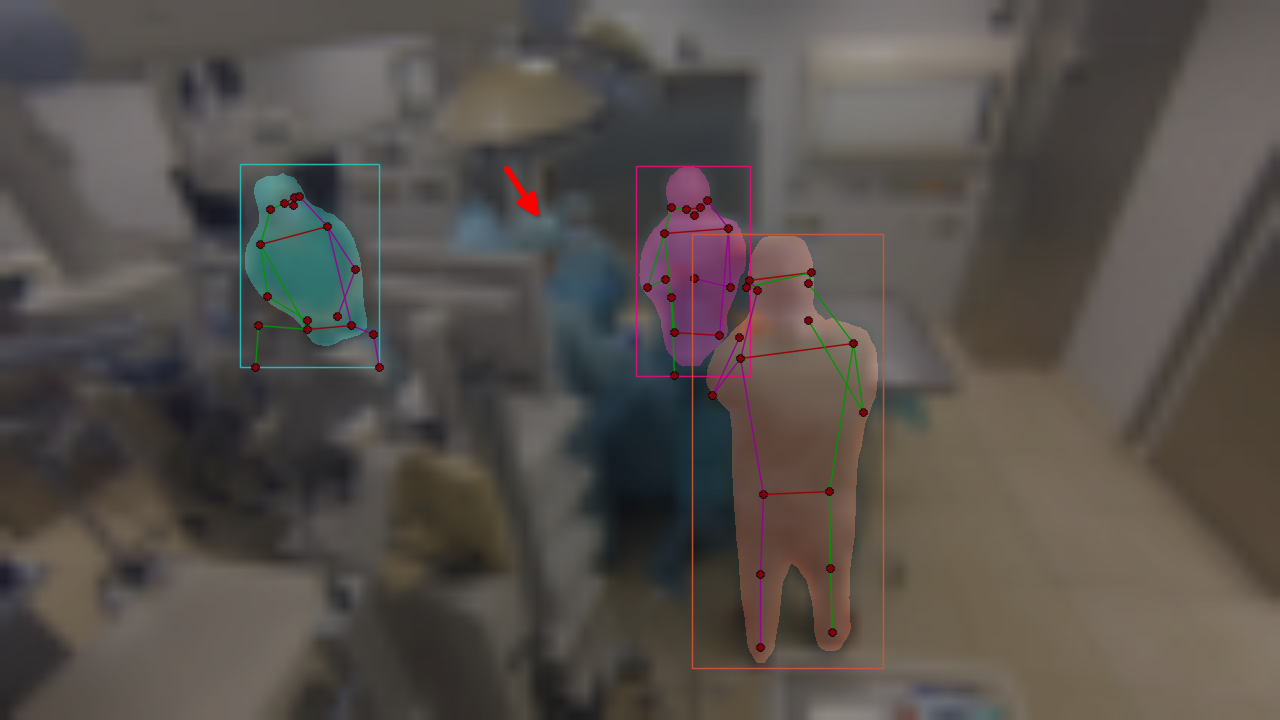}} \\

			\raisebox{5.0\normalbaselineskip}[0pt][0pt]{\parbox[b]{3mm}{\rotatebox[origin=c]{90}{Data-distillation}}}       & \fbox{\includegraphics[width=2.3in]{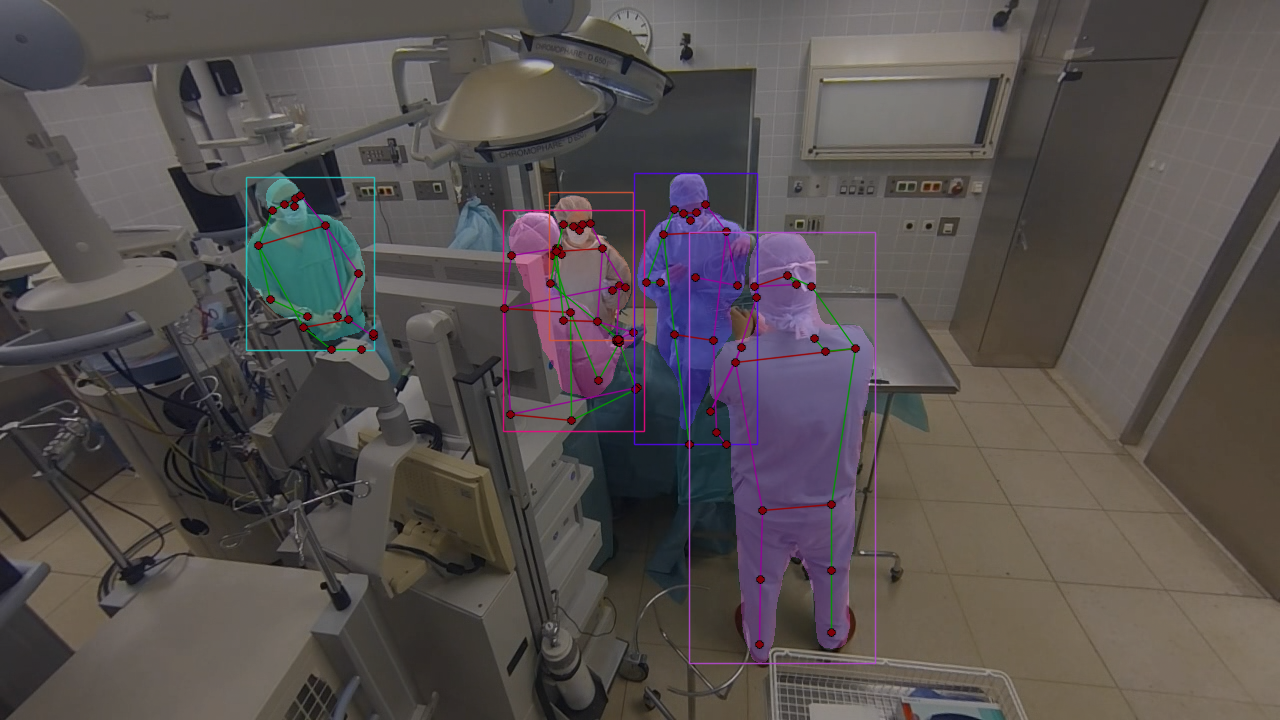}}   & \fbox{\includegraphics[width=2.3in]{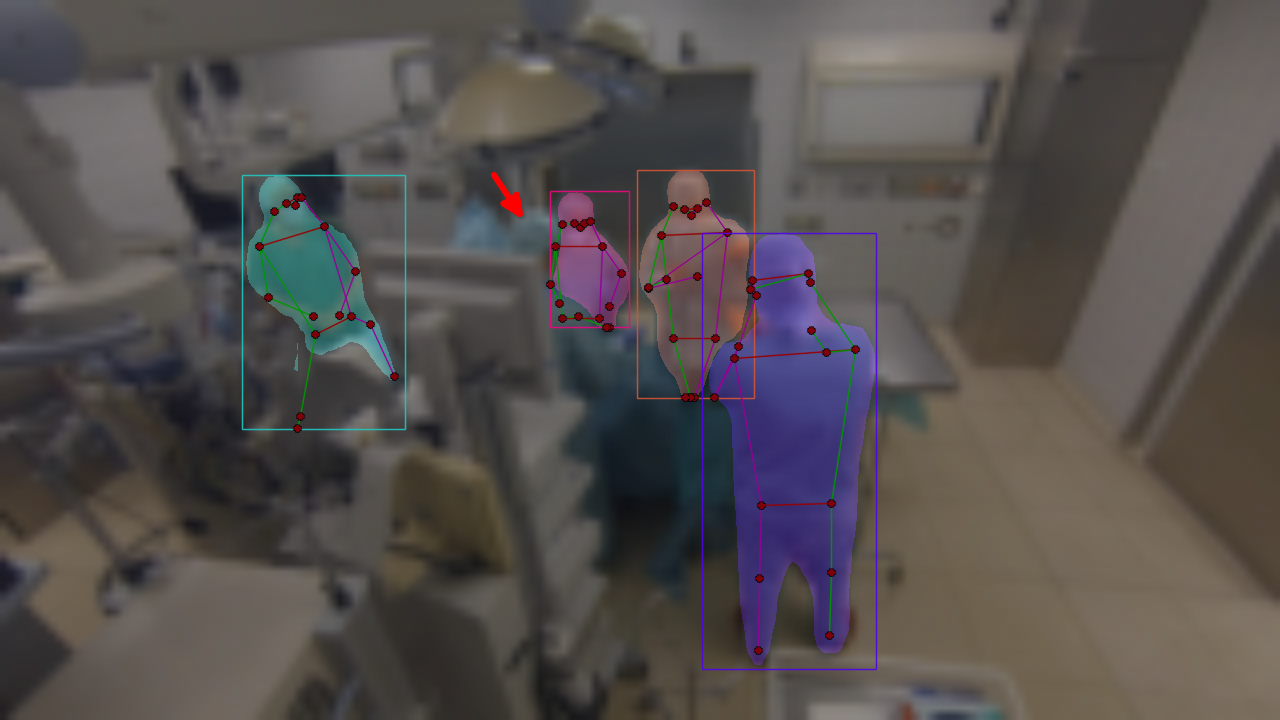}}   & \fbox{\includegraphics[width=2.3in]{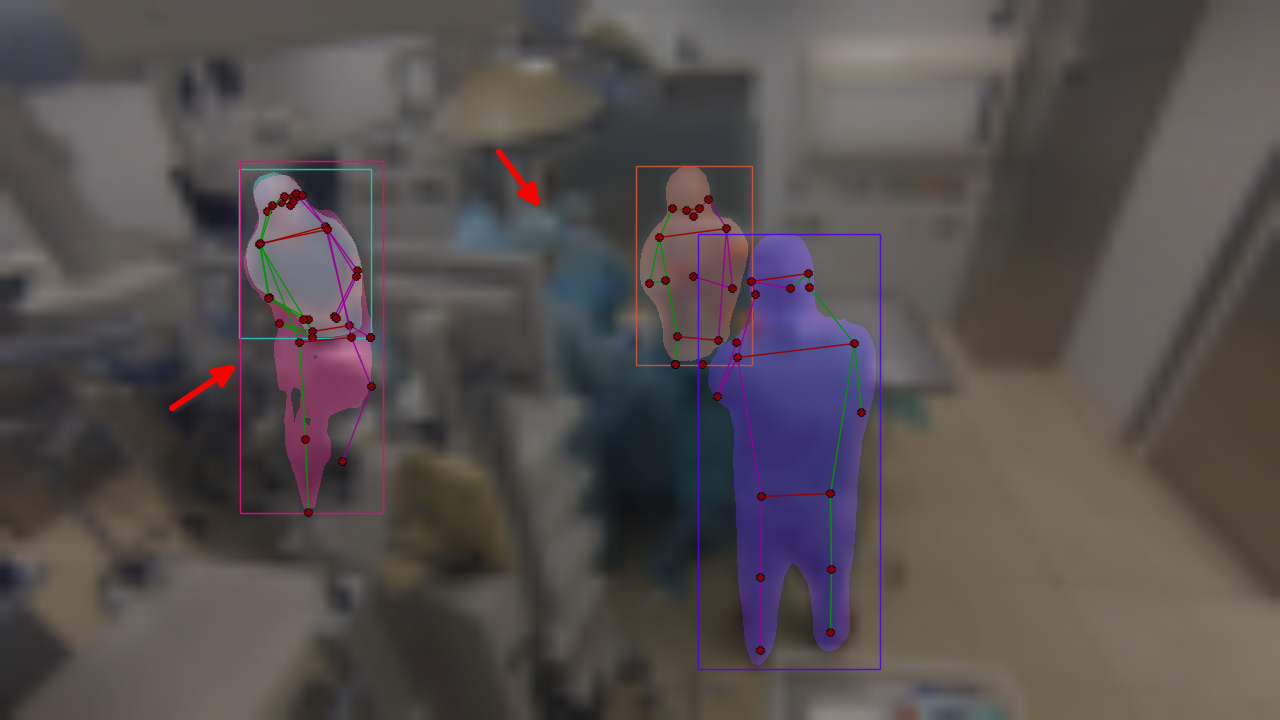}}   \\

			\raisebox{5.0\normalbaselineskip}[0pt][0pt]{\parbox[b]{3mm}{\rotatebox[origin=c]{90}{ORPose}}}                  & \fbox{\includegraphics[width=2.3in]{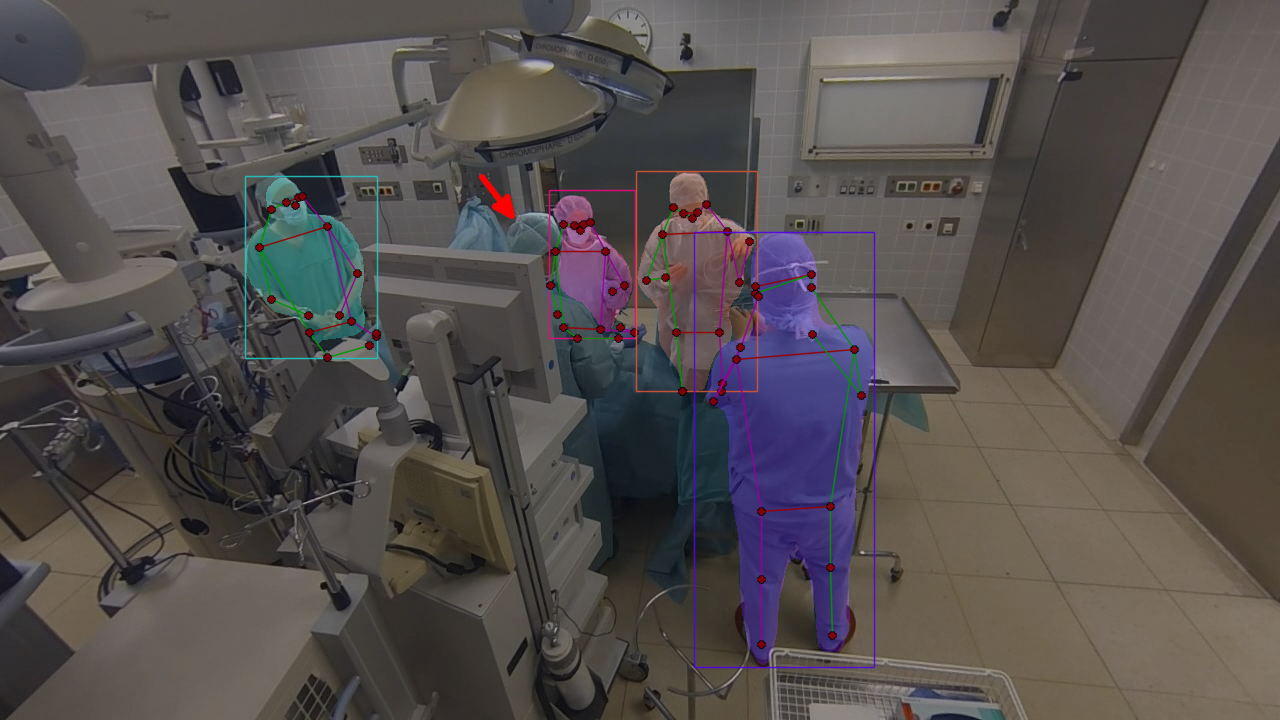}}       & \fbox{\includegraphics[width=2.3in]{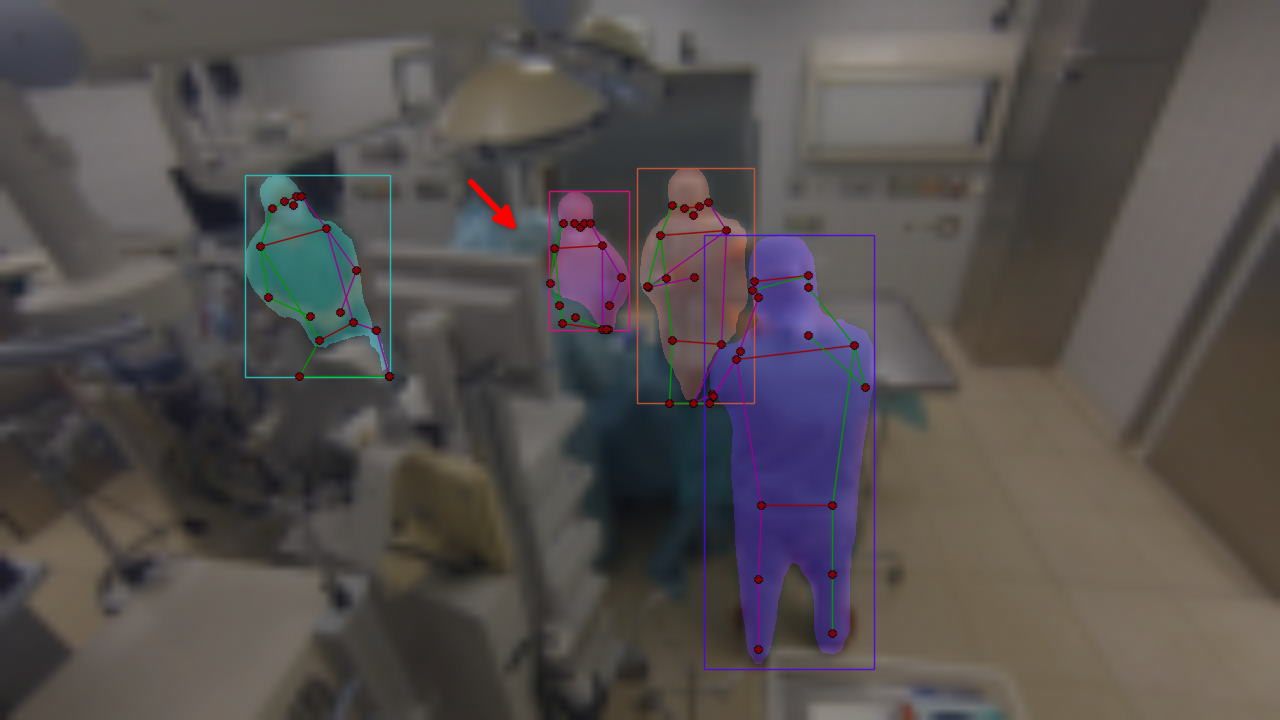}}       & \fbox{\includegraphics[width=2.3in]{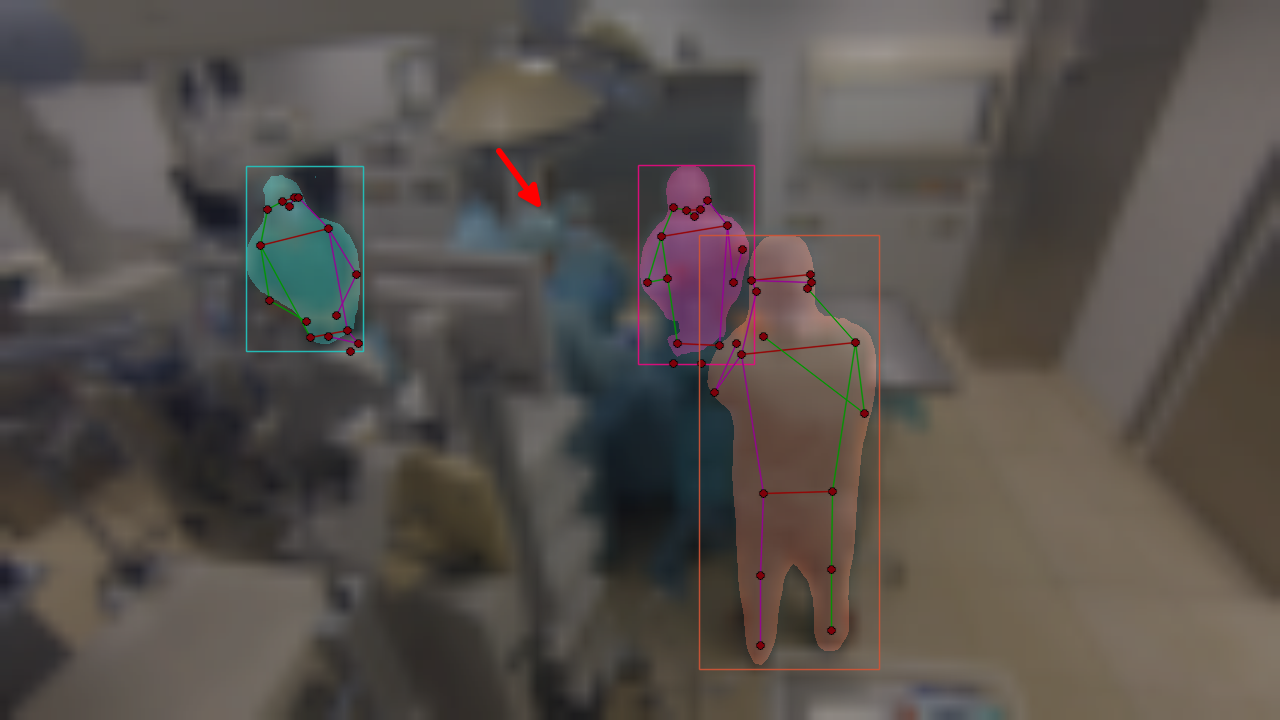}}       \\

			\raisebox{5.0\normalbaselineskip}[0pt][0pt]{\parbox[b]{3mm}{\rotatebox[origin=c]{90}{\emph{\textbf{AdaptOR}}}}} & \fbox{\includegraphics[width=2.3in]{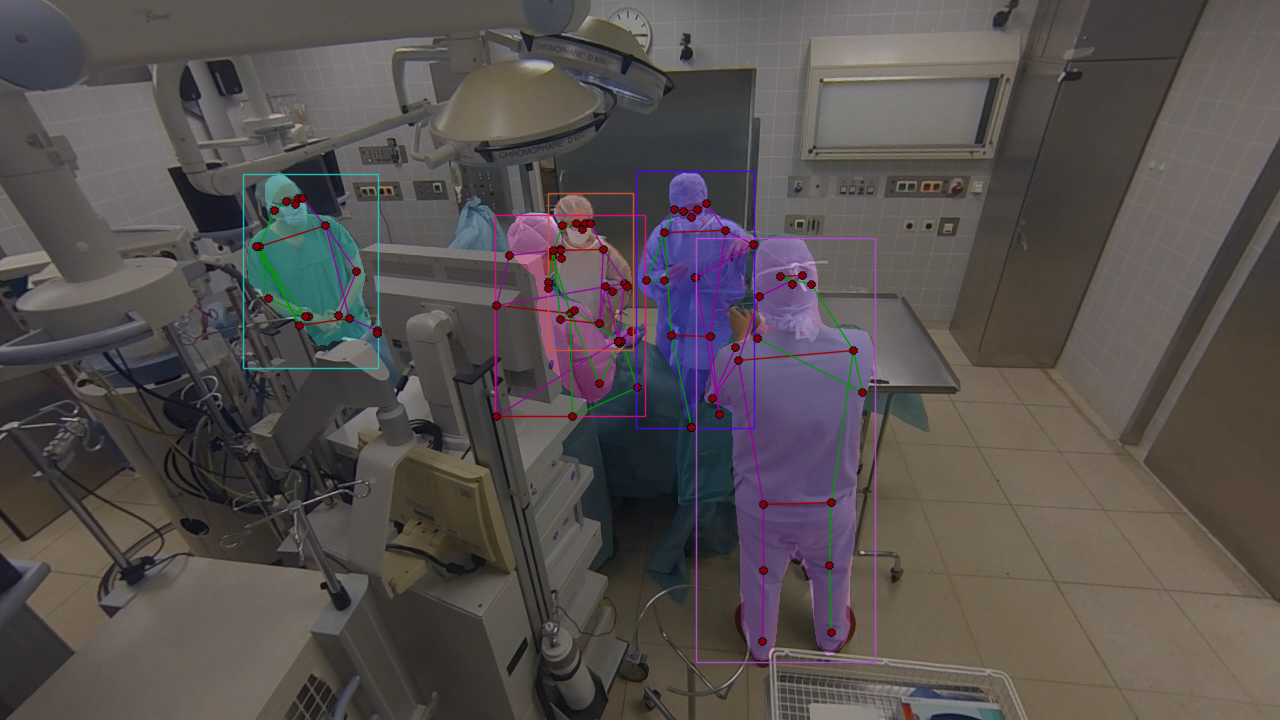}}      & \fbox{\includegraphics[width=2.3in]{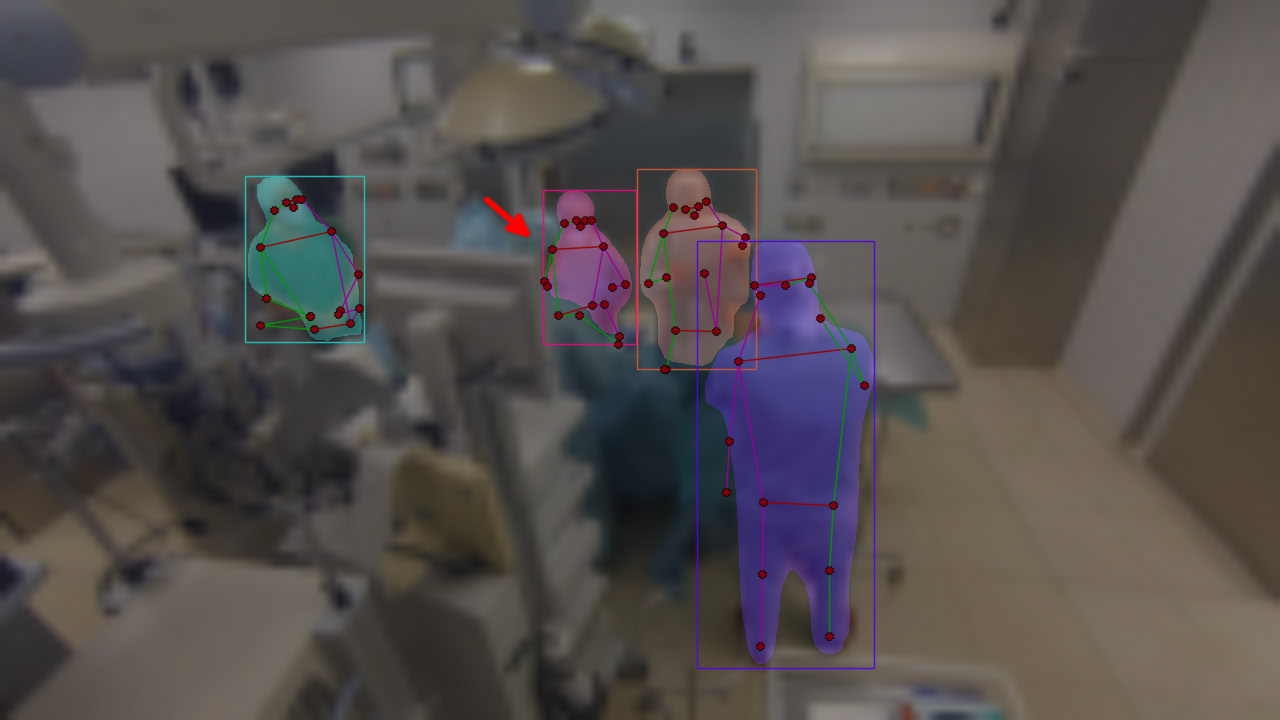}}      & \fbox{\includegraphics[width=2.3in]{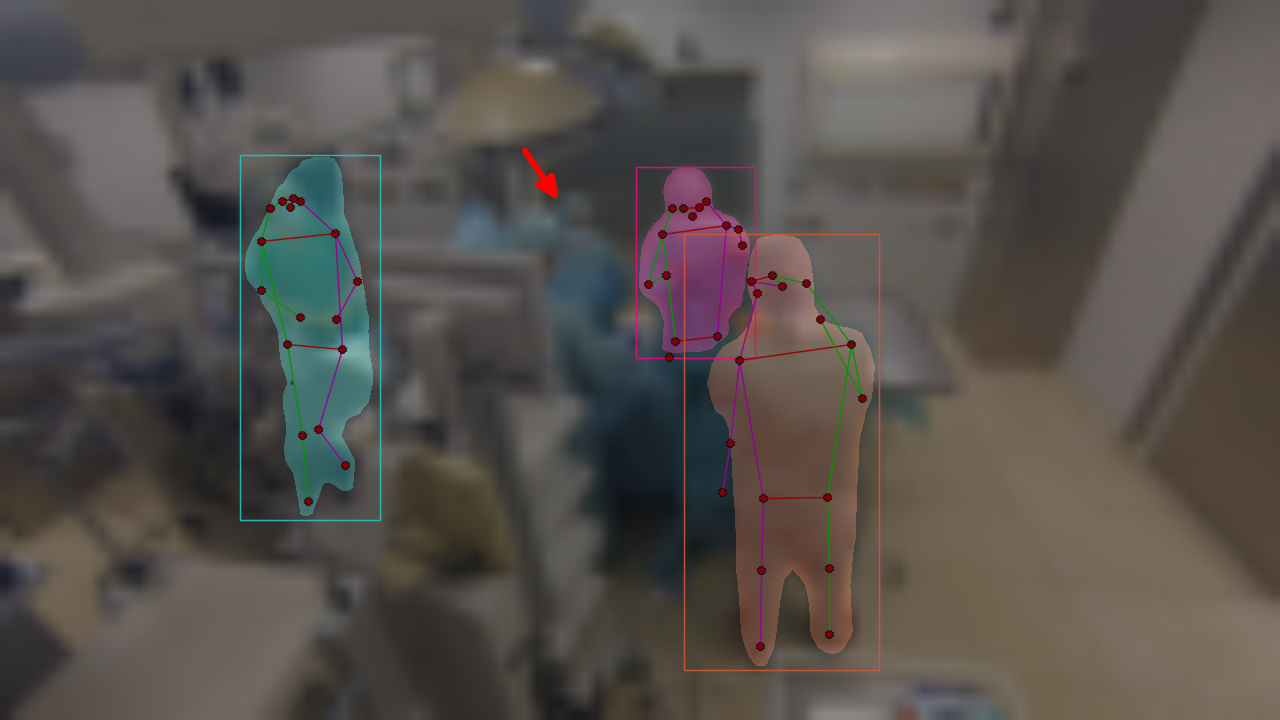}}      \\
		\end{tabular}
	}
	\captionof{figure}[]{\small{{\blue Qualitative results for bounding box detection, pose estimation, and instance segmentation on a sample \emph{TUM-OR-test} image for the baseline approaches and \emph{AdaptOR}.}}}
	\label{tab:table-qual-tum-or}
\end{table*}

\subsection{KM-PL} We modify the pseudo-labeling \citep{lee2013pseudo} approach to generate the pseudo labels on a single-scale image on the unlabeled target domain data. The authors in \citep{sohn2020simple} recently use a similar approach with advanced data augmentations for the object detection task.

\subsection{KM-DDS} KM-DDS \citep{radosavovic2018data} is also a pseudo-labeling approach, but instead of generating pseudo labels on a single scale, it aggregates the labels from multiple scales with random horizontal flipping transformations. Authors use the approach for multi-class object detection and human pose estimation. We further extend it to generate pseudo labels for the masks. Similar to the authors, we use scaling and random horizontal flipping transformations on nine predefined image sizes ranging from 400 to 1200 pixels with a step size of 100. Here, the image size corresponds to the shorter side of the image; the size of the longer side of the image is computed by maintaining the same aspect ratio.
\subsection{KM-ORPose} KM-ORPose \citep{srivastav2020human} uses the \emph{teacher-student} learning paradigm for the domain adaptation in the OR for joint person detection and 2D/3D human pose estimation. It combines the knowledge-distillation \citep{hinton2015distilling, zhang2019fast} - using complex three-stage models - along with data-distillation \citep{radosavovic2018data} to generate accurate pseudo labels. In the first stage, it uses cascade-mask-rcnn \citep{cai2019cascade} with the deformable convolution \citep{dai2017deformable} based resnext-152 backbone \citep{xie2017aggregated} to generate the person bounding boxes. We use the same network to get the pseudo masks as well. In the second stage, it uses the HRNet-w48 model (384x288 input size) \citep{SunXLW19} to get the pseudo labels for the poses. KM-ORPose is a strong baseline as it uses a complex multi-stage teacher model to generate accurate pseudo labels for the training.

\section{Experiments and results}
\subsection{Datasets and evaluation metrics}
We use COCO \citep{lin2014microsoft} as source domain dataset. It contains 57k images, and the ground truth labels have 150k instances of person bounding box, segmentation mask, and 17 body keypoints. The test dataset of COCO, called \emph{COCO-val}, contain 5k images with 10777 person instances.

We train and evaluate our approach on the two target domain OR datasets: MVOR \citep{srivastav2018mvor,srivastav2020human} and TUM-OR \citep{belagiannis2016parsing}. MVOR contains data captured during real surgical interventions, whereas TUM-OR contains OR images from simulated surgical activities. The unlabelled training datasets of MVOR and TUM-OR contain 80k and 1.5k images, respectively. The testing dataset of MVOR, called \emph{MVOR+}, and TUM-OR, called \emph{TUM-OR-test}, contain 2196 images with 5091 person instances and 2400 images with 11611 person instances, respectively. The \emph{MVOR+} dataset is extended from the public \emph{MVOR} dataset \citep{srivastav2018mvor,srivastav2020human}. Before the extension, it consists of 4699 person bounding boxes, 2926 2D upper body poses with 10 keypoints, (and 1061 3D upper body poses). The fully-annotated extension called \emph{MVOR+} consists of 5091 person bounding boxes, and 5091 body poses with 17 keypoints in the COCO format. The original \emph{TUM-OR-test} consists of only the upper-body bounding boxes with six common COCO keypoints. These annotations are not suitable for our evaluation purpose; hence we annotate the \emph{TUM-OR-test} using a semi-automatic approach. We first use a state-of-the-art person detector \citep{cai2019cascade} to get the person bounding boxes and manually correct all the bounding boxes. We then run the HRNet model \citep{SunXLW19} on all the corrected bounding boxes to get the poses. The predicted poses are corrected using the keypoint annotation tool\footnote{\url{https://github.com/visipedia/annotation_tools}}. 
{\blue An overview of the datasets used in this work is shown in the Table \ref{table:datasets}}.

The image sizes of \emph{MVOR+} and \emph{TUM-OR-test} datasets are 640x480 and 1280x720, respectively. We also conduct experiments with downsampled images using the scaling factors 8x, 10x, and 12x, yielding images of size 80x64, 64x48, and 53x40 for the \emph{MVOR+} dataset and 160x90, 128x72, and 107x60 for the \emph{TUM-OR-test} dataset.

We use the Average Precision \emph{$AP_{0.5:0.95}$} metric from COCO \citep{lin2014microsoft} for the evaluation. The bounding box evaluation metric \emph{$AP_{person}^{bb}$} uses intersection over union (IoU) over boxes, and the pose estimation evaluation metric \emph{$AP_{person}^{kp}$} uses the object keypoint similarity (OKS) over person keypoints to compare the ground-truth and the predictions. Both \emph{MVOR+} and \emph{TUM-OR-test} do not have a ground-truth for the person instance segmentation masks. Hence, we evaluate the mask predictions by computing a tight bounding box on the prediction masks and comparing them with ground-truth bounding boxes called \emph{$AP_{person}^{bb}$ (from mask)}. We also show extensive qualitative results for the instance segmentation and pose estimation in the supplementary video. The instance segmentation on the source domain COCO images is evaluated using the \emph{$AP_{person}^{mask}$} which uses IoU over masks to compare the ground-truth and the predictions.

\subsection{Experiments}
\subsubsection{Source domain fully supervised training} \label{sdfs}
The models are trained on the source domain COCO dataset in a fully supervised manner for three experiments: supervised ImageNet initialization with Frozen batch normalization (BN) \citep{he2016deep}, self-supervised MOCO-v2 initialization \citep{chen2020simple,he2020momentum} with Cross-GPU BN \citep{peng2018megdet}, and self-supervised MOCO-v2 initialization \citep{chen2020simple,he2020momentum} with group normalization (GN) \citep{wu2018group}. The goal of these experiments is to obtain one suitable \emph{source-only} baseline as an initialization model for the UDA experiments. The last model with self-supervised MOCO-v2 initialization and GN, called \emph{kmrcnn+}, is further used in the SSL experiments and extended in UDA experiments.
\subsubsection{AdaptOR: unsupervised domain adaptation (UDA) on target domains} \label{stduda}
The UDA experiments on source domain COCO datasets and target domains MVOR and TUM-OR datasets are conducted to train the \emph{kmrcnn++} model for eight sets of experiments. The first four experiments are for the target domain MVOR and the last four for TUM-OR. For each target domain, the first three experiments train the \emph{kmrcnn++} model on three constructed baseline methods: KM-PL, KM-DDS, and KM-ORPose, respectively, and the fourth experiment trains the \emph{kmrcnn++} model on our \emph{AdaptOR} method. Eleven ablation experiments are conducted with the source domain COCO dataset and the target domain MVOR dataset: the first experiment evaluates the contribution of disentangled feature normalization, the next five different types of strong augmentations, and the {\blue last five different unsupervised loss weights loss values $\lambda$}.
\subsubsection{AdaptOR-SSL: semi-supervised learning (SSL) on source-domain} \label{sdssl}
The SSL experiments on the source domain COCO dataset are conducted for four experiments where we train the \emph{kmrcnn+} model using 1\%, 2\%, 5\%, and 10\% of COCO dataset as the labeled set and the rest of the data as the unlabeled set. The \emph{kmrcnn+} model uses the regular GN layers instead of disentangled feature normalization layers. We use the same labeled and unlabeled images and training iterations as used by Unbiased-teacher \citep{liu2021unbiased}, the current state-of-the-art in SSL for object detection. 

{\blue \subsubsection{Domain adaptation on AdaptOR-SSL model} \label{sdudaonssl}
\emph{AdaptOR} assumes it has access to all the source-domain labels in the previous experiments. We conduct a final experiment to see how  \emph{AdaptOR} performs when initialized from a source-domain model trained with less source domain data. We take a \emph{AdaptOR-SSL} model trained using 10\% labeled and 90\% unlabeled source domain data and use it to initialize \emph{AdaptOR}.}

\subsection{Implementation details}
The source domain fully supervised training experiments, explained in section \ref{sdfs}, are conducted with batch size 16 and learning rate 0.02 for 270k iterations with multi-step (210k and 250k) learning rate decay on eight V100 GPUs. 

The \emph{AdaptOR} and the \emph{AdaptOR-on-AdaptOR-SSL} experiments explained in section \ref{stduda}, \ref{sdudaonssl}, respectively, are conducted on four V100 GPUs with a labeled and unlabeled batch size of eight (four images/GPU) and a learning rate of 0.001. The experiments are conducted for 65k iterations for the MVOR dataset and 10k iterations for the TUM-OR dataset. Finally, the \emph{AdaptOR-SSL} experiments explained in \ref{sdssl} are conducted on four V100 GPUs following the linear learning rate scaling rule \citep{goyal2017accurate}.

The spatial augmentations from rand-augment \citep{cubuk2020randaugment} consist of ``inversion'', ``auto-contrast'', ``posterize'', ``equalize'', ``solarize'', ``contrast-variation'', ``color-jittering'', ``sharpness-variations'', and  ``brightness-variations'' {\blue implemented using a python image library\footnote{\url{https://github.com/jizongFox/pytorch-randaugment}}}. 
The random cut-out \citep{devries2017improved} augmentation places square boxes of random sizes chosen between 40 to 80 pixels at random locations in the image. The random-resize operation for the \emph{weakly} and \emph{strongly} augmented images resize the image to a size randomly sampled from 600 to 800 pixels for SSL experiments following \citep{he2017mask}. For the UDA experiments, we choose the random-resize range from 480 to 800 pixels to provide more size variability in the data augmentation and match the original size of the MVOR dataset (640x480). The image size corresponds to the shorter side of the image.

We use a detectron2 framework \citep{wu2019detectron2} to run all the experiments with automatic mixed precision (AMP) \citep{micikevicius2017mixed}. We use bounding box threshold $\delta_{bbox} = 0.7$, keypoint threshold $\delta_{kp} = 0.1$, mask threshold $\delta_{mask} = 0.5$, EMA decay rate $\alpha = 0.9996$, unsupervised loss weight $\lambda = 3.0$ for \emph{AdaptOR}, and $\lambda = 2.0$ for \emph{AdaptOR-SSL}.

\begin{figure*}[t!]
	\centering
	\begin{subfigure}[t]{0.8\textwidth}
		\centering
		\resizebox{\linewidth}{!}{
			\begin{tabular}{ccc}
				\raisebox{15mm}{\multirow{2}{*}{\includegraphics[width=.4\linewidth]{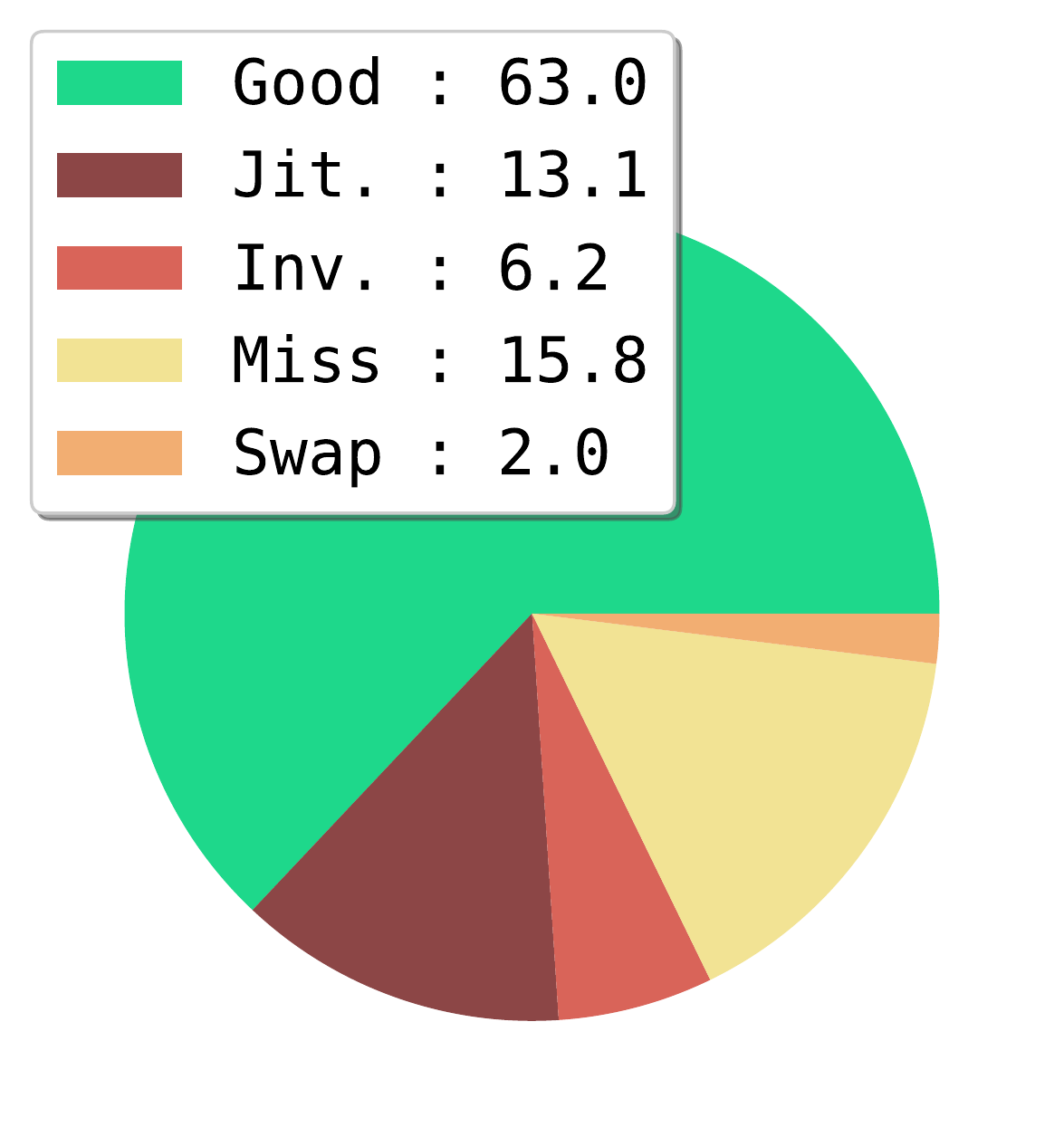}}}
					& \includegraphics[width=.99\linewidth]{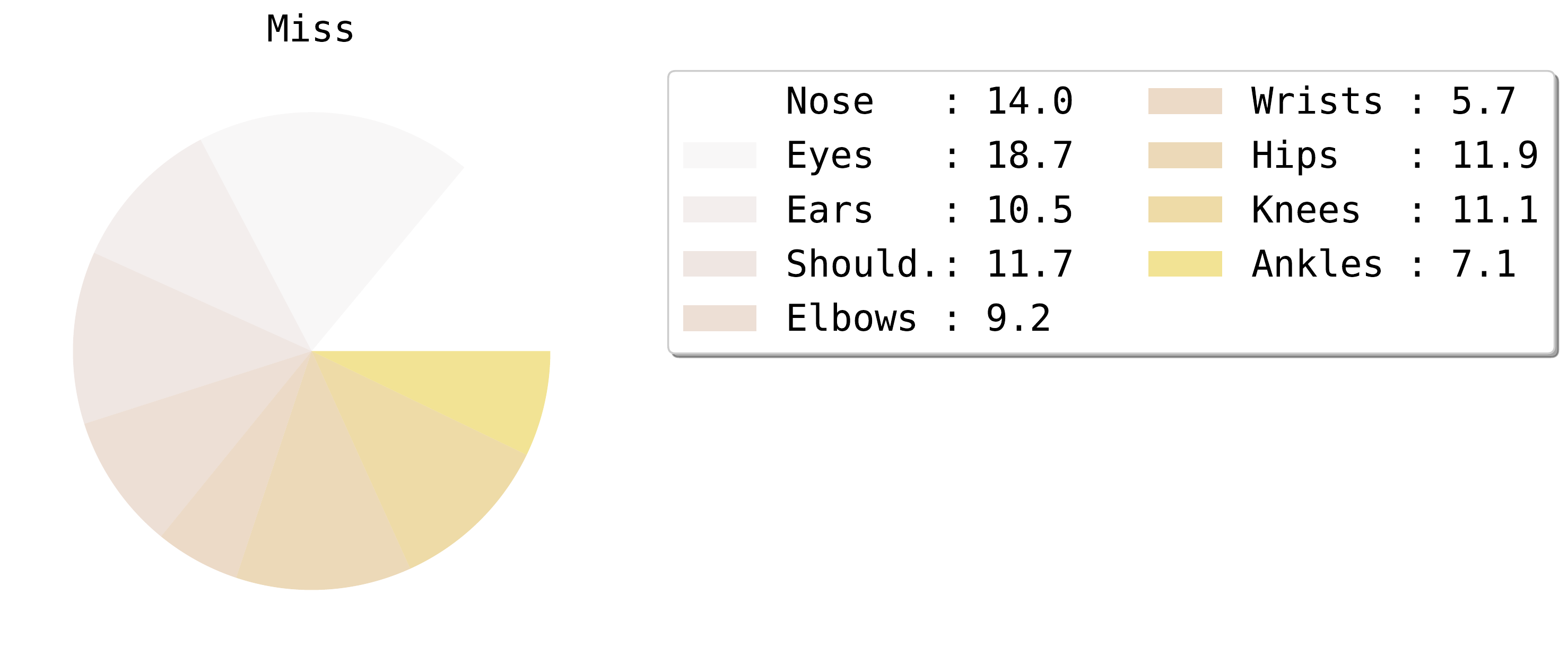}
					& \includegraphics[width=.99\linewidth]{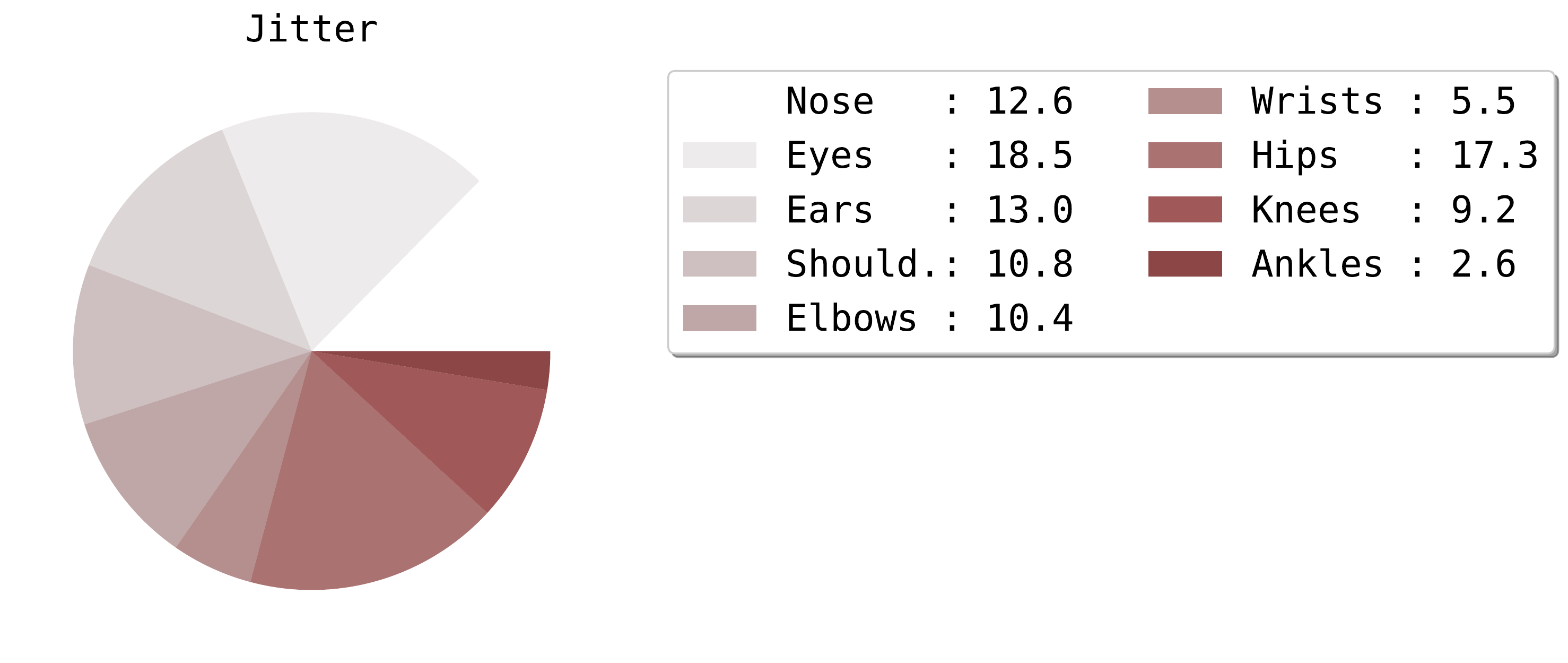}    \\
					& \includegraphics[width=.99\linewidth]{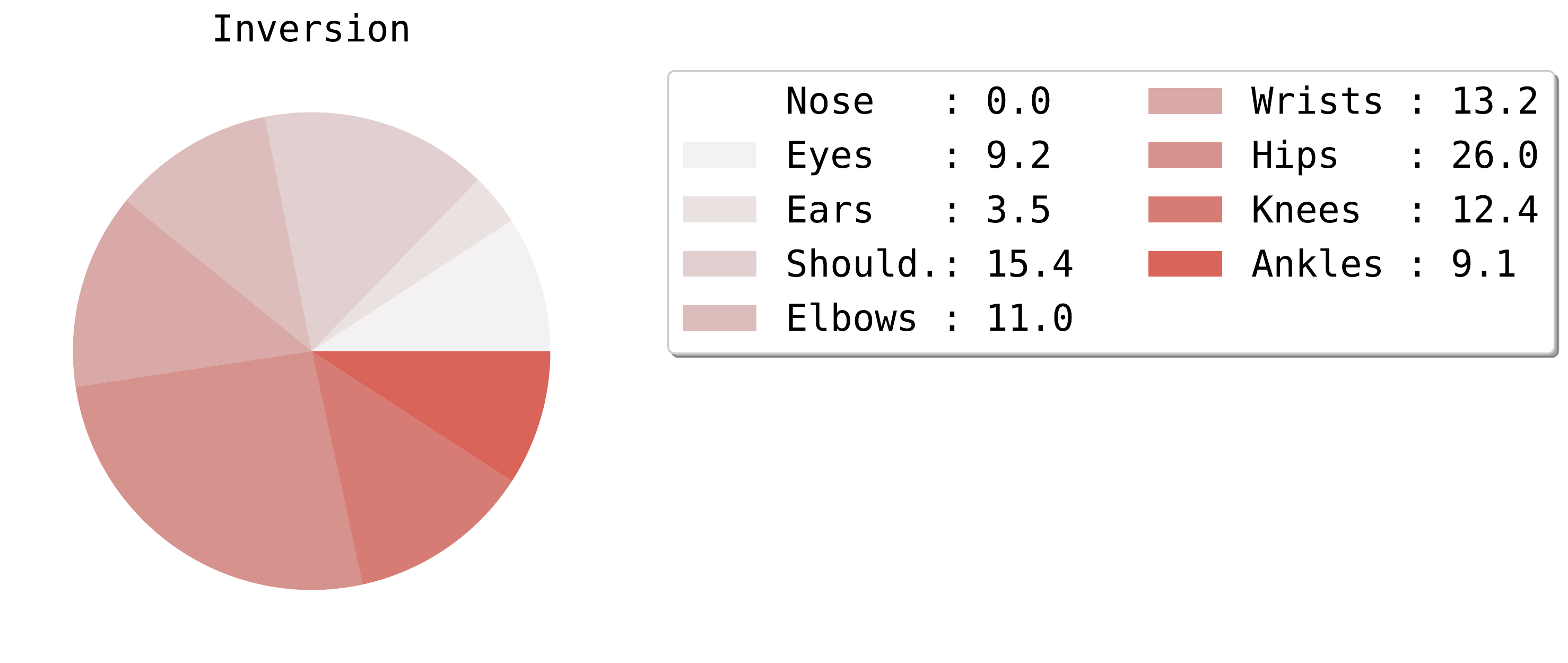}
					& \includegraphics[width=.99\linewidth]{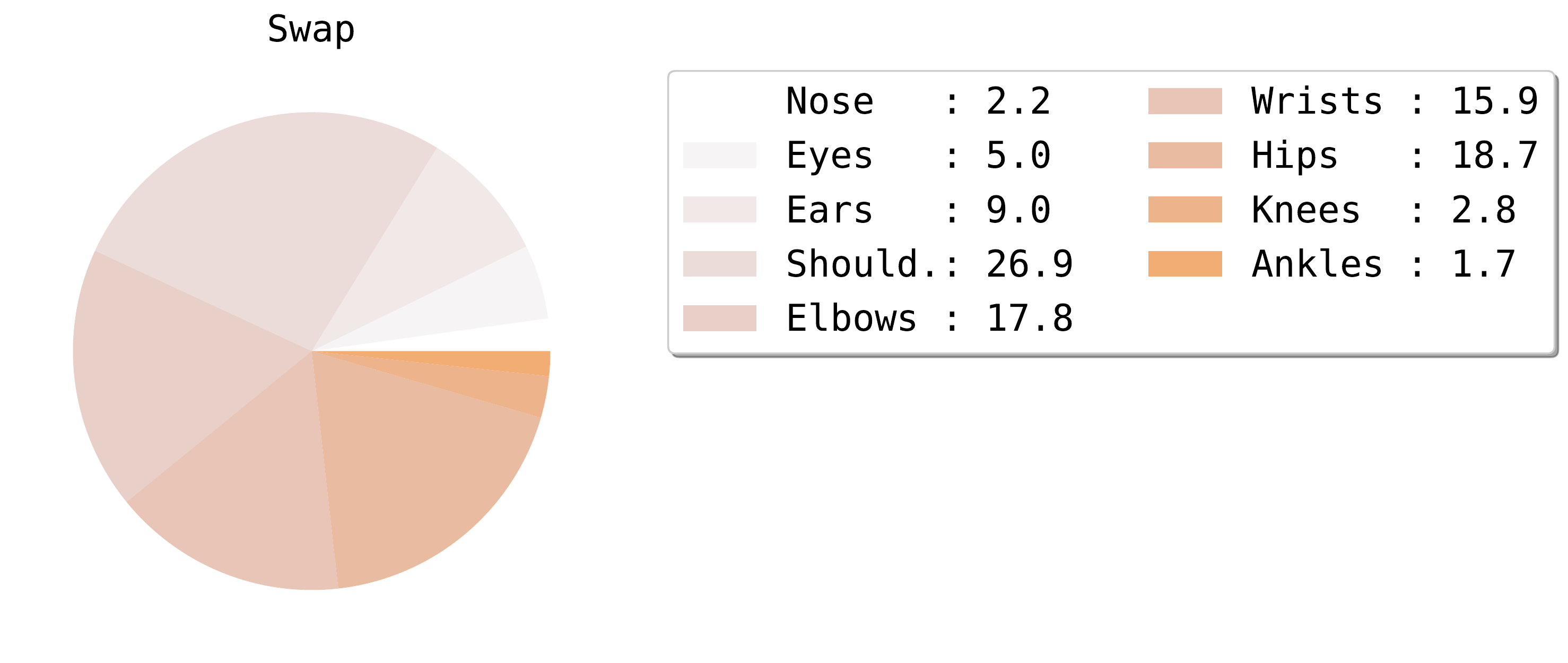}      \\
			\end{tabular}
		}
		\caption{{\small\emph{source-only}}}
	\end{subfigure}
	\begin{subfigure}[t]{0.8\textwidth}
		\centering
		\resizebox{\linewidth}{!}{
			\begin{tabular}{ccc}
				\raisebox{15mm}{\multirow{2}{*}{\includegraphics[width=.4\linewidth]{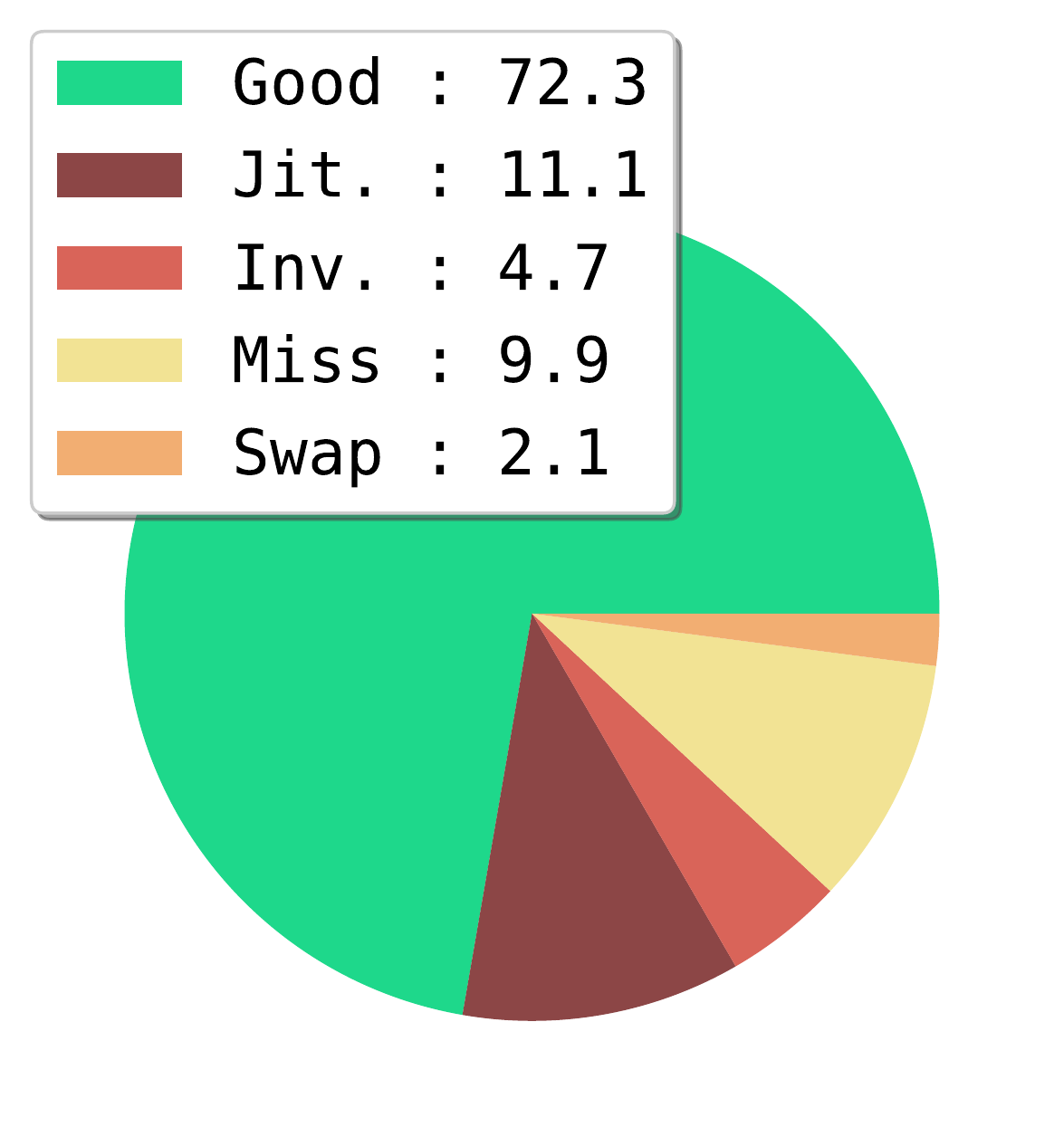}}}
					& \includegraphics[width=.99\linewidth]{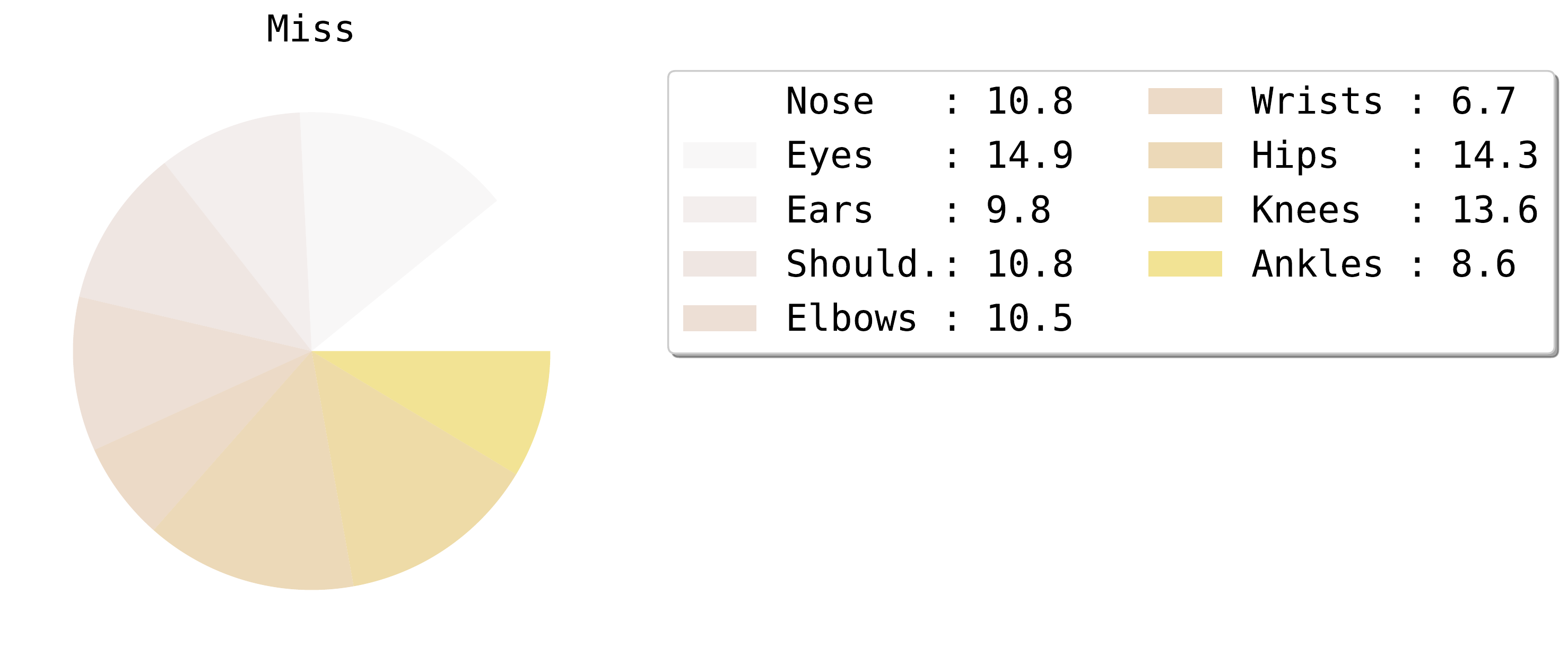}
					& \includegraphics[width=.99\linewidth]{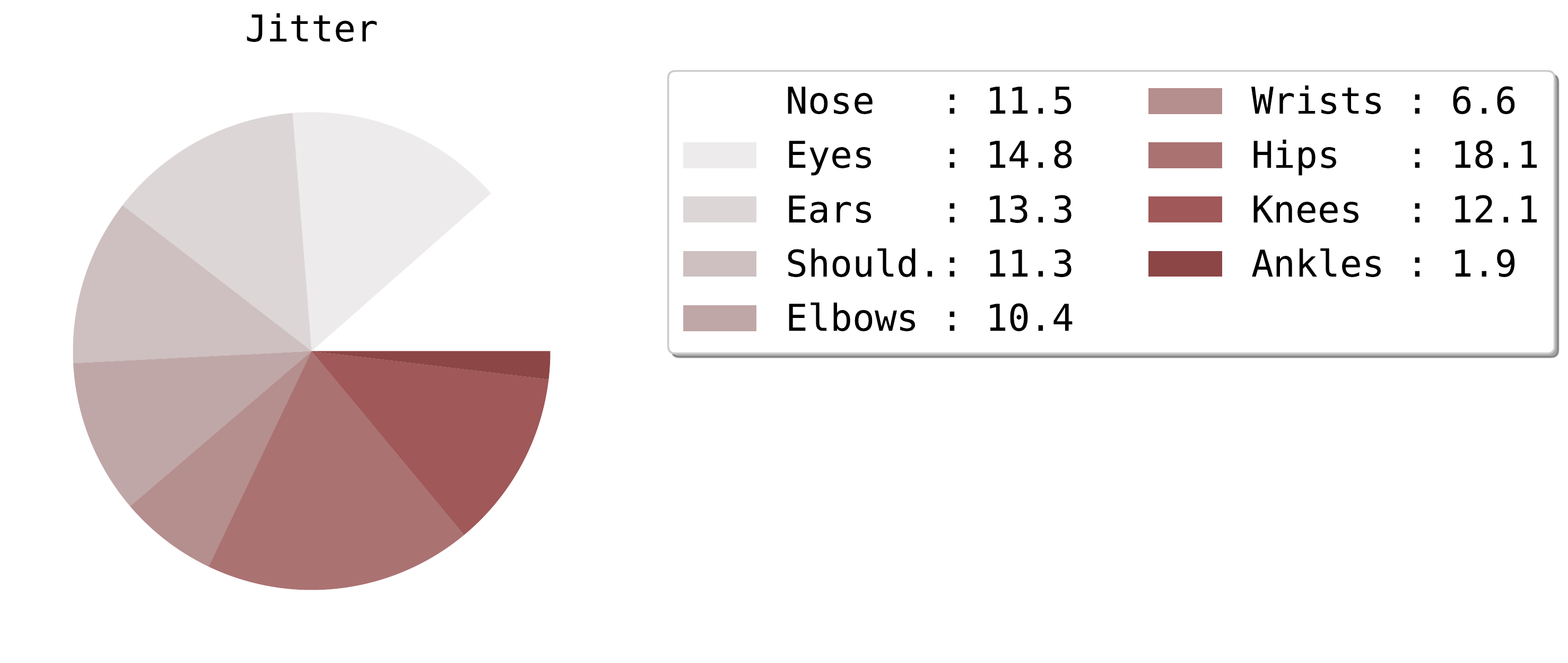}    \\
					& \includegraphics[width=.99\linewidth]{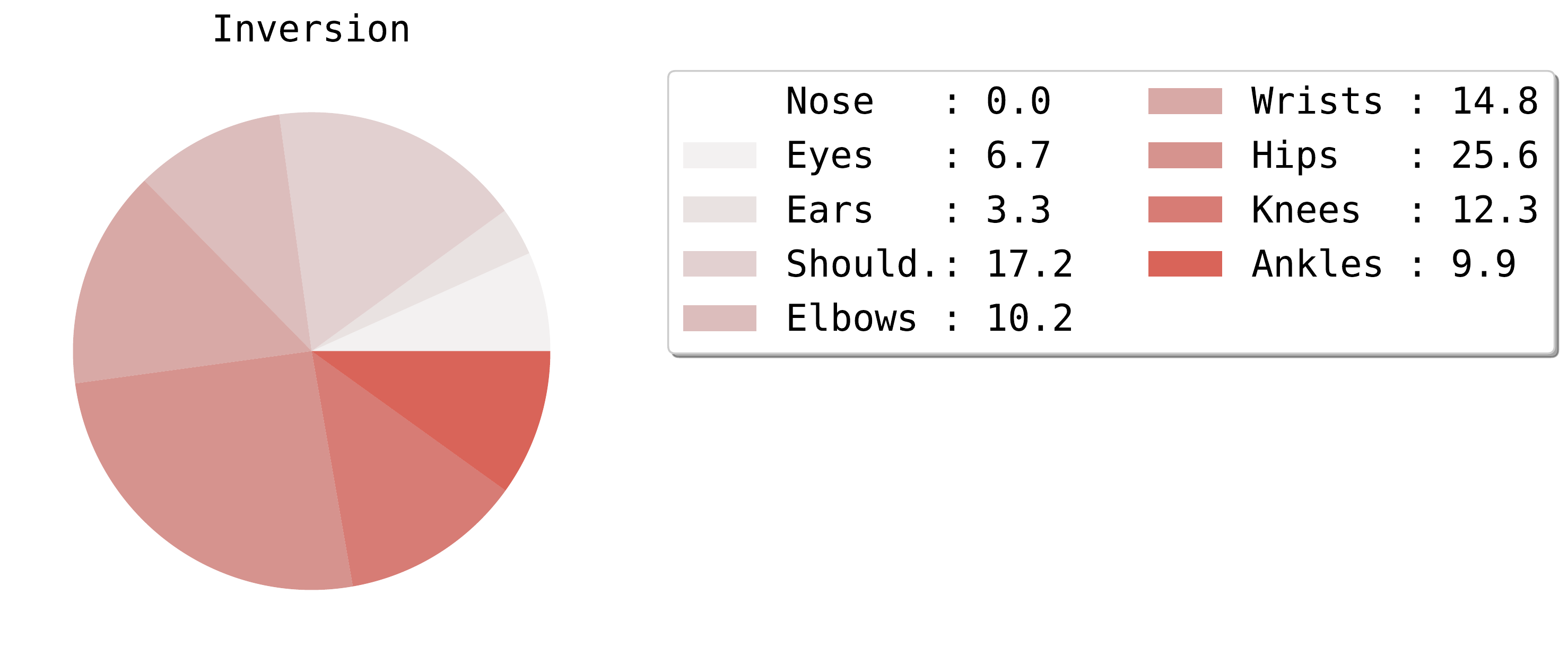}
					& \includegraphics[width=.99\linewidth]{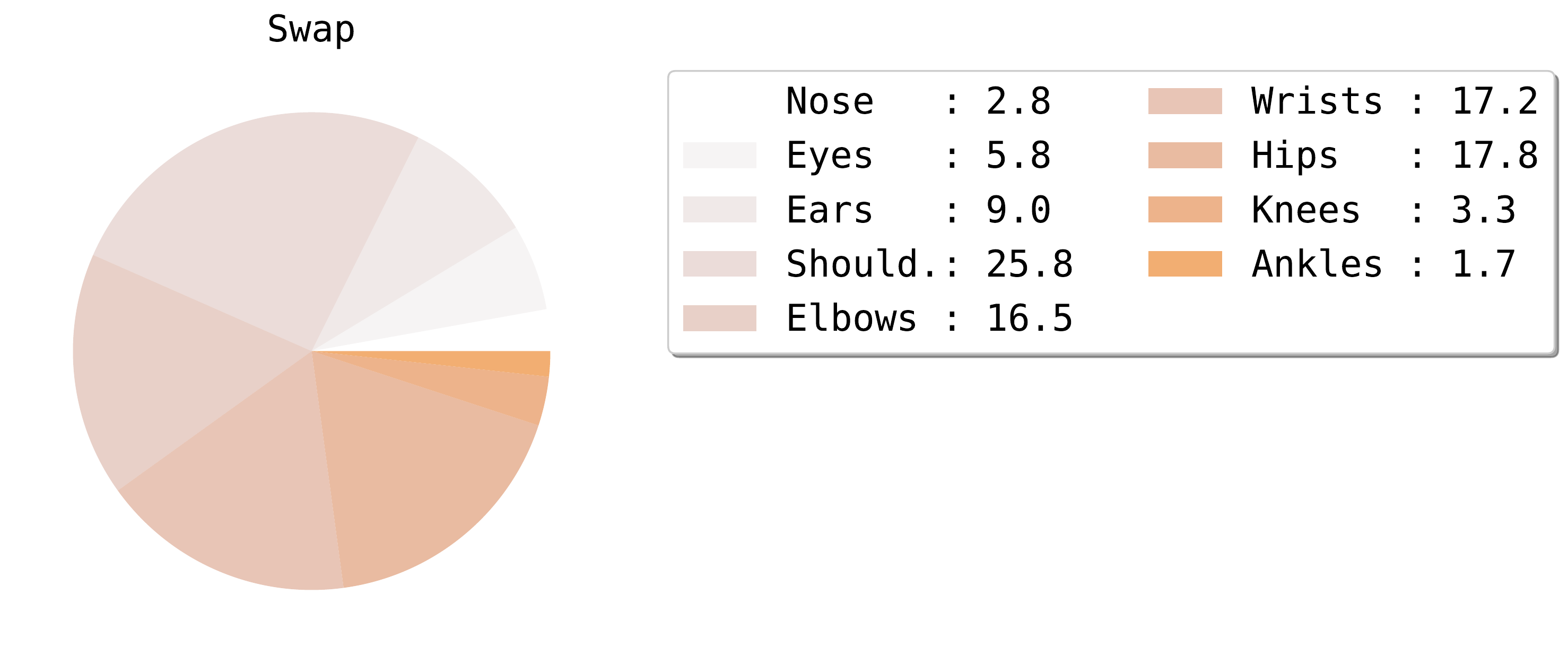}      \\
			\end{tabular}
		}
		\caption{{\small\emph{AdaptOR}}}
	\end{subfigure}
	\caption{\small{{\blue Localization errors at individual keypoint level for the pose estimation task before and after the domain adaptation. ``Jitter'', ``Inversion'', ``Swap'', and ``Miss'' are various localization errors defined in \citep{ruggero2017benchmarking}: ``Jitter'' error is the error in predicted keypoint w.r.t close proximity of the correct ground truth, ``Inversion'' error is due to the right-left swap of the body part, ``Swap'' is the error in assigning predicted keypoint to a wrong person, and ``Miss'' error is due to completely missing the correct ground truth location. We use the author's code repository \citep{ruggero2017benchmarking}\protect\footnotemark ~ for plotting the results.}}}	
	
	\label{figure:result-analysis}
\end{figure*}
\footnotetext{\url{https://github.com/matteorr/coco-analyze}}

\begin{figure*}[t!]
	%\hfill
	\centering
	\includegraphics[width=.99\linewidth]{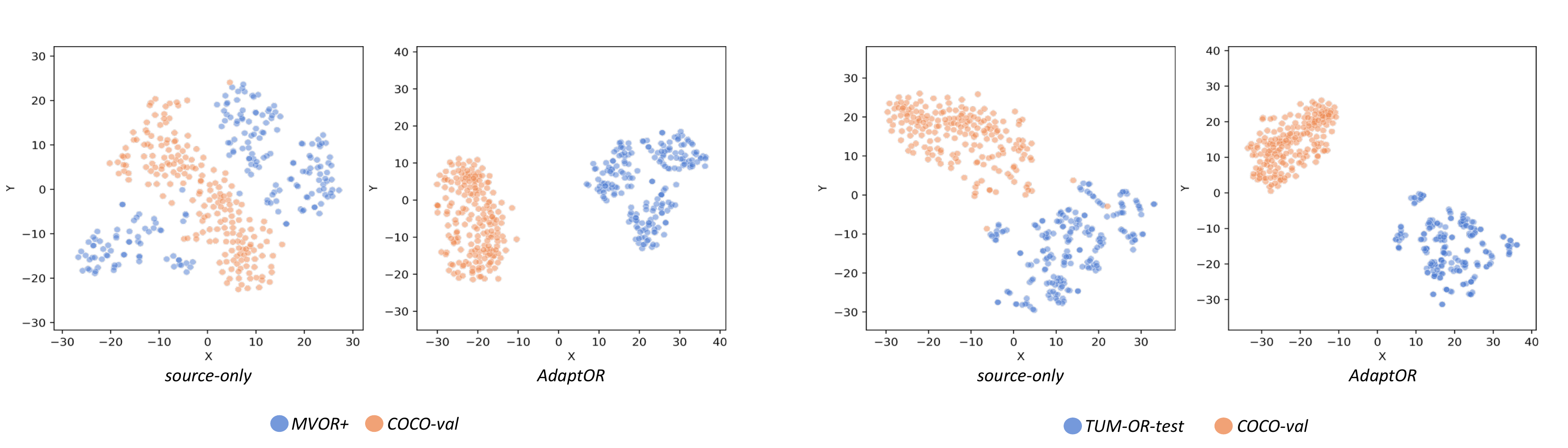}
	\caption{\small{{\blue t-sne feature visualization \citep{van2008visualizing} of the \emph{layer5} resnet features of the backbone model on random 200 images of the source and the target domain test datasets. The \emph{source-only} model uses only the \emph{GN(S)} layers whereas the \emph{AdapOR} uses separate \emph{GN(S)} and \emph{GN(T)} layers for the source and the target domain images, respectively. The \emph{AdapOR} model appropriately segregates the source and the target domain image features from the two domains helping in improving the domain adaptation for the downstream heads.}}}
	\label{figure:dfn-comp}
\end{figure*}

\section{Results}
\subsection{Source domain fully supervised training}
Table \ref{table:source} shows the results of \emph{kmrcnn} and \emph{kmrcnn+} models trained on the source domain COCO dataset. The \emph{kmrcnn} trained using self-supervised MoCo-v2 weights with Cross-GPU BN \citep{peng2018megdet} obtains improvement of approximately 1\% in all the three metrics compared to supervised ImageNet weights using frozen BN. The \emph{kmrcnn+} using GN performs equally well but with less training time. The \emph{kmrcnn+} model is therefore further used in the SSL experiments and extended in UDA experiments.

\subsection{AdaptOR: unsupervised domain adaptation (UDA) on target domains}
Table \ref{table:source-target-da} and figure \ref{fig:results_graphs} show the result of our unsupervised domain adaptation experiments using \emph{AdaptOR}. The first and the second half in table \ref{table:source-target-da} show the results for \emph{MVOR+} and \emph{TUM-OR-test} datasets, respectively. We evaluate the models at four downsampling scales (1x, 8x, 10x, and 12x). As the model is trained on unlabeled image sizes from 480 to 800 pixels (shorter side), we evaluate the model on nine target resolutions (480, 520, 560, 600, 640, 680, 720, 760, and 800), i.e., for a given downsampling scale, we down-sample the image with the scale and up-sample it to the given target resolution. The target resolution also corresponds to the shorter size of the image to maintain the aspect ratio. We use bilinear interpolation for the downsampling and up-sampling. The results in Table \ref{table:source-target-da} show the mean and standard deviation of the results computed on all the target resolutions for bounding box detection \emph{$AP_{person}^{bb}$}, pose estimation \emph{$AP_{person}^{kp}$}, and instance segmentation \emph{$AP_{person}^{bb}$ (from mask)} on a given downsampling scale.

The first row shows the \emph{source-only} results for the \emph{kmrcnn+} model trained on source domain images and evaluated on the target domain. The significant decrease in the low-resolution results of the  \emph{kmrcnn+} is likely because such heavily downsampled images are not present in the source domain. The improved result for the KM-DDS approach compared to KM-PL shows the effects of generating pseudo labels using the multi-scale and flipping transformation. The bounding box and segmentation results for the KM-ORPose are slightly worse than the KM-PL and KM-DDS. It may be because KM-ORPose uses a state-of-the-art object detector trained on all the 80 class categories from COCO whereas, KM-PL and KM-DDS use the model trained specifically for the person class. The \emph{AdaptOR} performs significantly better compared to baseline approaches, especially on the low-resolution at different target resolutions, see figure \ref{fig:results_graphs}, suggesting the potential of our approach for low-resolution images in the privacy-sensitive OR environment. We observe a slight decrease in the accuracy for \emph{$AP_{person}^{kp}$} metric on original size, likely due to the use of the multi-stage complex teacher model to generate the pseudo poses. Instead, our approach improves the given model in a model agnostic way without relying on an external teacher model to generate the pseudo labels. We also plot the results at individual scales in the figure \ref{fig:results_graphs}. The figure \ref{tab:table-qual1} and \ref{tab:table-qual-tum-or} show qualitative results comparing our approach with the baseline approaches. 

{\blue We further analyze the impact of different localization errors at the keypoint level before and after the domain adaptation using an approach described in \citep{ruggero2017benchmarking}. As shown in Fig. ~\ref{figure:result-analysis}, after domain adaptation, our approach correctly detects more keypoints while reducing the impact of different localization errors.} Additional qualitative results for the UDA experiments on \emph{MVOR+} and \emph{TUM-OR-test} are presented in the supplementary video\footnote{\url{https://youtu.be/gqwPu9-nfGs}}

\subsubsection{Ablation experiments}
\emph{6.2.1.1 Disentangled feature normalization}\\
{\blue Fig.~\ref{figure:dfn-comp} shows t-sne feature visualization \citep{van2008visualizing} of the \emph{layer5} resnet features of the backbone model illustrating the appropriate segregation of the features after the domain adaptation}. We also conduct experiments to quantify the use of two separate GN layers, \emph{GN(S)} and  \emph{GN(T)}, in the feature extractor for domain-specific normalization compared to either a single GN layer or a single frozen BN layer. The first row in Table \ref{table:splitgn-design} shows the results for the \emph{krcnn} \citep{he2017mask,wu2019detectron2kpn} model using frozen BN \citep{he2016deep} layers for joint bounding box detection and pose estimation. We take the source domain COCO trained weights from detectron2 \citep{wu2019detectron2} library and train it on the MVOR dataset. The second row shows the results for the \emph{kmrcnn+} model using a single GN layer for both domains. We also evaluate \emph{kmrcnn++} where we use the GN layers corresponding to source domain \emph{GN(S)} to evaluate on the target domain (\emph{kmrcnn++ GN(S)}). We obtain significantly better results by using our design of the two separate GN layers for feature normalization.\\
\begin{table}[t!]
	\centering
	\caption{\small{Ablation study comparing the \emph{kmrcnn++} model using the two GN layer-based design for feature normalization with the \emph{kmrcnn+} that uses only a single layer. We also compare it with a \emph{krcnn} model using single frozen BN, and \emph{kmrcnn++ GN(S)}, the same \emph{kmrcnn++} model but using the GN layers corresponding to the source domain.}}
	\vspace{-2mm}
	\scalebox{0.78}{
		\begin{tabular}{l|cccc}
			\toprule
			\multirow{3}{*}{Models} & \multicolumn{4}{c}{\textbf{\emph{MVOR+}}} \Tstrut \Bstrut                                                                                                                                               \\
			\cline{2-5}
			                        & \textbf{~1x~}                                                                                         & \textbf{~8x~}           & \textbf{~10x~}          & \textbf{~12x~}  \Tstrut                     \\
			\cline{2-5}
			                        & \multicolumn{4}{c}{$\mathit{AP_{person}^{bb}}$ (mean$\pm$std)}   \Tstrut \Bstrut                                                                                                               \\
			\hline
			\emph{krcnn}            & 59.00$\pm$0.35                                                                                        & 56.78$\pm$0.37          & 55.87$\pm$0.34          & 54.43$\pm$0.36 \Tstrut                      \\
			\emph{kmrcnn+}          & 60.71$\pm$0.16                                                                                        & 58.75$\pm$0.33          & 58.03$\pm$0.31          & 56.97$\pm$0.39 \Tstrut                      \\
			\emph{kmrcnn++ GN(S)}   & 59.64$\pm$0.46                                                                                        & 55.86$\pm$0.48          & 53.84$\pm$0.64          & 51.61$\pm$0.74 \Tstrut                      \\\hline
			\emph{kmrcnn++}         & \textbf{61.41$\pm$0.40}                                                                               & \textbf{59.48$\pm$0.35} & \textbf{58.55$\pm$0.36} & \textbf{57.33$\pm$0.43} \Tstrut
			\\\hline
			                        & \multicolumn{4}{c}{$\mathit{AP_{person}^{kp}}$ (mean$\pm$std)} \Tstrut \Bstrut                                                                                                                 \\
			\hline
			\emph{krcnn}            & 57.96$\pm$0.32                                                                                        & 55.48$\pm$0.62          & 53.34$\pm$0.55          & 50.50$\pm$0.44  \Tstrut                     \\
			\emph{kmrcnn+}          & 47.15$\pm$0.30                                                                                        & 45.27$\pm$0.44          & 43.89$\pm$0.44          & 42.01$\pm$0.44                      \Tstrut \\
			\emph{kmrcnn++ GN(S)}   & 58.64$\pm$0.40                                                                                        & 52.37$\pm$0.41          & 49.51$\pm$0.46          & 46.08$\pm$0.51 \Tstrut                      \\\hline
			\emph{kmrcnn++}         & \textbf{60.86$\pm$0.38}                                                                               & \textbf{57.35$\pm$0.61} & \textbf{55.42$\pm$0.66} & \textbf{52.60$\pm$0.60} \Tstrut             \\
			\hline
			                        & \multicolumn{4}{c}{$\mathit{AP_{person}^{bb\ (from\ mask)}}$ (mean$\pm$std)} \Tstrut \Bstrut                                                                                                   \\
			\hline
			\emph{krcnn}            & -                                                                                                     & -                       & -                       & - \Tstrut                                   \\
			\emph{kmrcnn+}          & 55.18$\pm$0.25                                                                                        & 53.85$\pm$0.42          & 53.28$\pm$0.5           & 52.44$\pm$0.58 \Tstrut                      \\
			\emph{kmrcnn++ GN(S)}   & 58.22$\pm$0.46                                                                                        & 54.77$\pm$0.62          & 53.06$\pm$0.67          & 50.70$\pm$0.72 \Tstrut                      \\\hline
			\emph{kmrcnn++}         & \textbf{59.34$\pm$0.40}                                                                               & \textbf{57.44$\pm$0.42} & \textbf{56.62$\pm$0.41} & \textbf{55.39$\pm$0.51} \Tstrut             \\
			\hline
		\end{tabular}
	}
	\label{table:splitgn-design}
	\vspace{-1mm}
\end{table}
\begin{table}[t!]
	\centering
	\caption{\small{Ablation study quantifying the different augmentations on the strongly transformed image used by the student model for the training. Here, sr: \emph{strong-resize}, ra: random-augment, rc: random-cut, and geom: geometric transformations consisting of random-resize and random-flip.}}
	\vspace{-2mm}
	\scalebox{0.67}{
		\begin{tabular}{lccc|cccc}
			\toprule
			           &            &            &            & \multicolumn{4}{c}{\textbf{\emph{MVOR+}}} \Tstrut \Bstrut \Tstrut \Bstrut                                                                                                                    \\
			\cline{5-8}
			sr         & ra         & rc         & geom       & \textbf{~1x~}                                                                                         & \textbf{~8x~}           & \textbf{~10x~}          & \textbf{~12x~}  \Tstrut          \\
			\cline{5-8}
			           &            &            &            & \multicolumn{4}{c}{$\mathit{AP_{person}^{bb}}$ (mean$\pm$std)}   \Tstrut \Bstrut                                                                                                    \\
			\hline
			           &            & Baseline   &            & 56.61$\pm$0.34                                                                                        & 40.42$\pm$2.17          & 34.87$\pm$2.47          & 29.61$\pm$2.69 \Tstrut           \\\hline
			\xmark     & \xmark     & \xmark     & \xmark     & 58.06$\pm$0.28                                                                                        & 45.14$\pm$1.70          & 40.19$\pm$2.09          & 35.45$\pm$2.28 \Tstrut           \\
			\checkmark & \xmark     & \xmark     & \xmark     & 58.34$\pm$0.34                                                                                        & 58.03$\pm$0.31          & 57.25$\pm$0.33          & 55.97$\pm$0.33 \Tstrut           \\
			\checkmark & \xmark     & \xmark     & \checkmark & 59.64$\pm$0.34                                                                                        & 58.74$\pm$0.30          & 58.01$\pm$0.36          & 56.80$\pm$0.32 \Tstrut           \\
			\checkmark & \checkmark & \xmark     & \xmark     & 58.43$\pm$0.31                                                                                        & 57.72$\pm$0.31          & 56.99$\pm$0.33          & 55.58$\pm$0.29 \Tstrut           \\
			\checkmark & \checkmark & \checkmark & \xmark     & 59.79$\pm$0.54                                                                                        & 58.38$\pm$0.44          & 57.48$\pm$0.45          & 56.21$\pm$0.46 \Tstrut           \\\hline
			\checkmark & \checkmark & \checkmark & \checkmark & \textbf{61.41$\pm$0.40}                                                                               & \textbf{59.48$\pm$0.35} & \textbf{58.55$\pm$0.36} & \textbf{57.33$\pm$0.43}  \Tstrut
			\\\hline
			           &            &            &            & \multicolumn{4}{c}{$\mathit{AP_{person}^{kp}}$ (mean$\pm$std)} \Tstrut \Bstrut                                                                                                      \\
			\hline
			           &            & Baseline   &            & 50.55$\pm$0.39                                                                                        & 23.99$\pm$2.25          & 16.86$\pm$2.16          & 11.31$\pm$1.91 \Tstrut           \\\hline
			\xmark     & \xmark     & \xmark     & \xmark     & 52.32$\pm$0.30                                                                                        & 31.33$\pm$1.56          & 24.48$\pm$2.07          & 18.19$\pm$1.97 \Tstrut           \\
			\checkmark & \xmark     & \xmark     & \xmark     & 54.22$\pm$0.39                                                                                        & 53.53$\pm$0.63          & 51.65$\pm$0.58          & 49.13$\pm$0.58 \Tstrut           \\
			\checkmark & \xmark     & \xmark     & \checkmark & 57.07$\pm$0.31                                                                                        & 55.41$\pm$0.62          & 53.68$\pm$0.55          & 51.19$\pm$0.48 \Tstrut           \\
			\checkmark & \checkmark & \xmark     & \xmark     & 54.51$\pm$0.24                                                                                        & 52.67$\pm$0.62          & 50.74$\pm$0.68          & 47.97$\pm$0.50 \Tstrut           \\
			\checkmark & \checkmark & \checkmark & \xmark     & 57.44$\pm$0.37                                                                                        & 54.73$\pm$0.47          & 52.64$\pm$0.47          & 49.96$\pm$0.49 \Tstrut           \\\hline
			\checkmark & \checkmark & \checkmark & \checkmark & \textbf{60.86$\pm$0.38}                                                                               & \textbf{57.35$\pm$0.61} & \textbf{55.42$\pm$0.66} & \textbf{52.60$\pm$0.60} \Tstrut
			\\\hline
			           &            &            &            & \multicolumn{4}{c}{$\mathit{AP_{person}^{bb\ (from\ mask)}}$ (mean$\pm$std)} \Tstrut \Bstrut                                                                                        \\
			\hline
			           &            & Baseline   &            & 54.95$\pm$0.37                                                                                        & 37.98$\pm$2.21          & 32.58$\pm$2.37          & 27.56$\pm$2.48  \Tstrut          \\\hline
			\xmark     & \xmark     & \xmark     & \xmark     & 56.08$\pm$0.32                                                                                        & 42.12$\pm$1.78          & 37.19$\pm$2.13          & 32.56$\pm$2.27 \Tstrut           \\
			\checkmark & \xmark     & \xmark     & \xmark     & 55.81$\pm$0.38                                                                                        & 55.66$\pm$0.46          & 54.94$\pm$0.43          & 53.73$\pm$0.51 \Tstrut           \\
			\checkmark & \xmark     & \xmark     & \checkmark & 57.14$\pm$0.35                                                                                        & 56.52$\pm$0.38          & 55.84$\pm$0.42          & 54.62$\pm$0.41 \Tstrut           \\
			\checkmark & \checkmark & \xmark     & \xmark     & 56.06$\pm$0.32                                                                                        & 55.50$\pm$0.33          & 54.70$\pm$0.41          & 53.30$\pm$0.40 \Tstrut           \\
			\checkmark & \checkmark & \checkmark & \xmark     & 57.58$\pm$0.50                                                                                        & 56.34$\pm$0.45          & 55.48$\pm$0.50          & 54.21$\pm$0.62 \Tstrut           \\\hline
			\checkmark & \checkmark & \checkmark & \checkmark & \textbf{59.34$\pm$0.40}                                                                               & \textbf{57.44$\pm$0.42} & \textbf{56.62$\pm$0.41} & \textbf{55.39$\pm$0.51} \Tstrut
			\\\hline
		\end{tabular}
	}
	\label{table:ablation}
	\vspace{-1mm}
\end{table}

\begin{figure}[t!]
	%\hfill
	\centering
	\includegraphics[width=.99\linewidth]{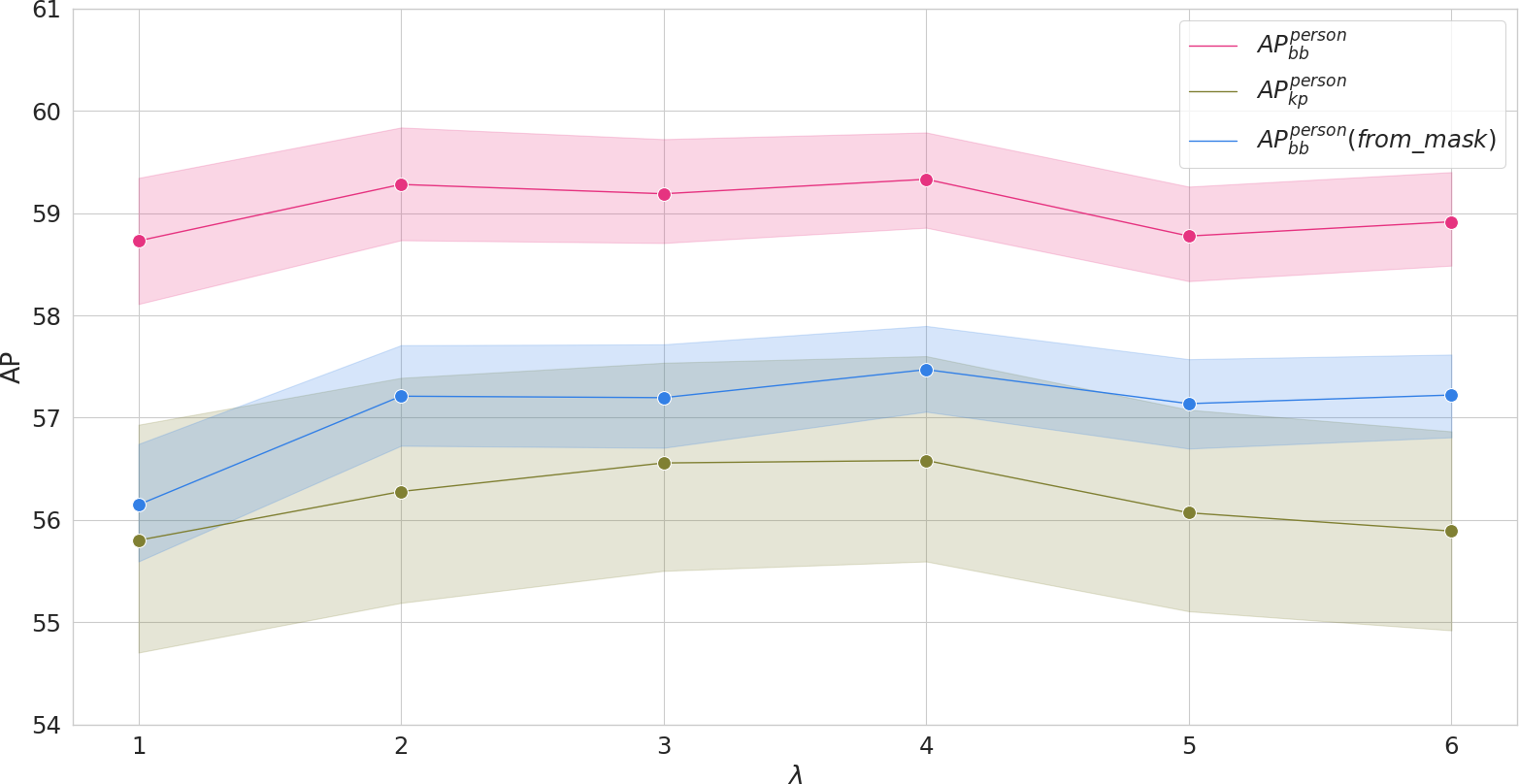}
	\caption{\small{{\blue Results for different values of unsupervised loss weight ($\lambda$) on the \emph{MVOR+} dataset. Results show the mean and confidence interval computed using different downsampling scales (1x, 8x, 10x, and 12x) and target resolutions (480, 520, 560, 600, 640, 680, 720, 760, and 800).}}}
	\label{figure:lambda-weights}
\end{figure}

\begin{table*}[t!]
	\centering
	\setlength{\fboxsep}{0pt}%
	\setlength{\fboxrule}{1.5pt}%
	\setlength\tabcolsep{0.5pt}%
	\vspace{-2mm}
	\scalebox{0.95}{
		\begin{tabular}{cc}
			\centering
			\fbox{\includegraphics[width=3.6in]{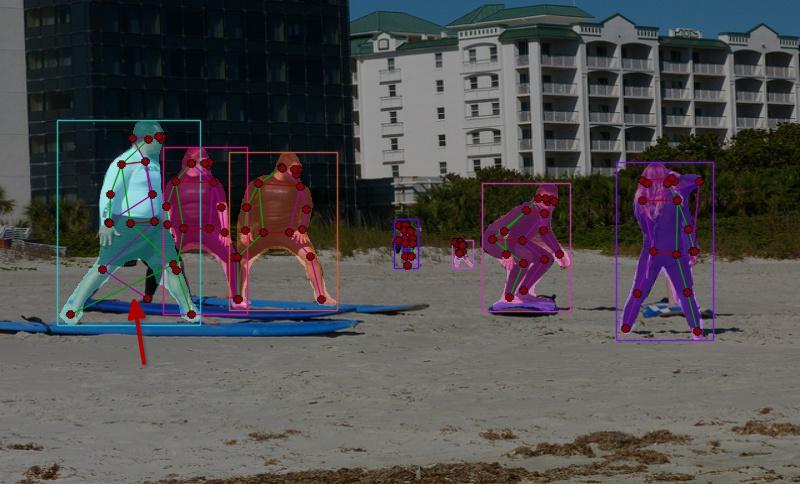}} & \fbox{\includegraphics[width=3.6in]{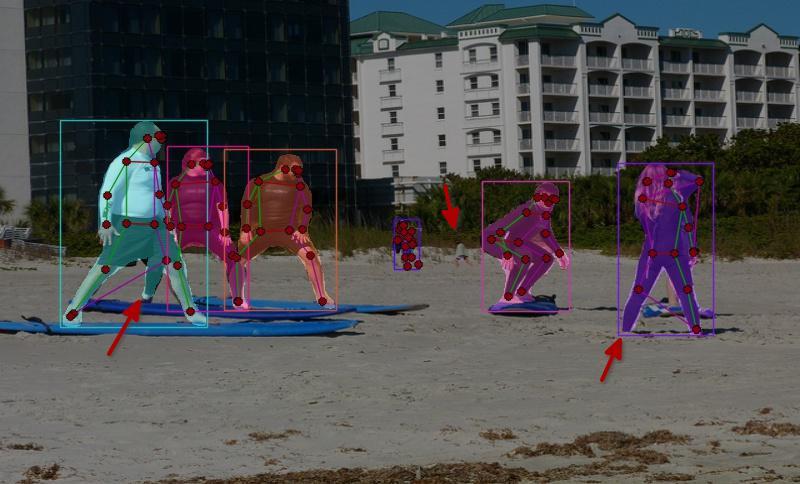}}   \\
			1\% supervision                                                            & 2\% supervision                                                              \\
			\fbox{\includegraphics[width=3.6in]{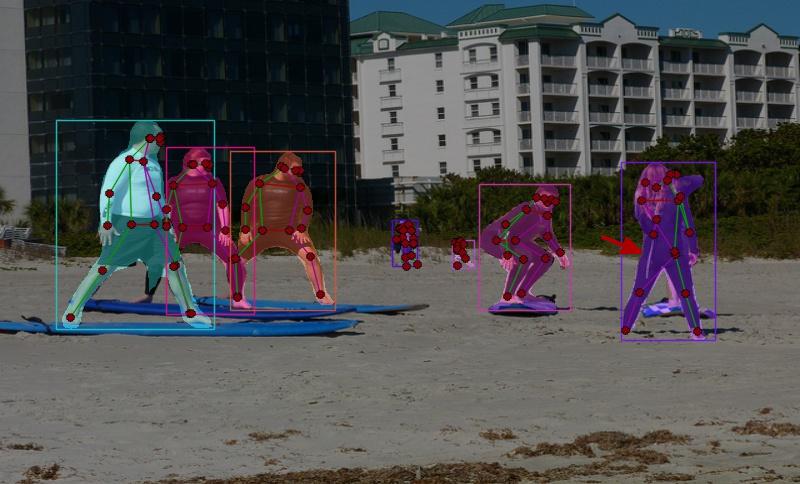}} & \fbox{\includegraphics[width=3.6in]{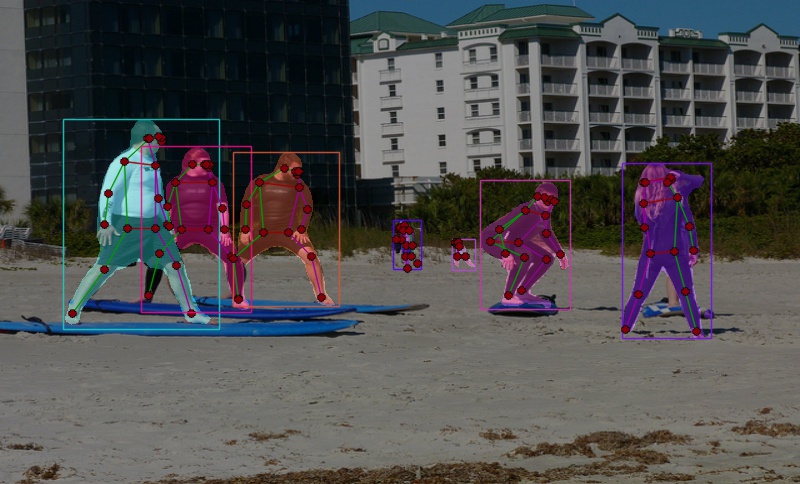}} \\
			5\% supervision                                                            & 100\% supervision                                                            \\
		\end{tabular}
	}
	\captionof{figure}[]{\small{Qualitative results on a sample image from \emph{COCO-val} dataset with x\%(x=1,2,5,100) of labeled supervision. We use the \emph{AdaptOR-SSL} for 1\%, and 5\% labeled supervision with the rest of the data as the unlabeled data. We see comparable qualitative results with 1\% of labeled supervision to 100\% of labeled supervision. The red arrows show either missed detections or localization errors.}}
	\vspace{-1mm}
	\label{tab:table-qual3}
\end{table*}

\begin{table*}[htb!]
	\centering
	\caption{\small{Results for \emph{AdaptOR-SSL} on \emph{COCO-val} dataset under the semi-supervised learning setting with x\%(x=1,2,5,10) of labeled supervision. We compare it with the fully supervised baselines trained on the same labeled data without using any unlabeled data. The \emph{supervised} baseline uses only the random resize and random-flip data augmentations as used in \citep{he2017mask} whereas \emph{supervised++} uses the same data augmentation pipeline as in \emph{AdaptOR-SSL} containing \emph{weakly} and \emph{strongly} augmented labeled images. We also compare it with the current state-of-the-art SSL object detector Unbiased-Teacher \citep{liu2021unbiased} for the person bounding box detection task. The inference is performed on a single scale of 800 pixels (shorter side) following the same settings as used in \citep{he2017mask, liu2021unbiased}.}}
	\vspace{-2mm}
	\scalebox{0.94}{
		\begin{tabular}{l|cccc|cccc|cccc}
			\toprule
			\multirow{2}{*}{Methods}    & \multicolumn{4}{c|}{$\mathit{AP_{person}^{bb}}$} \Tstrut \Bstrut & \multicolumn{4}{c|}{$\mathit{AP_{person}^{kp}}$} \Tstrut \Bstrut & \multicolumn{4}{c}{$\mathit{AP_{person}^{mask}}$} \Tstrut \Bstrut                                                                                                                                                                  \\
			\cline{2-13}
			                            & ~1\%~                                                            & ~2\%~                                                            & ~5\%~                                                             & ~10\%~         & ~1\%~          & ~2\%~          & ~5\%~          & ~10\%~         & ~1\%~          & ~2\%~          & ~5\%~          & ~10\%~ \Tstrut         \\
			\hline
			\emph{supervised}           & 22.09                                                            & 28.43                                                            & 34.52                                                             & 37.88          & 15.91          & 23.58          & 30.96          & 37.77          & 18.13          & 23.93          & 29.34          & 33.47  \Tstrut         \\
			\emph{supervised++}         & 28.59                                                            & 34.27                                                            & 41.18                                                             & 43.60          & 25.78          & 32.14          & 41.45          & 46.51          & 24.18          & 29.14          & 35.40          & 37.83  \Tstrut         \\\hline
			Unbiased-Teacher            & 39.18                                                            & 40.76                                                            & 43.72                                                             & 46.64          & -              & -              & -              & -              & -              & -              & -              & - \Tstrut              \\
			\emph{\textbf{AdaptOR-SSL}} & \textbf{42.57}                                                   & \textbf{45.37}                                                   & \textbf{49.90}                                                    & \textbf{52.70} & \textbf{38.22} & \textbf{44.08} & \textbf{49.79} & \textbf{56.65} & \textbf{36.06} & \textbf{38.96} & \textbf{43.10} & \textbf{45.46} \Tstrut \\ \hline
			\bottomrule
		\end{tabular}
	}
	\label{table:source:ssl}
	\vspace{-1mm}
\end{table*}

\begin{table}[bt!]
	\centering
	\caption{\small{{\blue Performance comparison when applying \emph{AdaptOR-SSL} models trained with 1\%, 2\%, 5\%, and 10\% source domain labels to the target domain of \emph{MVOR+} (see ``Before UDA'' results). When we apply the \emph{AdaptOR} approach on the \emph{AdaptOR-SSL} model (trained using 10\% source domain labels), we observe an improvement in the performance (see ``After UDA'' results). Results corresponding to 100\% source domain labeled supervision in ``Before UDA'' and ``After UDA'' are obtained from Table \ref{table:source} and \ref{table:source-target-da}, respectively.}}}
	\vspace{-2mm}
	\scalebox{0.78}{
		\begin{tabular}{l|cccc}
			\toprule
			\multirow{3}{*}{models} & \multicolumn{4}{c}{\textbf{\emph{MVOR+}}} \Tstrut \Bstrut                                                                                                         \\
			\cline{2-5}
			                        & \textbf{~1x~}                                                                                         & \textbf{~8x~}  & \textbf{~10x~} & \textbf{~12x~}  \Tstrut \\
			\cline{2-5}
			                        & \multicolumn{4}{c}{$\mathit{AP_{person}^{bb}}$ (mean$\pm$std)} \Tstrut \Bstrut                                                                           \\
			\hline
			Before UDA              &                                                                                                       &                & \Tstrut                                  \\
			\emph{1\%}              & 48.33$\pm$0.67                                                                                        & 39.89$\pm$1.97 & 34.46$\pm$2.44 & 28.63$\pm$3.25  \Tstrut \\
			\emph{2\%}              & 48.28$\pm$0.64                                                                                        & 41.12$\pm$2.00 & 35.93$\pm$2.16 & 30.51$\pm$2.54 \Tstrut  \\
			\emph{5\%}              & 51.27$\pm$0.48                                                                                        & 43.11$\pm$2.08 & 37.95$\pm$2.14 & 31.75$\pm$2.72 \Tstrut  \\
			\emph{10\%}             & 53.95$\pm$0.65                                                                                        & 44.74$\pm$1.92 & 39.83$\pm$2.03 & 34.13$\pm$2.60 \Tstrut  \\
			\emph{100\%}            & 56.61$\pm$0.34                                                                                        & 40.42$\pm$2.17 & 34.87$\pm$2.47 & 29.61$\pm$2.69 \Tstrut  \\\hline
			After UDA               &                                                                                                       &                & \Tstrut                                  \\
			\emph{10\%}             & 57.58$\pm$0.56                                                                                        & 55.80$\pm$0.60 & 54.70$\pm$0.51 & 53.44$\pm$0.38 \Tstrut  \\
			\emph{100\%}            & 61.41$\pm$0.40                                                                                        & 59.48$\pm$0.35 & 58.55$\pm$0.36 & 57.33$\pm$0.43 \Tstrut  \\
			\hline
			                        & \multicolumn{4}{c}{$\mathit{AP_{person}^{kp}}$ (mean$\pm$std)} \Tstrut \Bstrut                                                                           \\
			\hline
			Before UDA              &                                                                                                       &                & \Tstrut                                  \\
			\emph{1\%}              & 25.28$\pm$1.06                                                                                        & 16.64$\pm$1.17 & 12.72$\pm$1.90 & 08.34$\pm$1.94 \Tstrut  \\
			\emph{2\%}              & 30.16$\pm$0.58                                                                                        & 21.44$\pm$1.91 & 16.28$\pm$2.22 & 11.35$\pm$2.39 \Tstrut  \\
			\emph{5\%}              & 37.09$\pm$0.30                                                                                        & 25.93$\pm$2.22 & 20.12$\pm$2.38 & 13.84$\pm$2.50 \Tstrut  \\
			\emph{10\%}             & 41.51$\pm$0.58                                                                                        & 28.57$\pm$1.88 & 22.57$\pm$2.15 & 16.17$\pm$2.38 \Tstrut  \\
			\emph{100\%}            & 50.55$\pm$0.39                                                                                        & 23.99$\pm$2.25 & 16.86$\pm$2.16 & 11.31$\pm$1.91 \Tstrut  \\\hline
			After UDA               &                                                                                                       &                & \Tstrut                                  \\
			\emph{10\%}             & 48.52$\pm$0.50                                                                                        & 45.73$\pm$0.56 & 43.74$\pm$0.47 & 40.90$\pm$0.44 \Tstrut  \\
			\emph{100\%}            & 60.86$\pm$0.38                                                                                        & 57.35$\pm$0.61 & 55.42$\pm$0.66 & 52.60$\pm$0.60 \Tstrut  \\
			\hline
			                        & \multicolumn{4}{c}{$\mathit{AP_{person}^{bb\ (from\ mask)}}$ (mean$\pm$std)} \Tstrut \Bstrut                                                             \\
			\hline
			Before UDA              &                                                                                                       &                & \Tstrut                                  \\
			\emph{1\%}              & 47.54$\pm$0.78                                                                                        & 38.37$\pm$2.32 & 32.44$\pm$2.73 & 26.32$\pm$3.42 \Tstrut  \\
			\emph{2\%}              & 47.96$\pm$0.90                                                                                        & 39.32$\pm$2.29 & 33.54$\pm$2.30 & 27.87$\pm$2.44 \Tstrut  \\
			\emph{5\%}              & 50.55$\pm$0.74                                                                                        & 41.09$\pm$2.30 & 35.68$\pm$2.16 & 29.45$\pm$2.69 \Tstrut  \\
			\emph{10\%}             & 52.79$\pm$0.69                                                                                        & 42.63$\pm$2.17 & 37.18$\pm$2.10 & 31.41$\pm$2.60 \Tstrut  \\
			\emph{100\%}            & 54.95$\pm$0.37                                                                                        & 37.98$\pm$2.21 & 32.58$\pm$2.37 & 27.56$\pm$2.48 \Tstrut  \\\hline
			After UDA               &                                                                                                       &                & \Tstrut                                  \\
			\emph{10\%}             & 55.60$\pm$0.52                                                                                        & 54.07$\pm$0.58 & 53.00$\pm$0.49 & 51.55$\pm$0.35 \Tstrut  \\
			\emph{100\%}            & 59.34$\pm$0.40                                                                                        & 57.44$\pm$0.42 & 56.62$\pm$0.41 & 55.39$\pm$0.51 \Tstrut  \\
			\hline
		\end{tabular}
	}
	\label{table:init-exps}
	\vspace{-1mm}
\end{table}

\emph{6.2.1.2 Components of AdaptOR}\\
Table \ref{table:ablation} shows the ablation experiments to see the effect of using different types of augmentations on the strongly transformed images used by the student model during training. The results show that the \emph{strong-resize} augmentations are needed to adapt the model to the low-resolution OR images. The geometric transform exploiting the \emph{transformation equivariance constraints} significantly improves the results, especially for the pose estimation task, where we also utilize the chirality transforms to map the flipped keypoints to the horizontally flipped image. The results are further improved using the random-augment and random-cut augmentations.

{\blue \emph{6.2.1.3 Effect of unsupervised loss weight ($\lambda$) values}\\
Unsupervised loss weight ($\lambda$) controls the proportion of the total loss attributed to the unsupervised loss for the target domain. As the aim is to adapt the model to the target domain, higher value of $\lambda$ generally leads to better performance. Fig.~\ref{figure:lambda-weights} shows the ablation results for different values of unsupervised loss weight ($\lambda$). We observe that the increase in the $\lambda$ increases the accuracy; however, it starts to decrease after the $\lambda$ value of 4.0.}

\subsection{AdaptOR-SSL: semi-supervised learning (SSL) on source-domain}
Table \ref{table:source:ssl} shows the results of SSL experiments using \emph{AdaptOR-SSL} on the COCO dataset with 1\%, 2\%, 5\%, and 10\% labeled supervision. The results with 100\% labeled supervision are presented in Table \ref{table:source}. The first two rows in Table \ref{table:source:ssl} show the results of two fully supervised baselines: \emph{supervised} and \emph{supervised++}. The \emph{supervised} baseline uses random-resize and random-flip augmentations as used \citep{he2017mask}, whereas the \emph{supervised++} uses the our data augmentation pipeline containing \emph{weakly} and \emph{strongly} augmented labeled images. We observe significant improvement in the results by utilizing our data augmentation pipeline. We also compare our bounding box detection results with the current state-of-the-art SSL approach for object detection, Unbiased-teacher \citep{liu2021unbiased}: a multi-class object bounding box detection approach using \emph{self-training} and \emph{mean-teacher} based SSL approach. Different from ours, it uses fully supervised ImageNet weights for initialization and does not exploit the \emph{transformation equivariance constraints} using geometric augmentations. As the Unbiased-teacher performs bounding box detection on 80 COCO classes, we compare our results with their person category results \emph{$AP_{person}^{bb}$} from the model obtained from their GitHub repository\footnote{\url{https://github.com/facebookresearch/unbiased-teacher}}. We observe significant improvement in results attributed to our initialization using the self-supervised method, feature normalization using GN, exploitation of the geometric constraints on the unlabeled data, and single class training for person category exploiting mask and the keypoint annotations. Fig.~\ref{tab:table-qual3} shows the qualitative results from the models trained with 1\%, 2\%, 5\% and 100\% labels. We also show the qualitative results in the supplementary video on some YouTube videos and observe comparable qualitative results with 1\% of labeled supervision w.r.t 100\% of labeled supervision.

{\blue \subsection{Domain adaptation on AdaptOR-SSL model}
Table \ref{table:init-exps} shows results when we evaluate \emph{AdaptOR-SSL} models trained using 1\%, 2\%, 5\%, and 10\% source domain labels to our \emph{MVOR+} target domain. We observe a significant decrease in the results (see ``Before UDA'' results in the Table \ref{table:init-exps}). As a final experiment, we initialize our \emph{AdaptOR} approach with \emph{AdaptOR-SSL} model trained using 10\% source domain labels. We observe an increase in the performance after the domain adaptation. However, there still exists a gap of around 4\% for \emph{$AP_{person}^{bb}$} and \emph{$AP_{person}^{bb\ (from\ mask)}$}, and 12\% for \emph{$AP_{person}^{kp}$}. These results show the need to develop effective domain adaptation approaches in the presence of limited source domain labels.}

\section{Conclusion}
Manual annotations, especially for spatial localization tasks, are considered the main bottleneck in the design of AI systems. With advances in digital technology providing a wide variety of visual signals, the modern OR has started to use AI to develop next-generation smart assistance systems. However, the progress is hindered due to the cost and privacy concerns for obtaining manual annotations. In this work, we tackle the joint person pose estimation and instance segmentation task needed to analyze OR activities and propose an unsupervised domain adaptation approach to adapt a model trained on a labeled source domain to an unlabeled target domain. We propose a new \emph{self-training} based framework with advanced data augmentations to generate pseudo labels for the unlabeled target domain. The high-quality effectiveness in the pseudo labels is ensured by applying explicit geometric constraints of the different augmentations on the unlabeled input image. We also introduce disentangled feature normalization for the statistically different source and the target domains and use the \emph{mean-teacher} paradigm to stabilize the training. Evaluation of the method on the two target domain datasets, \emph{MVOR+} and \emph{TUM-OR-test}, with extensive ablation studies, show the effectiveness of our approach. We further demonstrate that the proposed approach can effectively be adapted to the low-resolution images of the target domain, as needed to ensure OR privacy, even up to a downsampling factor of 12x. Finally, we illustrate the generality of our approach as the SSL method on the large-scale \emph{COCO} dataset, where we obtain better results with as few as 1\% of labeled annotations.
%a domain adaptation approach that does not require manual labels from the target OR domain for multi-task of joint person bounding box detection, human pose estimation, and instance segmentation. The significant improvement against the competitive baselines and the generalizability on the unconstrained \emph{in the wild} images as an SSL method could potentially scale our approach to other target domains as well. 
% The unsupervised learning - the hallmark of human intelligence -
% ``The revolution will not be supervised" \cite{efrosbeyondsupervised}. 
\section{Acknowledgements}
This work was supported by French state funds managed by the ANR within the Investissements d'Avenir program under reference ANR-16-CE33-0009 (DeepSurg) and ANR-10-IAHU-02 (IHU Strasbourg). This work was granted access to the HPC resources of IDRIS under the allocation 20XX-[AD011011631R1] made by GENCI.
\bibliographystyle{8_model2-names.bst}\biboptions{authoryear}
\bibliography{9_references}

\end{document}